\documentclass[10pt,twocolumn,letterpaper]{article}

\usepackage[pagenumbers]{cvpr} 

\usepackage{algorithm}
\usepackage{algpseudocode}
\usepackage[table]{xcolor}
\usepackage{multirow}
\usepackage{cuted}
\usepackage{colortbl}

\usepackage[export]{adjustbox}
\usepackage{siunitx,xcolor,colortbl,booktabs}
\definecolor{rowgray}{RGB}{248,248,248}
\newcolumntype{d}{S[table-format=1.2]} 
\definecolor{cvprblue}{rgb}{0.21,0.49,0.74}
\usepackage[pagebackref,breaklinks,colorlinks,allcolors=cvprblue]{hyperref}

\usepackage{bbm}
\usepackage{float}
\usepackage{comment}

\def\paperID{40753}
\def\confName{CVPR}
\def\confYear{2026}

\usepackage{array,booktabs,multirow,makecell}
\newcolumntype{L}{>{\raggedright\arraybackslash}m{0.17\textwidth}}
\newcolumntype{C}{>{\centering\arraybackslash}m{0.055\textwidth}}
\definecolor{AUCgray}{gray}{0.95}
\definecolor{APgray}{gray}{1}

\title{Training-free Detection of Generated Videos via Spatial-Temporal Likelihoods}
\author{Omer Ben Hayun, Roy Betser, Meir Yossef Levi, Levi Kassel, Guy Gilboa\\
Technion -- Israel Institute of Technology, Haifa, Israel\\
{\tt\small \{omerben,roybe,me.levi,kassellevi\}@campus.technion.ac.il; guy.gilboa@ee.technion.ac.il}
}

\usepackage{listings}
\usepackage{xcolor}

\definecolor{stringcolor}{RGB}{186,33,33}  

\lstdefinestyle{numberedlist} {
    language=Python,
    basicstyle=\ttfamily\small,
    keepspaces=true,
    columns=flexible,
    numbers=left,
    numberstyle=\tiny\color{gray},
    numbersep=5pt,
    stepnumber=1,
    stringstyle=\color{stringcolor},
    showstringspaces=false,
    escapeinside={@}{@},
}

\lstdefinestyle{smallsize} {
    language=Python,
    basicstyle=\ttfamily\small,
    keepspaces=true,
    columns=flexible,
    stringstyle=\color{stringcolor},
    showstringspaces=false,
    escapeinside={@}{@},
}

\usepackage{pifont}
\newcommand{\cmark}{\ding{51}} 
\newcommand{\xmark}{\ding{55}} 

\usepackage{amsthm}
\newtheorem{theorem}{Theorem}
\newtheorem{lemma}[theorem]{Lemma}

\usepackage{graphicx}

\newcommand{\perturbationtabular}[2][0.16\linewidth]{%
  \begin{adjustbox}{max width=0.9\textwidth}
  \begin{tabular}{|>{\raggedright\arraybackslash}c|ccccc|}
    \toprule
    Original
      & \multicolumn{5}{c|}{\includegraphics[valign=c,width=#1]{#2/0_original.png}} \\
    \midrule\midrule
        & \multicolumn{5}{c|}{Severity level} \\
    \cmidrule(lr){2-6}
        & 1 & 2 & 3 & 4 & 5 \\
    \midrule
    Gaussian blur
    &\includegraphics[valign=c,width=#1]{#2/1_gaussian_blur_level1.png}
      & \includegraphics[valign=c,width=#1]{#2/1_gaussian_blur_level2.png}
      & \includegraphics[valign=c,width=#1]{#2/1_gaussian_blur_level3.png}
      & \includegraphics[valign=c,width=#1]{#2/1_gaussian_blur_level4.png}
      & \includegraphics[valign=c,width=#1]{#2/1_gaussian_blur_level5.png} \\
    JPEG compression
      & \includegraphics[valign=c,width=#1]{#2/2_jpeg_compression_level1.png}
      & \includegraphics[valign=c,width=#1]{#2/2_jpeg_compression_level2.png}
      & \includegraphics[valign=c,width=#1]{#2/2_jpeg_compression_level3.png}
      & \includegraphics[valign=c,width=#1]{#2/2_jpeg_compression_level4.png}
      & \includegraphics[valign=c,width=#1]{#2/2_jpeg_compression_level5.png} \\
    Random resized crop
      & \includegraphics[valign=c,width=#1]{#2/3_resize_crop_level1_85pct.png}
      & \includegraphics[valign=c,width=#1]{#2/3_resize_crop_level2_70pct.png}
      & \includegraphics[valign=c,width=#1]{#2/3_resize_crop_level3_50pct.png}
      & \includegraphics[valign=c,width=#1]{#2/3_resize_crop_level4_30pct.png}
      & \includegraphics[valign=c,width=#1]{#2/3_resize_crop_level5_8pct.png} \\
    Gaussian noise
      & \includegraphics[valign=c,width=#1]{#2/4_gaussian_noise_level1.png}
      & \includegraphics[valign=c,width=#1]{#2/4_gaussian_noise_level2.png}
      & \includegraphics[valign=c,width=#1]{#2/4_gaussian_noise_level3.png}
      & \includegraphics[valign=c,width=#1]{#2/4_gaussian_noise_level4.png}
      & \includegraphics[valign=c,width=#1]{#2/4_gaussian_noise_level5.png} \\
    \bottomrule
  \end{tabular}
  \end{adjustbox}%
}

\begin{document}

\maketitle

\begin{strip}
	\centering
	\includegraphics[width=0.8\textwidth]{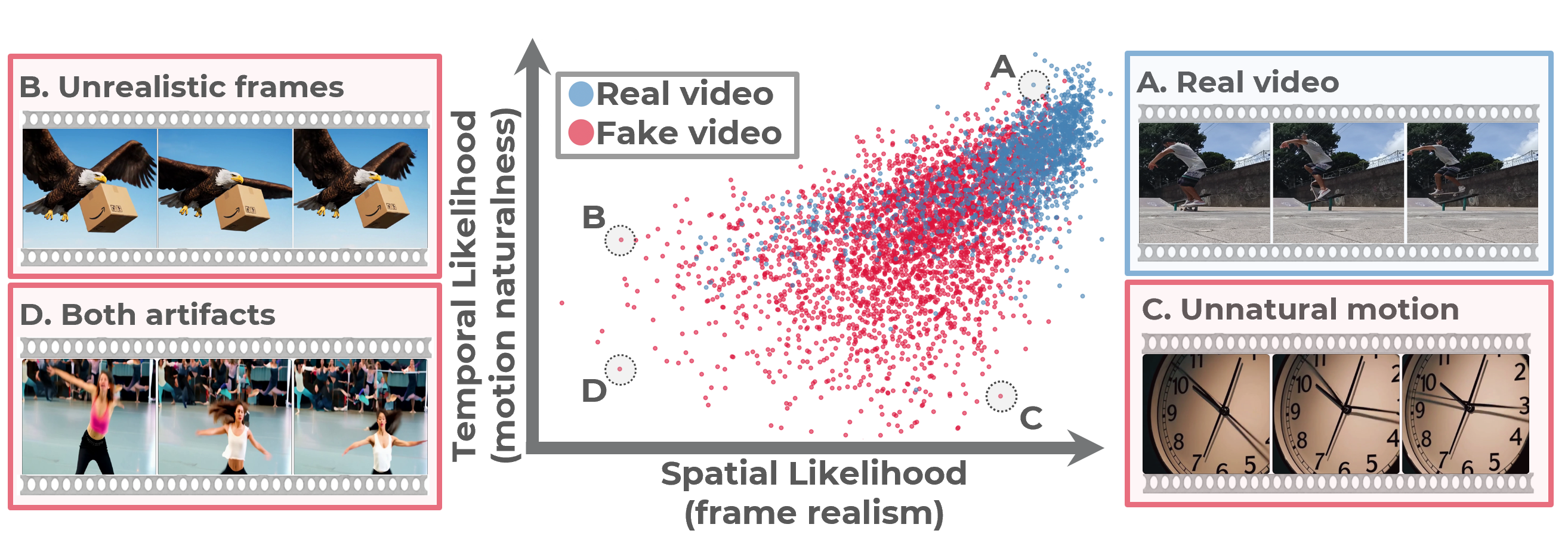}
	\captionsetup{hypcap=false}
	\captionof{figure}{\textbf{Spatio-temporal likelihoods per video.}
		Blue: real; red: fake (ComGenVid). Joint spatio-temporal likelihoods clearly separate real and fake videos; examples illustrate high/low spatial likelihood (frame realism) and temporal likelihood (motion naturalness).}
	\label{fig:teaser}
\end{strip}

\begin{abstract}
	Following major advances in text and image generation, the video domain has surged, producing highly realistic and controllable sequences.
	Along with this progress, these models also raise serious concerns about misinformation, making reliable detection of synthetic videos increasingly crucial.
	Image-based detectors are fundamentally limited because they operate per frame and ignore temporal dynamics, while supervised video detectors generalize poorly to unseen generators, a critical drawback given the rapid emergence of new models.
	These challenges motivate zero-shot approaches, which avoid synthetic data and instead score content against real-data statistics, enabling training-free, model-agnostic detection.
	We introduce \emph{STALL}, a simple, training-free, theoretically justified detector that provides likelihood-based scoring for videos, jointly modeling spatial and temporal evidence within a probabilistic framework.
	We evaluate STALL on two public benchmarks and introduce \emph{ComGenVid}, a new benchmark with state-of-the-art generative models.
	STALL consistently outperforms prior image- and video-based baselines.
	Code and data are available \href{https://omerbenhayun.github.io/stall-video}{here}.

\end{abstract}

\section{Introduction}
Generative modeling has progressed rapidly across modalities, enabling powerful text and image-generation capabilities built on large language models and diffusion-based image synthesizers~\citep{ho2020denoising, song2019generative, rombach2022high, raffel2020exploring, brown2020language}.
After major breakthroughs in text and image generation, the video domain has undergone a sharp leap forward in the past few years, with highly realistic, controllable video generation models producing long, high-fidelity sequences~\citep{chen2024videocrafter2, hacohen2024ltx, wan2025wan}.
These advances unlock strong benefits for creative workflows, content production, and media automation~\citep{times2025, using2023}.
At the same time, synthetic videos can be misused for misinformation, fraud, impersonation, and intellectual-property violations~\citep{appel2023generative, dandodiary2025deepfake, euiphelpdesk2024deepfake}, prompting platforms and regulators to require disclosure of AI-generated content and underscoring the urgency of reliable detection~\citep{kalra2025india, itu2025deepfakes}. Unlike deepfake detection, which focuses on manipulation of real content, we address a different problem: detecting fully generated videos, where every frame is synthetic.

In the image domain, early studies mainly relied on supervised classifiers, typically CNN-based models trained to distinguish real from synthetic images using large, labeled datasets~\citep{baraheem2023ai, bird2024cifake, cioni2024clip}.
While effective on known generators, these methods require continuous retraining as new generative models emerge and thus generalize poorly to unseen ones~\citep{epstein2023online}.
To reduce dependence on synthetic training data, later works explored unsupervised and semi-supervised approaches, leveraging large pretrained models~\citep{ojha2023towards, cozzolino2023raising}.
Recently, zero-shot image detectors have emerged, showing improved robustness and generalization~\citep{ricker2024aeroblade, he2024rigid, cozzolino2024zero, brokman2025manifold}. In this context, zero-shot means no additional training and no generated content available.
However, when applied to videos, image detectors assess authenticity only on a per-frame basis. As a result, they ignore temporal dependencies and miss artifacts that emerge across time, such as motion inconsistencies, that are invisible in any single frame.

In the video domain, progress has been more limited. Recent methods predominantly use supervised training to detect generated videos~\cite{vahdati2024beyond, bai2024ai, chen2024demamba, interno2025ai,zhang2025NSGVD}, but they inherit the same limitations as supervised image detectors: they require large labeled datasets and generalize poorly to unseen generators.
The first zero-shot detector for generated videos is D3~\citep{zheng2025d3}, introduced only recently.
It analyzes transitions between consecutive frames and relies solely on temporal cues, while ignoring per-frame visual content and spatial information. Moreover,
it lacks principled theoretical foundations, relying primarily on empirical hypotheses about real video dynamics.
Therefore, a critical gap remains: the need for a mathematically grounded video detector that jointly analyzes spatial content and temporal dynamics.

\begin{figure*}[t]
	\centering
	\includegraphics[width=0.9\textwidth]{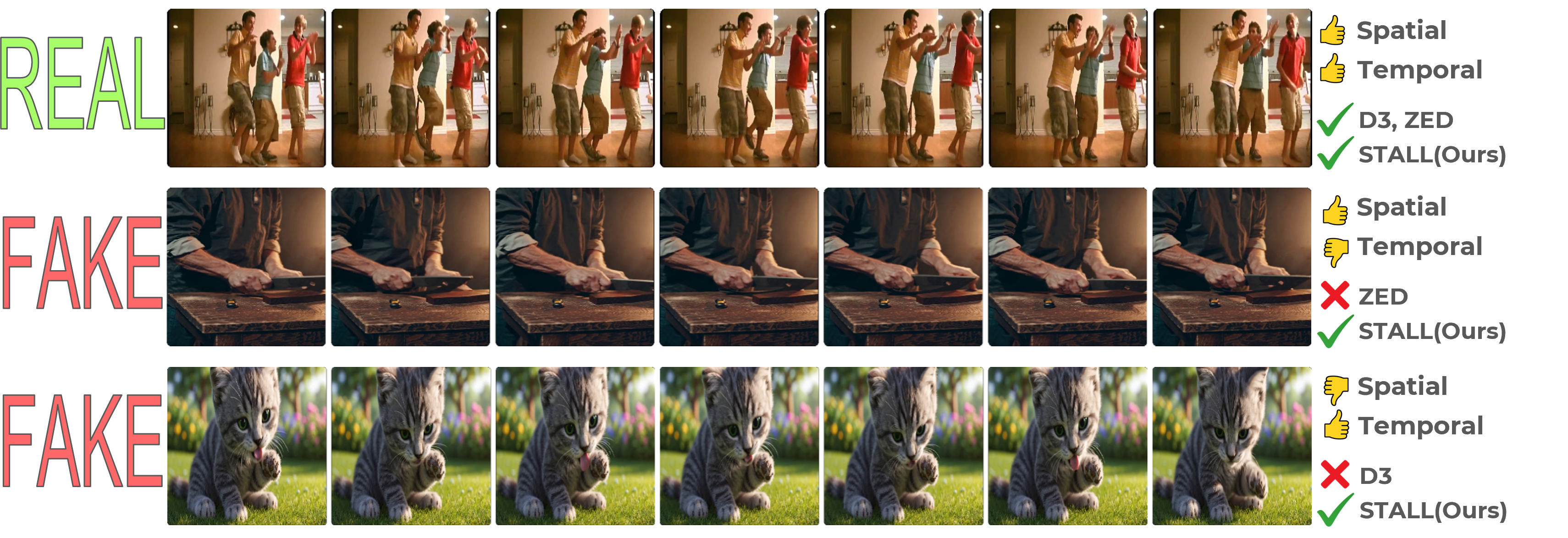}
	\caption{\textbf{Qualitative comparison of ZED, D3, and our method (STALL).} Each row shows a video clip with natural or unnatural spatial and temporal behavior, together with the corresponding predictions. ZED (spatial-only) misses in cases dominated by temporal inconsistency; D3 (temporal-only) fails when spatial realism is misleading. STALL fuses spatial and temporal likelihoods, yielding robust detection when either modality alone is insufficient. Additional examples with more details are given in Supp. D.7.}
	\label{fig:examples}
\end{figure*}

To address this gap, we introduce STALL, a zero-shot video detector that accounts for both spatial and temporal dimensions when determining whether a video is real or generated (see illustration in \Cref{fig:teaser}).
Our method leverages a probabilistic image-domain approach~\citep{betser2025whitened} and uses DINOv3~\citep{siméoni2025dinov3} to compute image likelihoods. We extend this approach with a temporal likelihood term that captures the consistency of transitions between frames.
Unlike prior approaches that are supervised, rely solely on spatial cues (image detectors), or focus exclusively on temporal dynamics~\cite{zheng2025d3}, our formulation jointly models both aspects and detects inconsistencies that emerge from their interaction (see \Cref{fig:examples} for qualitative examples).
STALL assumes access to a collection of real videos in its pre-processing stage, termed the \emph{calibration set}. With the abundance of publicly available videos, this is a very mild requirement.
The approach is training-free and requires no access to generated samples from any model.
The core of the algorithm is based on a new spatio-temporal likelihood model of real videos. This yields a principled measure of how well a video aligns with real-data statistics in space and time.

Our method achieves state-of-the-art performance on two established benchmarks~\citep{chen2024demamba, he2024videoscore} and on our newly introduced dataset comprising videos from recent high-performing generators~\citep{sora2024openai, veo32025deepmind}.
We curate this dataset to reflect the newest wave of high-fidelity video models, enabling evaluation on frontier systems.
The method is lightweight and efficient, operating without training, and is thus suitable for real-time or large-scale screening pipelines.
Across all experiments, it remains robust to common image perturbations, variations in frames-per-second (FPS), and ranges of hyperparameter settings.
Our main contributions are as follows:

\begin{itemize}
	\item \textbf{Temporal likelihood.} We extend spatial (image-domain) likelihoods to \emph{temporal} frame-to-frame transitions.
	\item \textbf{Theory-grounded zero-shot video detector.} A detector derived from a well-defined theory, which we empirically validate. This provides a principled, measurable tool for analyzing and debugging edge cases.
	\item \textbf{State-of-the-art across benchmarks.} We achieve state-of-the-art results on three challenging benchmarks and perform extensive evaluations demonstrating robustness and consistent performance across settings.
	\item \textbf{New benchmark.} We release \emph{ComGenVid}, a curated benchmark featuring recent high-fidelity video generators (e.g., Sora, Veo-3) to support future research.
\end{itemize}

\section{Background and Related work}
\noindent \textbf{Generated image detection.}
Early work trained supervised CNNs on labeled real and synthetic datasets, sometimes emphasizing hand-crafted artifacts, but generalization to unseen generators was limited~\citep{wang2020cnn,baraheem2023ai,epstein2023online,bird2024cifake,bammey2023synthbuster,martin2023detection,zhong2023rich,wang2023dire}.
Few-shot and semi or unsupervised variants improved data efficiency by leveraging pretrained features, yet typically retained some dependence on synthetic data or generator assumptions~\citep{zhang2022exposing,cioni2024clip,ojha2023towards,sha2023fake,cozzolino2023raising}.
Zero-shot methods avoid synthetic content exposure by comparing an image to transformed or reconstructed variants~\citep{ricker2024aeroblade, he2024rigid, cozzolino2024zero, brokman2025manifold}.
However, these image-only approaches are confined to per-frame spatial cues and ignore cross-frame temporal consistency, leaving them blind to anomalies that only manifest in motion or inter-frame transitions.

\noindent \textbf{Generated video detection.}
Unlike \emph{deepfakes}, which edit real footage (e.g., face swaps or lip-sync), we target \emph{fully generated} content, where the video is synthesized from scratch.
Supervised detectors train on labeled real and synthetic videos and report strong in-domain results but struggle in unseen models regime~\citep{vahdati2024beyond, bai2024ai}.
Recent work also couples new benchmarks with architectures: GenVideo with the DeMamba module~\citep{chen2024demamba}; VideoFeedback, which also presents VideoScore (human-aligned automatic scoring)~\citep{he2024videoscore}.
Parallel efforts explore MLLM-based supervised detectors that provide rationales but still require curated training data and tuning~\citep{wen2025busterx,gao2025david}.
D3, the first \emph{zero-shot} video detector, relies on second-order temporal differences, focusing on motion cues~\citep{zheng2025d3}. A similar first-order approach is presented in~\citep{brokman2026training}.
In contrast, our approach is directly probabilistic and \emph{jointly} scores spatial (per-frame) and temporal (inter-frame) likelihoods, addressing both appearance and dynamics in a single framework.

\noindent \textbf{Gaussian embeddings and likelihood approximation.}
Modern visual encoders such as CLIP~\citep{radford2021learning} learn high-dimensional embedding spaces with rich semantic structure.
Empirical studies have characterized geometric phenomena in CLIP representations, including the \emph{modality gap}, \emph{narrow-cone} concentration~\citep{liang2022mind}, and a \emph{double-ellipsoid} structure~\citep{levi2025double}.
Recent work demonstrates that CLIP embeddings are well-approximated by Gaussian distributions, enabling closed-form image likelihood approximation using whitening without additional training~\citep{betser2026general}. Whitening has also been shown effective for LLM activations~\citep{rachmil2025training}.
From a theoretical angle, the Maxwell–Poincaré lemma implies that uniform normalized high-dimensional features have approximately Gaussian projections~\citep{diaconis1984asymptotics}.
This principle has recently been leveraged to show that the InfoNCE objective asymptotically induces Gaussian structure in learned embeddings~\citep{betser2026infonce}.
Motivated by both empirical evidence and theoretical guarantees, we introduce a normalization step in the temporal embedding space to promote Gaussian statistics and compute faithful likelihood estimates.
Additionally, this Gaussian modeling approach extends to other vision encoders~\citep{siméoni2025dinov3,howard2019searching} and forms the basis of our spatio-temporal video likelihood score.

\begin{figure*}[t]
	\centering
	\includegraphics[width=\textwidth]{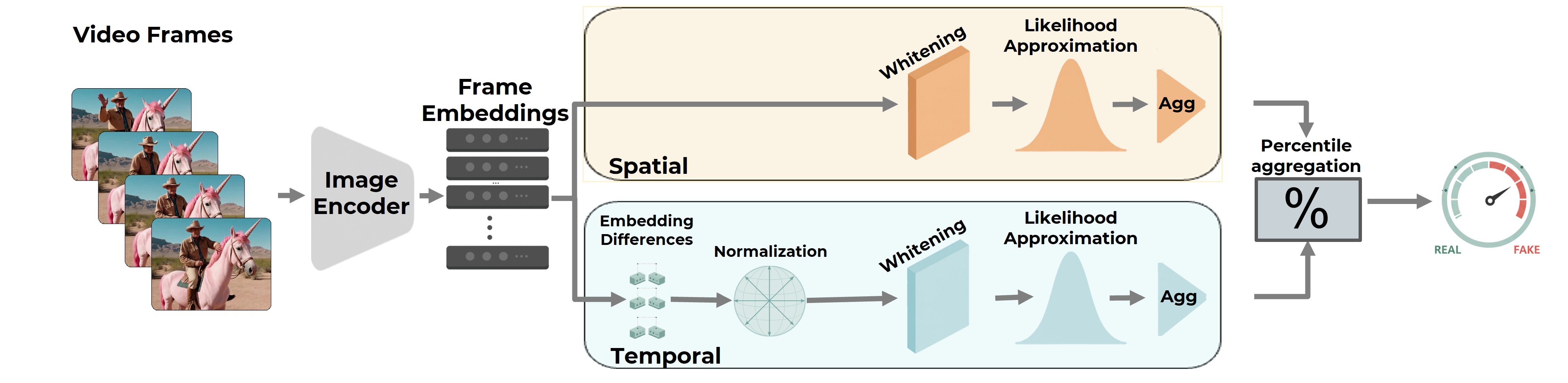}
	\caption{\textbf{Method overview.} A video is split into frames and encoded into embeddings. The spatial branch scores the likelihood of each frame embedding; the temporal branch normalizes inter-frame differences and scores their likelihood analogously. The two scores are then fused into a unified measure that separates AI-generated from real videos.}
	\label{fig:overview}
\end{figure*}

\section{Preliminaries}
We now introduce the mathematical tools and notations used throughout the paper. These concepts form the basis of our likelihood formulation and will be applied in the method section (\Cref{sec:method}).

\subsection{Whitening transform and Gaussian likelihood}
\noindent \textbf{Notation.}
Let $\mathcal{X}=\{x_i\}_{i=1}^N \subset \mathbb{R}^d$ and let $\mathbf{X}\in\mathbb{R}^{d\times N}$ be the column-stacked matrix.
Define the sample mean $\mu=\tfrac{1}{N}\sum_{i=1}^N x_i$ and centered vectors $\hat{x}_i=x_i-\mu$, with $\mathbf{\hat{X}}=[\hat{x}_1,\dots,\hat{x}_N]$.
The empirical covariance is $\mathbf{\Sigma}=\tfrac{1}{N}\mathbf{\hat{X}}\mathbf{\hat{X}}^\top$.

\noindent \textbf{Whitening transform.}
\label{sec:whitening}
We seek a linear transform $W\in\mathbb{R}^{d\times d}$ that admits:
\begin{equation}
	W^\top W \;=\; \mathbf{\Sigma}^{-1}.
	\label{eq:whiten_equiv}
\end{equation}
The whitening matrix is \emph{not unique}: if $W$ satisfies \Cref{eq:whiten_equiv}, then so does $RW$ for any orthogonal $R$. A common choice is PCA-whitening. Let the eigen-decomposition be $\mathbf{\Sigma}=V\Lambda V^\top$ with eigenvectors $V$ and eigenvalues $\Lambda=\mathrm{diag}(\lambda_1,\dots,\lambda_d)$. The PCA-whitening matrix is
\begin{equation}
	W_{\text{PCA}} \;=\; \Lambda^{-\tfrac{1}{2}} V^\top.
	\label{eq:pca_whiten}
\end{equation}
Given a vector $x$, the whitened representation is $y \;=\; W(x-\mu)$ and the whitened data matrix is
\begin{equation}
	\mathbf{Y} \;=\; W\,\mathbf{\hat{X}}.
	\label{eq:whiten_apply}
\end{equation}
Whitened embeddings have zero mean and identity covariance.

\noindent \textbf{Likelihood approximation.}
Under the zero-mean and identity-covariance properties, if the whitened coordinates follow a Gaussian distribution, then $y \sim \mathcal{N}(0, I_d)$.
Given this isotropic Gaussian model, the log-likelihood is:
\begin{equation}
	\ell(y) \;=\; \log p(y) \;=\; -\tfrac{1}{2}\Big(d\log(2\pi) + \|y\|_2^2\Big),
	\label{eq:isotropic_log_likelihood}
\end{equation}
where $\|y\|_2^2 = y^\top y$.
Given an embedding $x$, the whitened norm $\|W(x-\mu)\|_2^2$ thus provides a closed-form likelihood proxy \emph{when} the Gaussian assumption holds.

\subsection{Asymptotic Gaussian projections}
\label{sec:gaussian}
When vectors are uniformly distributed on the unit sphere in $\mathbb{R}^d$, their coordinates behave approximately Gaussian.
The Maxwell-Poincaré lemma~\cite{diaconis1984asymptotics,diaconis1987dozen} formalizes this: if $u \sim \mathrm{Unif}(\mathbb{S}^{d-1})$, then for each coordinate,
\begin{equation}
	\sqrt{d}\,u_j \xrightarrow[d\to\infty]{} \mathcal{N}(0,1).
	\label{eq:maxwell_poincare}
\end{equation}
More generally, for high-dimensional vectors with nearly uniform directions and concentrated norms, any fixed low-dimensional linear projection is well-approximated by a Gaussian.
Supplementary Material (Supp.) Section B.3 details the lemma and convergence rates.

\begin{algorithm}[t]
	\small
	\caption{STALL (Generated video detector)}
	\label{alg:method}
	\begin{algorithmic}[1]
		\Require Encoder $E$, calibration set $\mathcal{C}$, test video $v = \{f_t\}_{t=1}^{T}$
		\Statex
		\Statex \textbf{Calibration (pre-processing)}
		\State From $\mathcal{C}$: extract all frames per video, encode $\{x_{i,t}\}$.
		\State Compute spatial statistics $(\mu, W)$ (~\Cref{eq:pca_whiten}), using a single frame per video.
		\State Compute the temporal statistics $(\mu_\Delta, W_\Delta)$ (~\Cref{eq:pca_whiten}) using all normalized inter-frame differences for each video.
		\State Record calibration distributions of $s_{C_\mathrm{sp}}$ and $s_{C_\mathrm{temp}}$.
		\Statex
		\setcounter{ALG@line}{0}
		\Statex \textbf{Inference (test-time)}
		\State Compute $x_t = E(f_t)$; for frames $\{f_t\}_{t=1}^{T}$ of $v$.
		\State $y_t \gets W(x_t - \mu)$; for $t \in \{1, \cdots, T\}$
		\State $z_t \gets W_\Delta\!\left(\frac{x_{t+1}-x_t}
			{\|x_{t+1}-x_t\|} - \mu_\Delta\right)$; for $t \in \{1, \cdots, T-1 \}$

		\State $s_{\mathrm{sp}} \gets \text{max}(\{\ell(y_t)\}_{t=1}^{T})$ ($\ell(y_t)$ computed by~\Cref{eq:isotropic_log_likelihood}).
		\State $s_{\mathrm{temp}} \gets \text{min}(\{\ell(z_t)\}_{t=1}^{T-1})$ ($\ell(z_t)$ follows~\Cref{eq:isotropic_log_likelihood}).
		\State \Return $s_{\mathrm{video}} \gets \tfrac{1}{2}\big(\mathrm{perc}(s_{\mathrm{sp}})
			\;+\; \mathrm{perc}(s_{\mathrm{temp}})\big)$.

	\end{algorithmic}
\end{algorithm}

\section{Method: STALL}
\label{sec:method}
We propose \emph{STALL} (\textbf{S}patial-\textbf{T}emporal \textbf{A}ggregated \textbf{L}og-\textbf{L}ikelihoods), a zero-shot detector that \emph{jointly} scores videos via a spatial likelihood over per-frame embeddings and a temporal likelihood over inter-frame transitions.
A high-level overview of the method is shown in~\Cref{fig:overview}, and~\Cref{alg:method} summarizes the procedure.
Detailed algorithms for all method steps are provided in Supp. Section A.1.

\noindent \textbf{Notation.}
Let $\mathcal{C}=\{c^{(i)}\}_{i=1}^{N}$ denote a collection of videos.
A video $c \in \mathcal{C}$ consists of $T$ frames, written as $c=\{f_t\}_{t=1}^{T}$.
Each frame $f_t$ is mapped to a $d$-dimensional embedding using a vision encoder $E$, yielding $x_t = E(f_t) \in \mathbb{R}^{d}$.

\subsection{Spatial likelihood}
Prior work~\citep{betser2025whitened} in the image domain observed that \emph{whitened} CLIP embeddings are well-approximated by standard Gaussian coordinates, as verified on MSCOCO~\citep{lin2014microsoft}, using Anderson–Darling (AD) and D’Agostino–Pearson (DP) normality tests~\citep{anderson1954test, d1973tests}. Therefore, the norm in the whitened space correlates with the likelihood of an image.
We extend this result to the \emph{video} setting by extracting frame-level embeddings from real video datasets. We apply the whitening procedure discussed above (\Cref{sec:whitening}), and assess Gaussianity with the same tests, evaluating multiple encoders.
Under this Gaussian assumption, per-frame spatial likelihoods follow the closed-form log-likelihood in~\Cref{eq:isotropic_log_likelihood}.
Details and results are in Supp. Section B.1.

We estimate spatial likelihood statistics using a calibration set of $n$ real videos (see \Cref{sec:calibration}).
This step involves no training and is computed \emph{a priori} only once. It consists of estimating real-data statistics, which remain fixed throughout inference.
At inference time, for a test video $v=\{f_t\}_{t=1}^{T}$, each frame $f_t$ is encoded as $x_t = E(f_t)$, whitened to $y_t$ using~\Cref{eq:whiten_apply}, and assigned a log-likelihood $\ell(y_t)$ according to~\Cref{eq:isotropic_log_likelihood}.


\subsection{Temporal likelihood}
Spatial likelihoods score frames independently; they do not assess how \emph{transitions} evolve across time.
To capture motion consistency, we examine the embedding space and model frame-to-frame differences, $\Delta_t \;=\; x_{t+1} - x_t$.

\noindent \textbf{Normalization Induces Gaussianity.}
Empirically, the raw transition vectors ${\Delta_t}$ are \emph{not} well modeled by a Gaussian distribution (see Supp. Section B.1).
We observe that these high-dimensional transitions exhibit two key properties:
(1) \emph{Variable magnitudes}; their norms vary substantially across samples; and
(2) \emph{Random directions}; their orientations are approximately spanned in a uniform manner, since the underlying video motions are arbitrary and thus lack any preferred direction; see Supp. Section B.1 for empirical validation.
In high-dimensional spaces, uniformly distributed directions on the sphere behave similarly to Gaussian samples when projected onto any axis, as established by the Maxwell–Poincaré lemma~\cite{diaconis1984asymptotics, diaconis1987dozen} (\Cref{sec:gaussian}).
To obtain a stable probabilistic model, we normalize each transition vector as
$\tilde{\Delta}_t = \frac{\Delta_t}{\lVert \Delta_t \rVert}$,
placing all transition directions on the unit sphere.
Empirically, these normalized transitions exhibit Gaussian-like behavior, see illustration in \Cref{fig:temp_norm} and  quantitative results in Supp. Section B.1.

\begin{figure}[t]
	\centering
	\captionsetup[subfigure]{justification=centering}
	\begin{subfigure}[t]{0.49\linewidth}
		\centering
		\includegraphics[width=\linewidth]{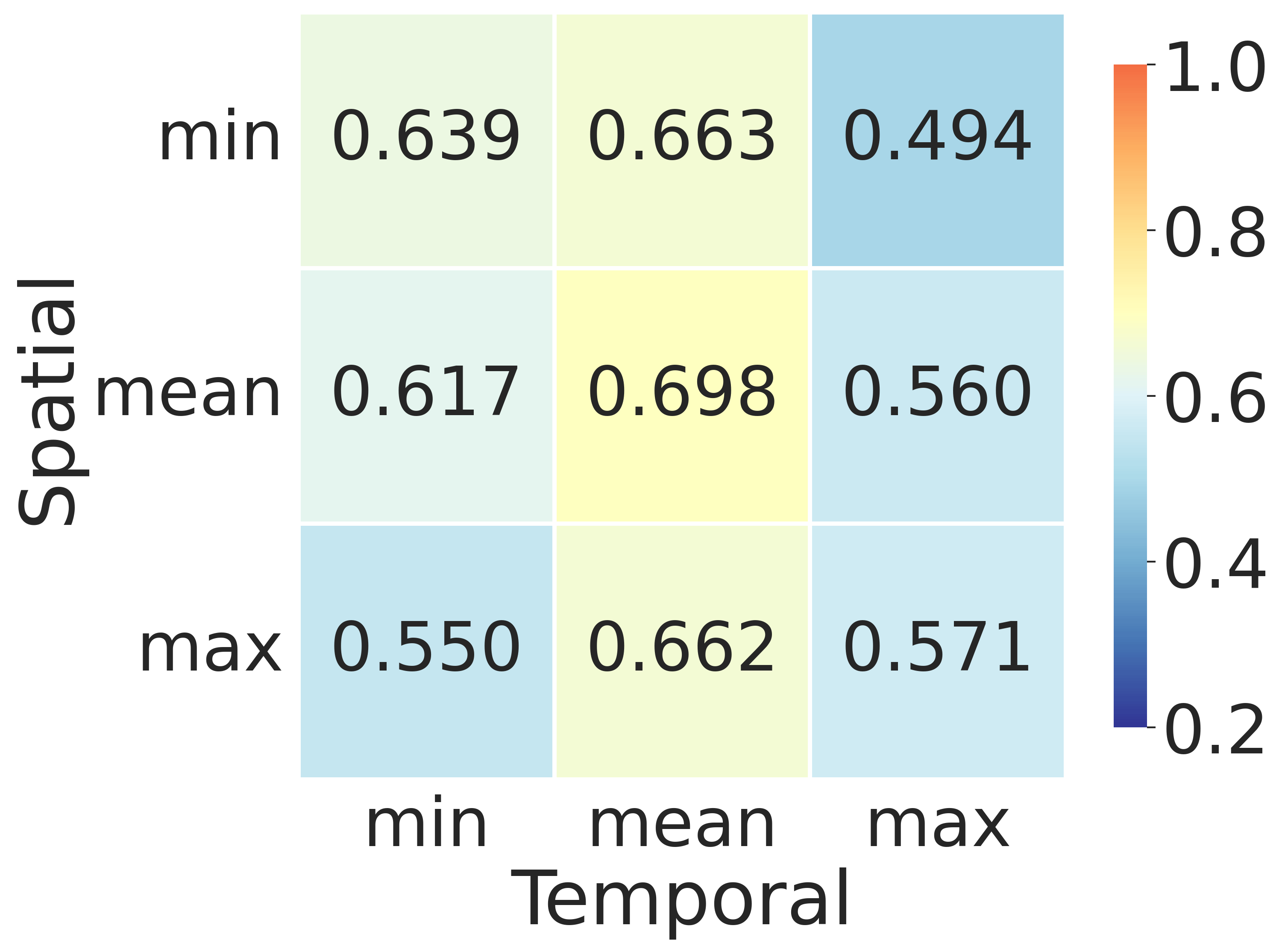}
		\caption{Pearson correlation}
		\label{fig:noise-auc}
	\end{subfigure}\hfill
	\begin{subfigure}[t]{0.49\linewidth}
		\centering
		\includegraphics[width=\linewidth]{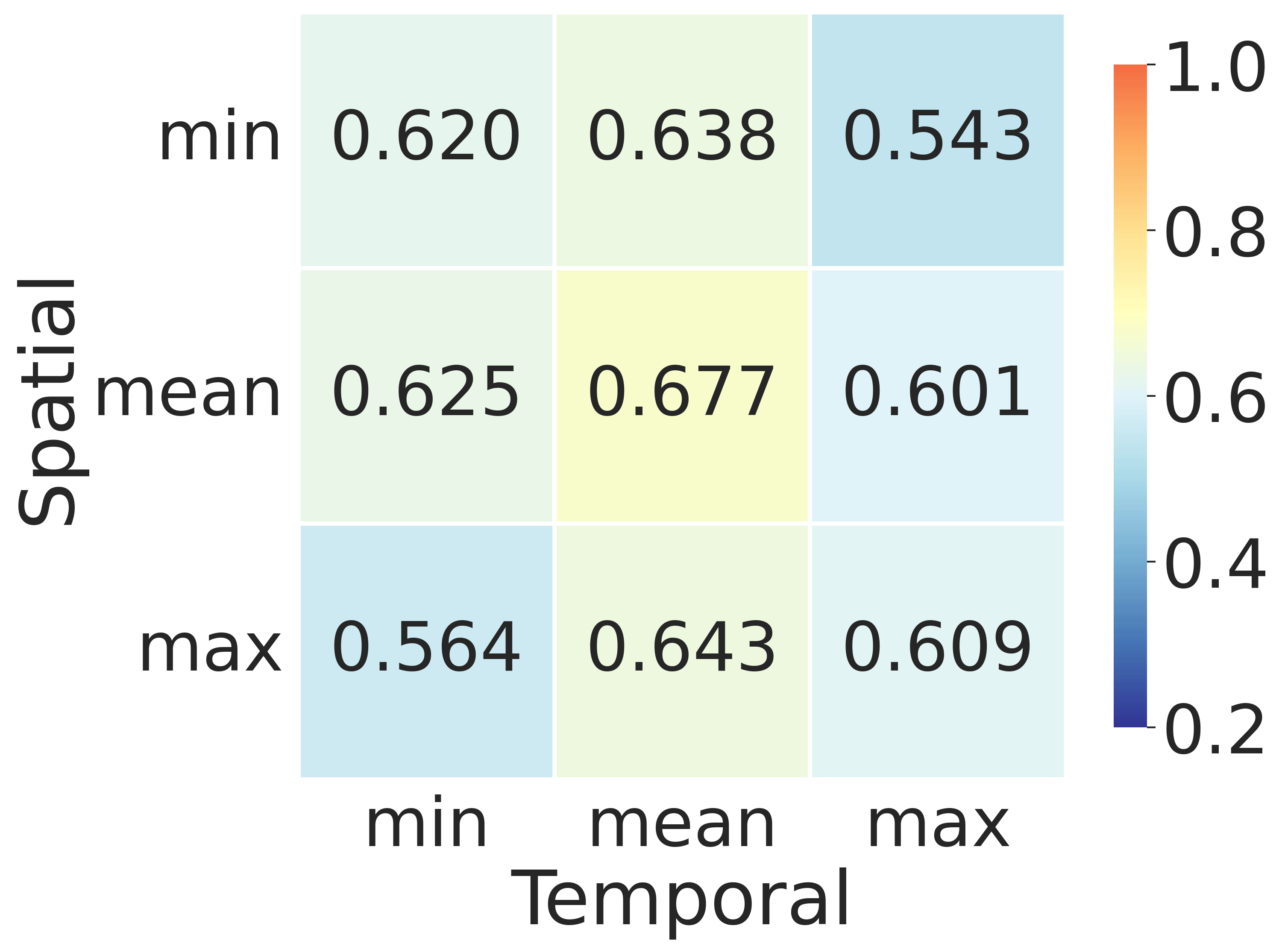}
		\caption{Spearman correlation}
		\label{fig:second}
	\end{subfigure}\hfill
	\caption{\textbf{Correlations among spatial and temporal aggregation methods.} Values computed on VATEX~\citep{wang2019vatex}, which is not used in our evaluations. When all individual likelihood detectors perform reasonably well on the evaluated benchmarks (see Supp.\ Section~D.2), lower correlations are desirable.}
	\label{fig:agg_metrics_corr}
\end{figure}


\noindent \textbf{Corner case:} if two consecutive frames are identical ($x_{t+1} = x_t$), their transition vector is $\Delta_t = x_{t+1} - x_t = 0$. Such transitions carry no temporal information and are deterministically discarded from the
temporal likelihood computation. If \emph{all} frames in a video are identical, i.e.,
$f_1 = f_2 = \dots = f_T$, the input effectively degenerates to a single image. In this case no temporal score is defined and the detector falls back to the spatial likelihood $s_{\mathrm{spatial}}(V)$, which analyzes the image domain.

\noindent Using the calibration set of real videos, we collect all normalized transition vectors
$\{\tilde{\Delta}_t\}$ and compute their empirical mean $\mu_{\Delta}$ and covariance
$\Sigma_{\Delta}$. At inference time, in a manner analogous to the spatial likelihood, we whiten the normalized transitions, $z_t$, using~\Cref{eq:whiten_apply}, and compute their log-likelihoods $\ell(z_t)$ according to~\Cref{eq:isotropic_log_likelihood}.
This yields the temporal log-likelihood of each transition in the video.
Generated videos often exhibit unnatural motion, resulting in transitions with low likelihood under this model.

\begin{table*}[htbp]
	\caption{\textbf{Zero-shot detection results.} Comparison of image- and video-based detectors on three benchmarks. Best in each row is \textbf{bold}; second best is \underline{underlined}. Our method achieves the highest average performance on all benchmarks and leads on most generators individually. It is also the only method that detects consistently across all models, maintaining AUC $> 0.5$.}
	\label{tab:zero-shot_results}
	\begin{adjustbox}{max width=\textwidth}
		\begin{tabular}{@{} | l l ||
				>{\columncolor{AUCgray}}d  >{\columncolor{APgray}}d   
				>{\columncolor{AUCgray}}d  >{\columncolor{APgray}}d   
				>{\columncolor{AUCgray}}d  >{\columncolor{APgray}}d | 
				>{\columncolor{AUCgray}}d  >{\columncolor{APgray}}d   
				>{\columncolor{AUCgray}}d  >{\columncolor{APgray}}d | 
				>{\columncolor{AUCgray}}d  >{\columncolor{APgray}}d |  
				@{}}
			\toprule
			\multirow{3}{*}{Benchmark}                                           & \multirow{3}{*}{Model}                                  & \multicolumn{6}{c|}{Image Detectors} & \multicolumn{6}{c|}{Video Detectors}                                                                                                                                                                                               \\
			                                                                     &                                                         &
			\multicolumn{2}{c}{\scriptsize AEROBLADE \cite{ricker2024aeroblade}} &
			\multicolumn{2}{c}{RIGID \cite{he2024rigid}}                         &
			\multicolumn{2}{c|}{ZED \cite{cozzolino2024zero}}                    &
			\multicolumn{2}{c}{D3 (L2) \cite{zheng2025d3}}                       &
			\multicolumn{2}{c|}{D3 (cos) \cite{zheng2025d3}}                     &
			\multicolumn{2}{c|}{\textbf{STALL (Ours)}}                                                                                                                                                                                                                                                                                                                                                                 \\
			\cline{3-14}
			                                                                     &                                                         & AUC                                  & AP                                   & AUC              & AP               & AUC              & AP               & AUC              & AP               & AUC              & AP               & AUC              & AP               \\
			\midrule
			\midrule
			\multirow[c]{12}{*}{VideoFeedback \cite{he2024videoscore}}           & AnimateDiff \cite{guo2023animatediff}                   & 0.57                                 & 0.55                                 & \underline{0.73} & \underline{0.74} & 0.65             & 0.62             & 0.49             & 0.49             & 0.61             & 0.57             & \textbf{0.83}    & \textbf{0.86}    \\
			                                                                     & Fast-SVD \cite{blattmann2023stable}                     & 0.52                                 & 0.51                                 & 0.54             & 0.56             & 0.45             & 0.48             & 0.76             & 0.77             & \underline{0.80} & \underline{0.79} & \textbf{0.89}    & \textbf{0.89}    \\
			                                                                     & LVDM \cite{he2022latent}                                & \textbf{0.88}                        & \textbf{0.90}                        & 0.65             & 0.57             & 0.76             & 0.70             & 0.42             & 0.49             & 0.31             & 0.41             & \underline{0.86} & \underline{0.89} \\
			                                                                     & LaVie                                                   & 0.50                                 & 0.50                                 & \underline{0.71} & \underline{0.73} & 0.29             & 0.37             & 0.51             & 0.47             & 0.49             & 0.46             & \textbf{0.81}    & \textbf{0.83}    \\
			                                                                     & ModelScope \cite{wang2023modelscope}                    & 0.60                                 & 0.56                                 & 0.66             & \underline{0.62} & \underline{0.69} & 0.59             & 0.51             & 0.52             & 0.42             & 0.46             & \textbf{0.81}    & \textbf{0.83}    \\
			                                                                     & Pika \cite{pika2023}                                    & 0.44                                 & 0.46                                 & 0.54             & 0.54             & 0.39             & 0.47             & \textbf{0.83}    & \textbf{0.84}    & \underline{0.81} & \underline{0.81} & 0.78             & 0.80             \\
			                                                                     & Sora \cite{sora2024openai}                              & 0.65                                 & 0.62                                 & 0.43             & 0.44             & 0.56             & \underline{0.62} & 0.62             & 0.56             & \underline{0.67} & 0.58             & \textbf{0.81}    & \textbf{0.82}    \\
			                                                                     & Text2Video\cite{khachatryan2023text2video}              & 0.67                                 & 0.63                                 & \underline{0.70} & \underline{0.68} & 0.55             & 0.49             & 0.15             & 0.33             & 0.22             & 0.36             & \textbf{0.83}    & \textbf{0.83}    \\
			                                                                     & VideoCrafter2 \cite{chen2024videocrafter2}              & 0.60                                 & 0.58                                 & \underline{0.80} & 0.76             & 0.53             & 0.50             & 0.69             & 0.71             & 0.80             & \underline{0.79} & \textbf{0.93}    & \textbf{0.94}    \\
			                                                                     & ZeroScope \cite{zeroscope_v2_2024}                      & \textbf{0.78}                        & \underline{0.78}                     & 0.65             & 0.59             & 0.70             & 0.62             & 0.35             & 0.45             & 0.35             & 0.44             & \textbf{0.78}    & \textbf{0.81}    \\
			                                                                     & Hotshot-XL~\citep{hotshotxl2023}                        & 0.20                                 & 0.34                                 & 0.51             & 0.58             & 0.44             & 0.45             & \underline{0.64} & \underline{0.67} & 0.60             & 0.62             & \textbf{0.79}    & \textbf{0.80}    \\
			\addlinespace
			\midrule
			\rowcolors{3}{}{}
			                                                                     & Average                                                 & 0.58                                 & 0.58                                 & \underline{0.63} & \underline{0.62} & 0.54             & 0.54             & 0.54             & 0.57             & 0.55             & 0.57             & \textbf{0.83}    & \textbf{0.85}    \\
			\midrule
			\multirow[c]{11}{*}{GenVideo \cite{chen2024demamba}}                 & Crafter \cite{chen2023videocrafter1}                    & 0.64                                 & 0.65                                 & 0.71             & 0.66             & 0.55             & 0.56             & \underline{0.79} & \textbf{0.82}    & 0.76             & 0.79             & \textbf{0.82}    & \underline{0.80} \\
			                                                                     & Gen2 \cite{esser2023structure}                          & 0.56                                 & 0.59                                 & 0.70             & 0.67             & 0.51             & 0.58             & \textbf{0.88}    & \textbf{0.90}    & \textbf{0.88}    & \textbf{0.90}    & \textbf{0.88}    & 0.89             \\
			                                                                     & Lavie \cite{wang2025lavie}                              & 0.58                                 & 0.59                                 & \underline{0.77} & \underline{0.76} & 0.39             & 0.42             & 0.68             & 0.68             & 0.67             & 0.68             & \textbf{0.85}    & \textbf{0.84}    \\
			                                                                     & ModelScope \cite{wang2023modelscope}                    & 0.60                                 & 0.60                                 & 0.62             & 0.59             & 0.61             & 0.57             & \underline{0.63} & \underline{0.64} & 0.60             & 0.63             & \textbf{0.78}    & \textbf{0.78}    \\
			                                                                     & MorphStudio \cite{morphstudio2023}                      & \underline{0.74}                     & \underline{0.73}                     & \underline{0.74} & 0.69             & 0.60             & 0.60             & 0.66             & 0.71             & 0.64             & 0.69             & \textbf{0.83}    & \textbf{0.84}    \\
			                                                                     & Show\_1 \cite{zhang2025show}                            & 0.48                                 & 0.50                                 & 0.53             & 0.52             & 0.45             & 0.47             & \underline{0.76} & \textbf{0.80}    & 0.75             & 0.79             & \textbf{0.82}    & \underline{0.80} \\
			                                                                     & Sora \cite{sora2024openai}                              & 0.73                                 & 0.70                                 & 0.49             & 0.48             & 0.71             & \underline{0.79} & \underline{0.75} & 0.75             & 0.74             & 0.74             & \textbf{0.79}    & \textbf{0.80}    \\
			                                                                     & WildScrape \cite{wei2023dreamvideocomposingdreamvideos} & 0.49                                 & 0.53                                 & 0.61             & 0.59             & 0.55             & 0.57             & \underline{0.65} & \textbf{0.69}    & 0.64             & \textbf{0.69}    & \textbf{0.72}    & 0.68             \\
			                                                                     & HotShot-XL \cite{hotshotxl2023}                         & 0.31                                 & 0.39                                 & \underline{0.64} & \underline{0.65} & 0.47             & 0.46             & 0.56             & 0.64             & 0.54             & 0.62             & \textbf{0.79}    & \textbf{0.78}    \\
			                                                                     & MoonValley \cite{moonvalley2022}                        & 0.75                                 & 0.78                                 & 0.72             & 0.66             & 0.63             & 0.72             & \textbf{0.81}    & \textbf{0.82}    & \textbf{0.81}    & \textbf{0.82}    & 0.72             & 0.75             \\
			\addlinespace
			\midrule
			                                                                     & Average                                                 & 0.59                                 & 0.61                                 & 0.65             & 0.63             & 0.55             & 0.57             & \underline{0.72} & \underline{0.74} & 0.70             & 0.74             & \textbf{0.80}    & \textbf{0.80}    \\
			\midrule
			\multirow[c]{3}{*}{ComGenVid (ours)}                                 & Sora \cite{sora2024openai}                              & \underline{0.72}                     & \underline{0.67}                     & 0.53             & 0.55             & 0.58             & 0.59             & 0.68             & 0.65             & 0.68             & 0.65             & \textbf{0.84}    & \textbf{0.85}    \\
			                                                                     & VEO3 \cite{veo32025deepmind}                            & 0.67                                 & 0.62                                 & 0.62             & 0.63             & 0.52             & 0.55             & \underline{0.79} & 0.76             & \underline{0.79} & \underline{0.78} & \textbf{0.86}    & \textbf{0.87}    \\
			\addlinespace
			\midrule
			                                                                     & Average                                                 & 0.69                                 & 0.64                                 & 0.57             & 0.59             & 0.55             & 0.57             & \underline{0.73} & \underline{0.71} & \underline{0.73} & \underline{0.71} & \textbf{0.85}    & \textbf{0.86}    \\
			\midrule
			\midrule
			All Benchmarks                                                       & Average                                                 & 0.62                                 & 0.61                                 & 0.61             & 0.59             & 0.57             & 0.58             & \underline{0.64} & \underline{0.65} & \underline{0.64} & \underline{0.65} & \textbf{0.82}    & \textbf{0.82}    \\
			\bottomrule
		\end{tabular}
	\end{adjustbox}
\end{table*}

\begin{figure}[t]
	\centering
	\includegraphics[width=0.4\textwidth]{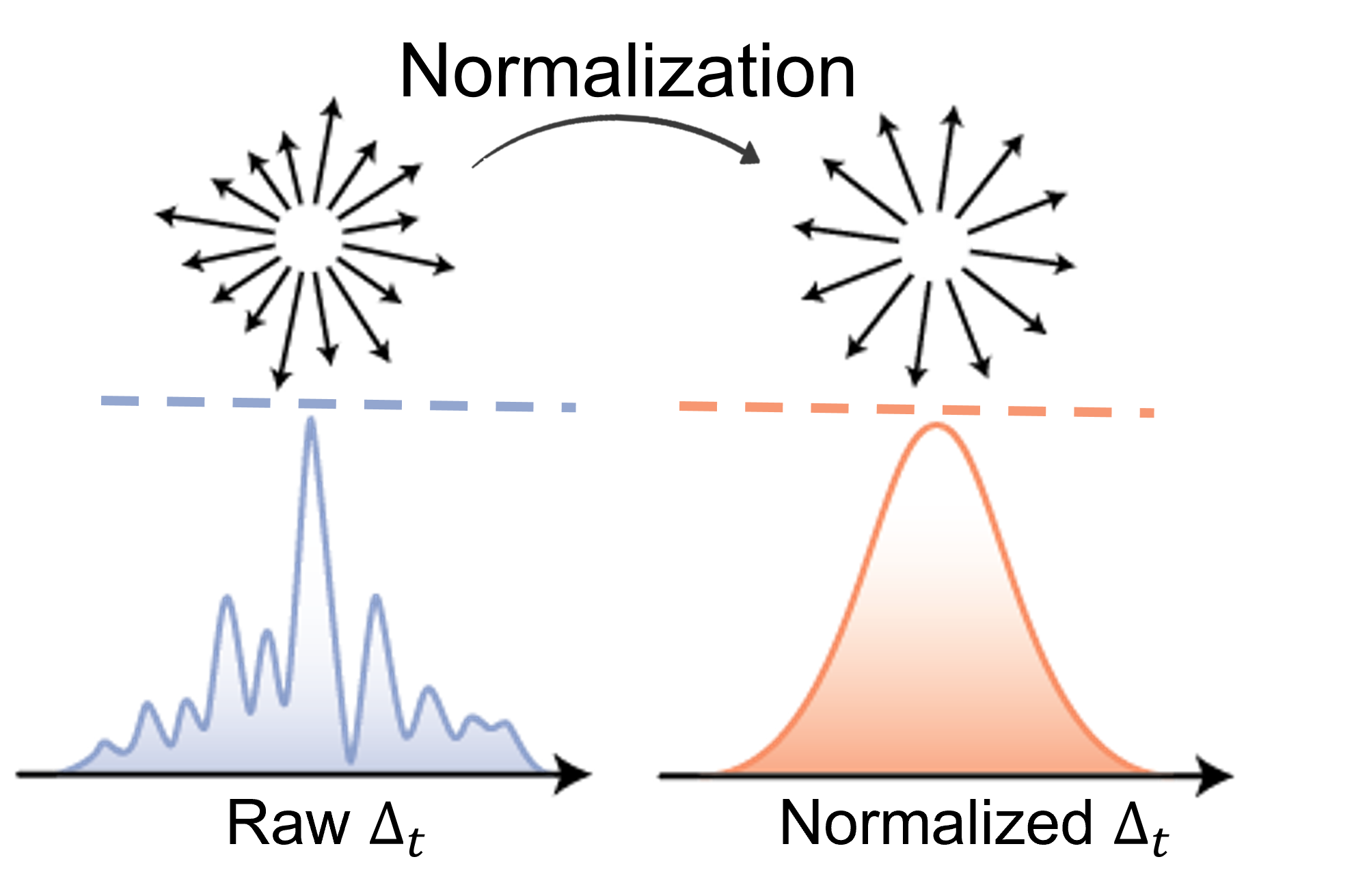}
	\caption{\textbf{Temporal embedding coordinates.}
		Raw coordinates of temporal embeddings (frame differences) are not Gaussian; after normalization, each coordinate is approximately Gaussian (full histogram comparison in Supp. B.2.1).}

	\label{fig:temp_norm}
\end{figure}

\subsection{Unified score}
We compute likelihood scores for each frame (spatial) and each frame-to-frame transition (temporal).
We first aggregate each list separately and then combine the two aggregates into a single video-level score.
We evaluate standard aggregation operators: minimum, maximum, and mean, on a set of real videos and measure the cross-domain correlations induced by each choice (\Cref{fig:agg_metrics_corr}).
Combining the minimum of one domain with the maximum of the other yields the lowest correlation, indicating complementary information.
Accordingly, we use the minimal temporal likelihood and the maximal spatial likelihood per video.
The method is robust to this selection; detection results for all combinations are reported in Supp.\ Section~D.2.

\noindent \textbf{Percentile scoring.}
Because spatial and temporal likelihoods lie on different scales, we avoid raw magnitudes and compare each score relative to real data, so decisions reflect how typical a video is under the calibration distribution.
We set aside the spatial and temporal scores from the calibration set and, at inference, convert a test score $s$ into a \emph{rank-based percentile} by counting how many calibration scores $s_1,\ldots,s_n$ satisfy $s_i \le s$ and dividing by $n$: $\mathrm{perc}(s)
	\;=\;
	\frac{1}{n}\,
	\bigl|\{\, i : s_i \le s \,\}\bigr|.$
We compute these percentiles separately for the spatial and temporal scores.


\noindent \textbf{Unified video score.}
The final video score aggregates the percentile-normalized components:
\begin{equation}
	s_{\mathrm{video}}(v) \;=\; \frac{1}{2}\Big(
	\text{perc}_{\mathrm{sp}}(v) + \text{perc}_{\mathrm{temp}}(v)
	\Big).
	\label{eq:unified_score}
\end{equation}
Percentile normalization makes both terms scale-free and less sensitive to extreme OOD values.
In ~\Cref{sec:ablations}, we ablate each component (spatial/temporal) alone and cross-component fusion (average vs.\ product) and find robustness across choices.
Each component is individually discriminative, and the unified score performs best.

\subsection{Calibration set}
\label{sec:calibration}
We use a \emph{calibration set} of real videos to compute whitening statistics and percentile ranges, aligning with zero-shot detection: no generated samples are used at any point, and ``in-distribution'' is defined solely by real data.
The calibration set is \emph{disjoint} from all evaluation benchmarks and any other data used elsewhere in this paper, ensuring no overlap or leakage.
This is not a limitation: every detector must define a decision boundary, and real-only calibration provides a principled, data-driven anchor for both spatial and temporal likelihoods.
Ablations are provided in Sec.~\ref{sec:ablations}.

\begin{figure*}[t]
	\centering

	\begin{subfigure}{0.3\textwidth}
		\centering
		\includegraphics[width=\linewidth]{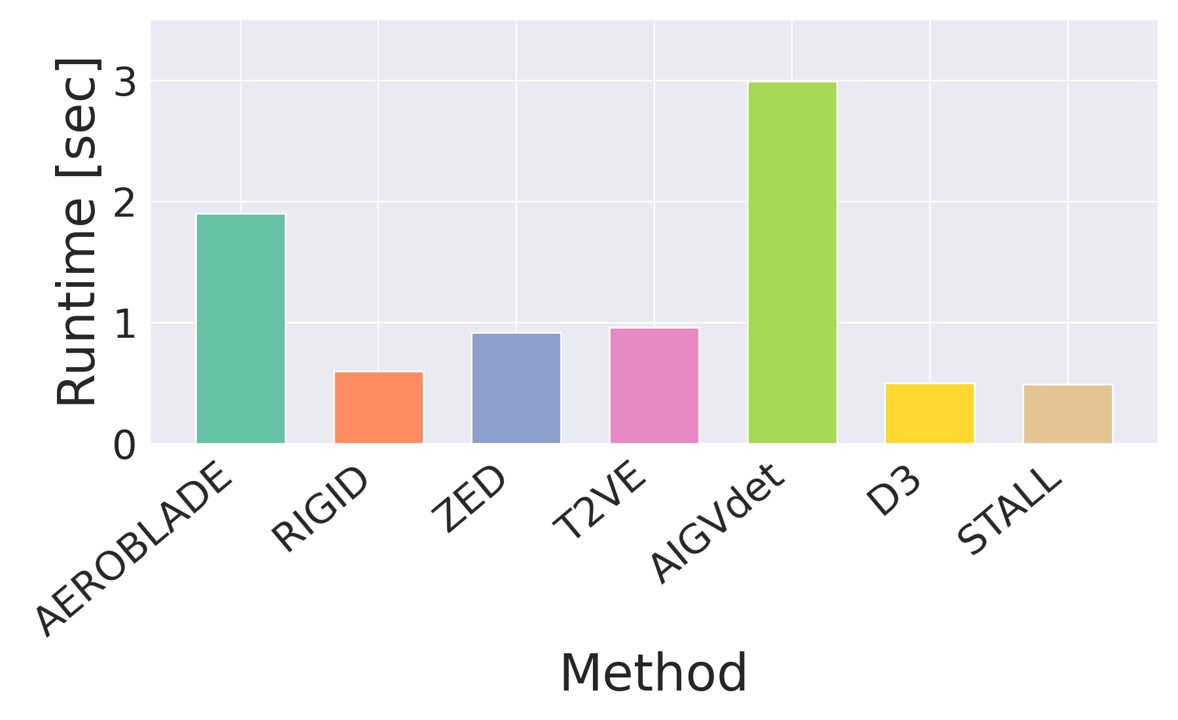}
		\caption{Latency comparison.}
		\label{fig:inference}
	\end{subfigure}\hfill
	\begin{subfigure}{0.3\textwidth}
		\centering
		\includegraphics[width=\linewidth]{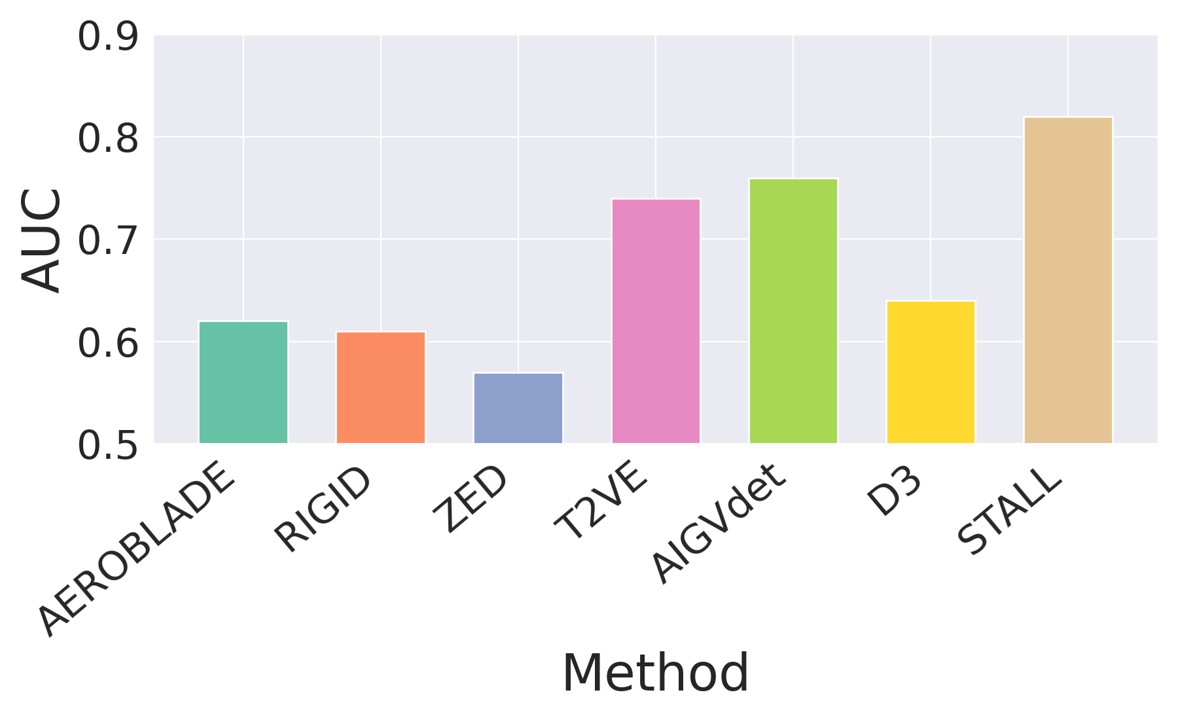}
		\caption{Performance comparison.}
		\label{fig:auc_all}
	\end{subfigure}\hfill
	\begin{subfigure}{0.3\textwidth}
		\centering
		\includegraphics[width=\linewidth]{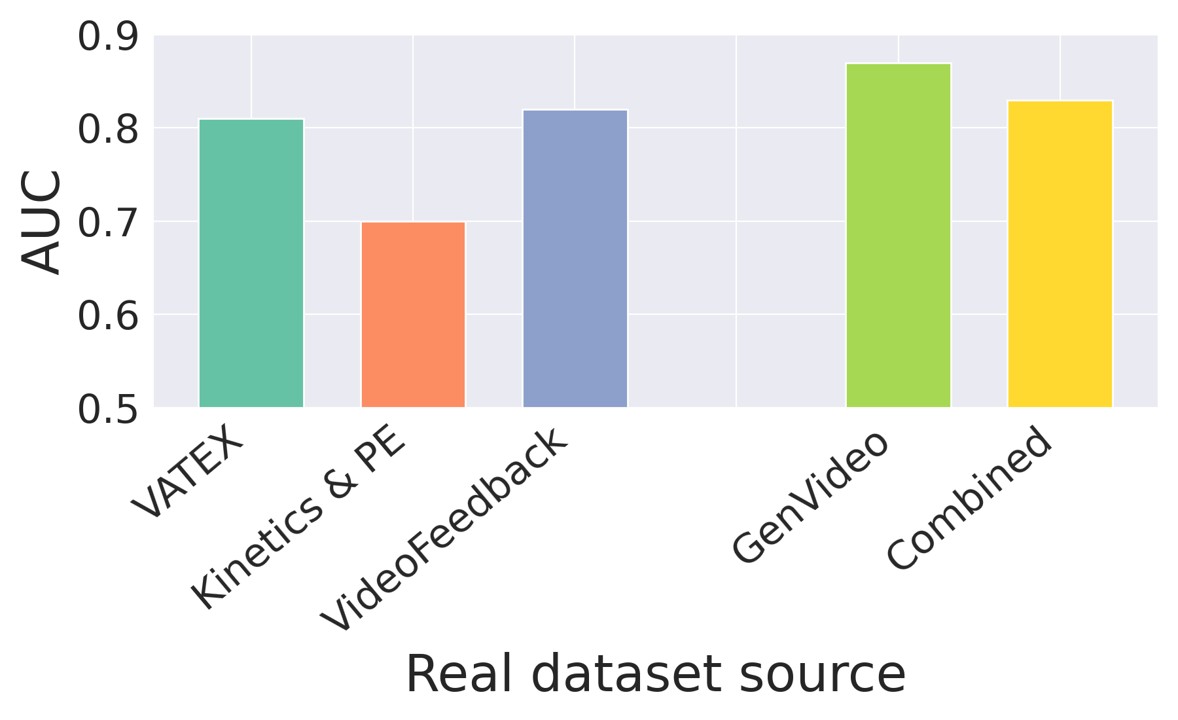}
		\caption{Calibration set ablation.}
		\label{fig:cal_data}
	\end{subfigure}

	\caption{\textbf{Comparison of detectors and calibration data.}
		We compare our method against six detectors: three image-based~\citep{ricker2024aeroblade,he2024rigid,cozzolino2024zero} and three video-based~\citep{zheng2025d3,T2VE_2025,bai2024ai}, for performance and efficiency.
		(a) Inference latency per video.
		(b) Average AUC across all three benchmarks; our method is both high-performing and efficient.
		(c) GenVideo~\citep{chen2024demamba} results using different datasets as the calibration set, showing that same-distribution calibration is only slightly better, indicating robustness to the calibration choice (see Supp. D.6.6).}
	\label{fig:method_comp}
\end{figure*}

\section{Evaluations}
\label{sec:eval}
\subsection{Experimental settings}
\label{sec:eval_dets}
\noindent \textbf{Datasets.}
We evaluate our detector on two benchmarks spanning real and generated videos.
\emph{VideoFeedback}~\citep{he2024videoscore} contains $\sim$33k generated videos from 11 text-to-video models~\citep{pika2023, khachatryan2023text2video, chen2024videocrafter2, wang2023modelscope, wang2025lavie, guo2023animatediff, he2022latent, hotshotxl2023, zeroscope_v2_2024, blattmann2023stable, brooks2024video} and $\sim$4k real videos drawn from two datasets~\citep{chen2024panda, anne2017localizing}.
\emph{GenVideo}~\citep{chen2024demamba} (test set) comprises $\sim$8.5k generated videos from 10 generative sets~\citep{wang2023modelscope, morphstudio2023, moonvalley2022, hotshot2023, zhang2025show, esser2023structure, chen2023videocrafter1, wang2025lavie, brooks2024video} and $\sim$10k real videos from a single dataset~\citep{xu2016msr}.
Across both benchmarks, the generative models constitute a diverse collection of diffusion-based text-to-video systems.
Additionally, we present \emph{ComGenVid}, a set of $\sim$3.5k generated videos from recent commercial models Veo3 and Sora~\citep{veo32025deepmind, sora2024openai}, designed to stress cross-model generalization.
We pair these with $\sim$1.7k real videos sampled from~\cite{chen2011collecting}.
For all evaluations, we subsample to use equal numbers of real and generated videos (determined by the smaller class in each split) to ensure fair metric comparisons. A complete breakdown of generative models, video counts, and dataset composition is given in Supp. Section C.

\begin{figure}[t]
	\centering

	\begin{subfigure}{0.23\textwidth}
		\centering
		\includegraphics[width=\linewidth]{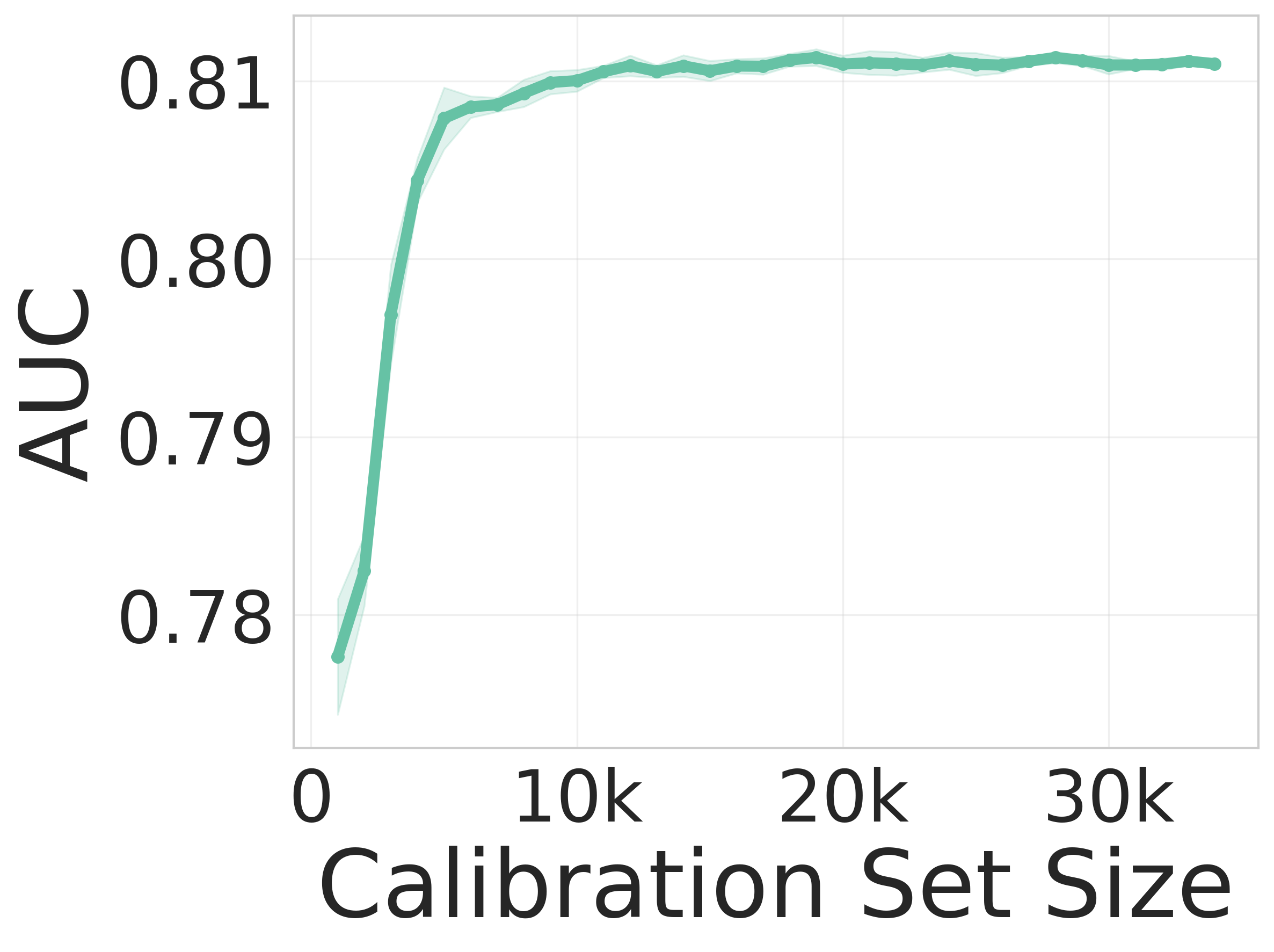}
		\caption{Calibration set size ablation.}
		\label{fig:cal_size}
	\end{subfigure}\hfill
	\begin{subfigure}{0.23\textwidth}
		\centering
		\includegraphics[width=\linewidth]{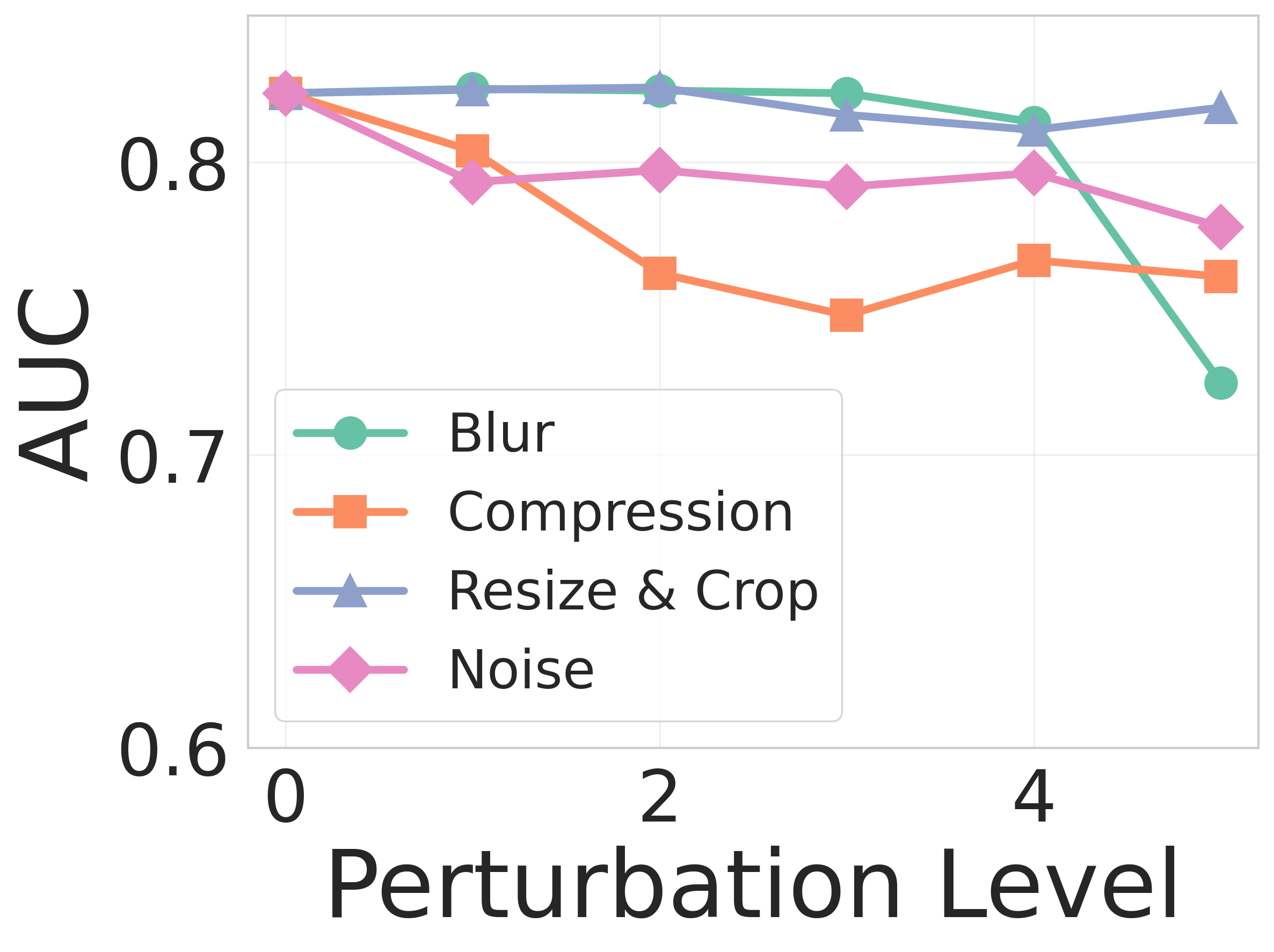}
		\caption{Image perturbations.}
		\label{fig:image_pert}
	\end{subfigure}\hfill
	\caption{\textbf{Robustness to calibration set size and image perturbations.}
		(a) Varying calibration-set size between 1k to 34k; each size is resampled 5 times and we report mean AUC $\pm$ standard deviation.
		(b) Robustness to four common image perturbations applied randomly to frames at five severity levels; our method maintains high separation across perturbation type and intensity. Both experiments are performed on GenVideo~\citep{chen2024demamba}.}

	\label{fig:cal_noise}
\end{figure}

\noindent \textbf{Metrics.}
We report Area Under the ROC Curve (AUC) and Average Precision (AP).
AUC measures the ability of the detector to separate real and generated videos by integrating the ROC curve (true-positive-rate vs.\ false-positive-rate across thresholds), while AP summarizes the precision–recall trade-off for the positive (generated) class.

\noindent \textbf{Implementation details.}
We use available official implementations for baselines: \emph{AEROBLADE}~\citep{ricker2024aeroblade} and \emph{D3} (both L2 and cosine similarity variants, see Supp. Section A.4), and the supervised detectors T2VE~\citep{T2VE_2025} and AIGVdet~\citep{bai2024ai} (official weights and code).
For \emph{RIGID}~\citep{he2024rigid} and \emph{ZED}~\citep{cozzolino2024zero}, we reimplemented the authors’ methods following the paper’s specifications (see Supp. Section A.2).
Image detectors operate per-frame, and we report the mean score over frames.
In all experiments we encode frames using DINOv3~\citep{siméoni2025dinov3} for our method, and use a fixed calibration set that is built from 33k real videos from VATEX~\citep{wang2019vatex}. This dataset is completely separate from any data used for evaluations.
We conduct ablations on calibration set size and dataset, encoder model and method components in the next section.

\noindent \textbf{Data curation and evaluation protocol.}
Following standard protocols~\cite{bai2024ai, zheng2025d3},
we standardize inputs to 8 or 16 frames.
For fair comparison, we sample all evaluated videos at 8\,FPS and 2\,s duration (16 frames).
The only exceptions are HotShot-XL and MoonValley~\citep{hotshotxl2023,moonvalley2022}, which generate 1\,s videos; for these we compare against real 1\,s videos at 8\,FPS (8 frames).
Image detectors operate per frame and the average score over all frames is evaluated.
We report results under this default setting (Supp.~A.3) and provide an ablation on FPS/duration sensitivity in the next section.

\begin{figure*}[t]
	\centering

	\begin{subfigure}{0.28\textwidth}
		\centering
		\includegraphics[width=\linewidth]{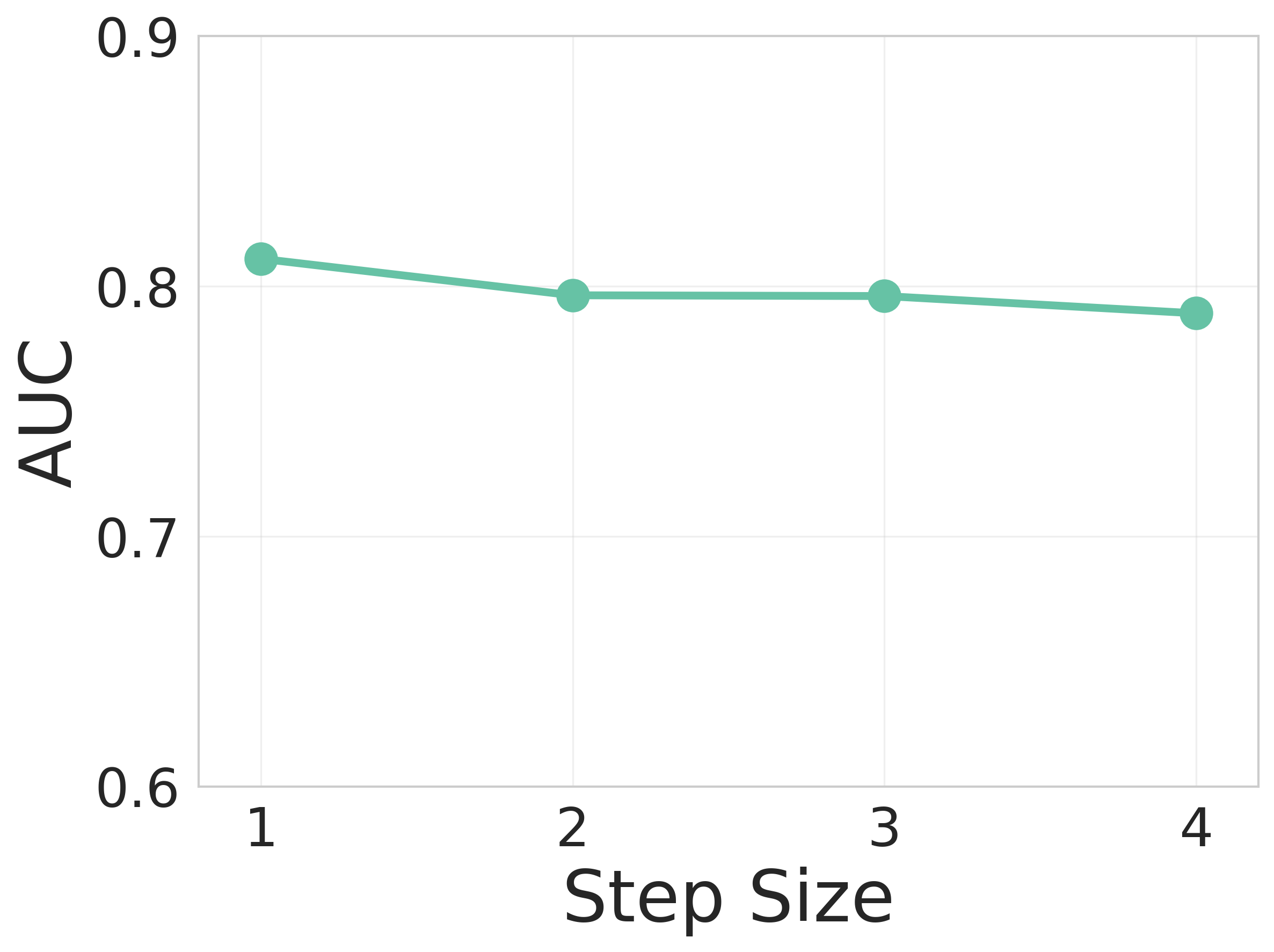}
		\caption{Step size ablation.}
		\label{fig:step_size}
	\end{subfigure}\hfill
	\begin{subfigure}{0.28\textwidth}
		\centering
		\includegraphics[width=\linewidth]{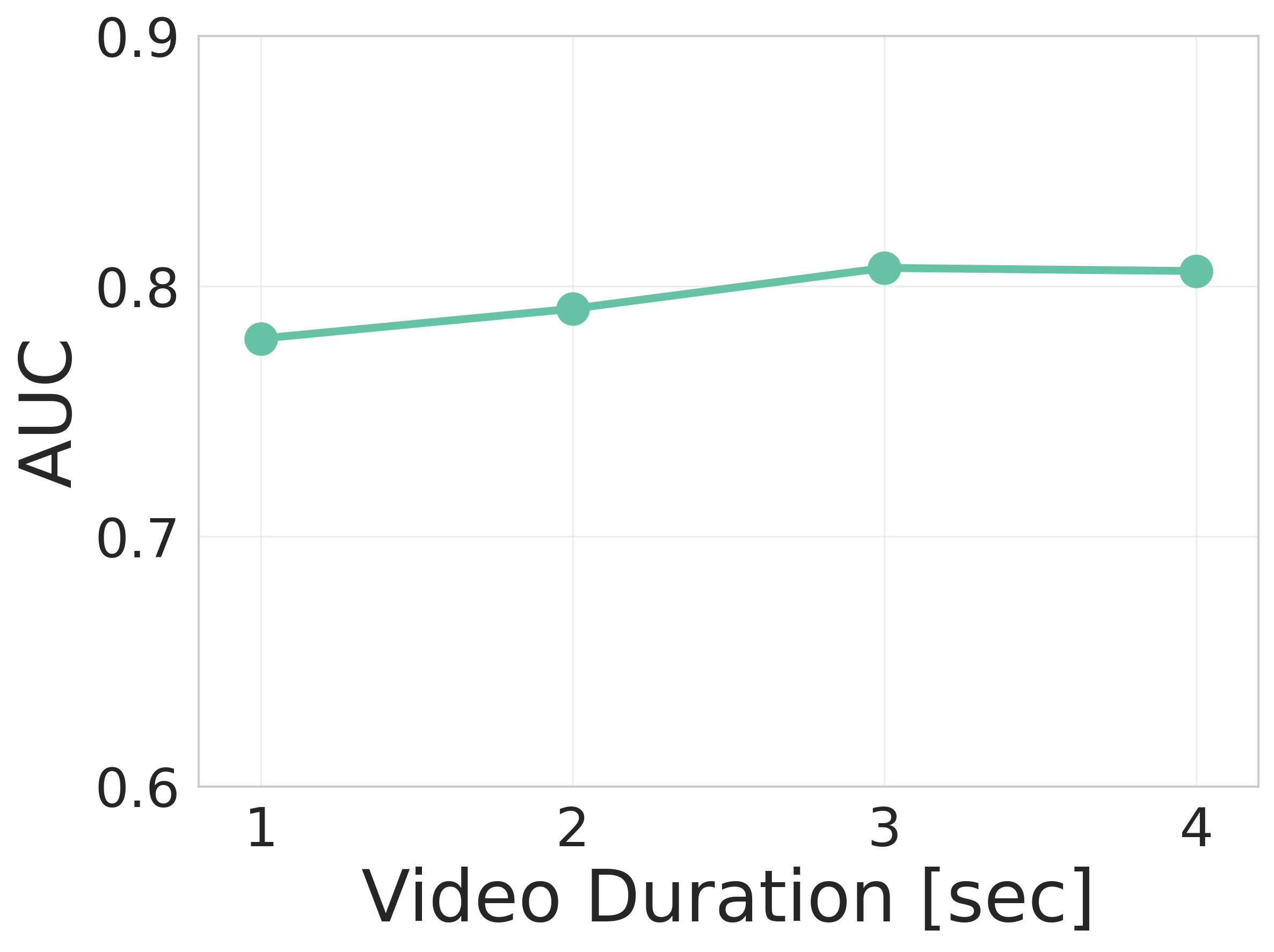}
		\caption{Length of video ablation.}
		\label{fig:vid_len}
	\end{subfigure}\hfill
	\begin{subfigure}{0.28\textwidth}
		\centering
		\includegraphics[width=\linewidth]{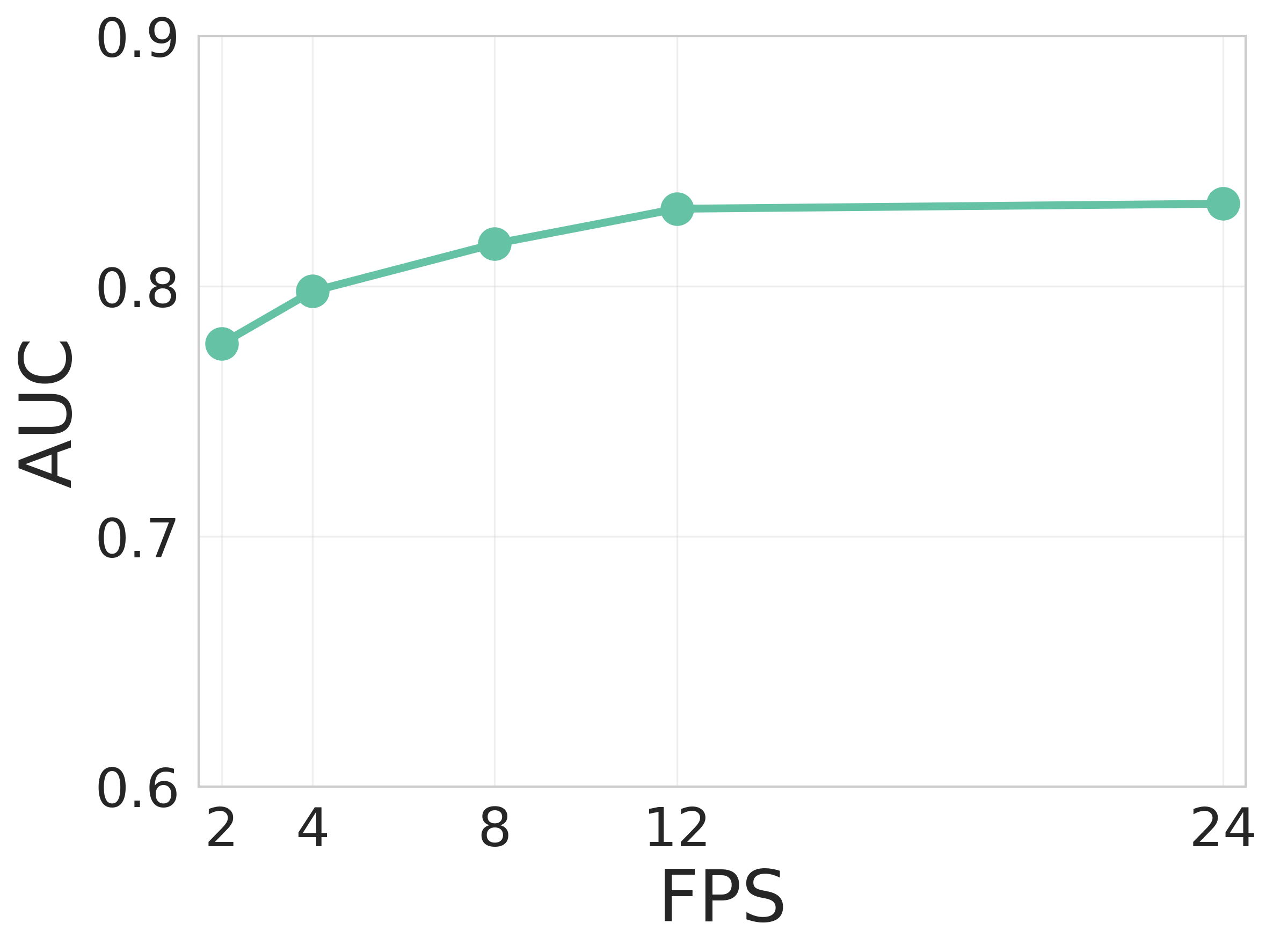}
		\caption{FPS ablation.}
		\label{fig:fps}
	\end{subfigure}

	\caption{\textbf{Temporal ablations.}
		Our method remains robust under all temporal variations:
		(a) temporal step size for likelihood computation;
		(b) video length;
		(c) frame rate (FPS).}

	\label{fig:temp_ablations}
\end{figure*}

\subsection{Results}
\label{sec:eval_results}

\noindent \textbf{Benchmark evaluations.}
\Cref{tab:zero-shot_results} reports zero-shot results across all three benchmarks.
Our method achieves the highest average performance on each benchmark and attains the best per-generator results in most cases; when not the top method, it remains competitive.
Notably, all other methods produce AUC values below 0.5 for some generators, indicating an inverted decision boundary; detectors that fit one model misclassify many examples from others.
Our method does not exhibit this failure mode and maintains consistent separation between real and generated samples.
In \Cref{fig:auc_all} we also include supervised video detectors; our zero-shot method still outperforms them, even though they are partially trained on the evaluated generators.

\noindent \textbf{Efficiency.}
We measure each method's inference time, measured per video (16 frames input), results are reported in \Cref{fig:inference}. STALL, together with D3~\citep{zheng2025d3} and RIGID~\citep{he2024rigid} are the fastest methods with 0.49s, 0.5s and 0.6s respectively. ZED~\citep{cozzolino2024zero} and T2VE~\citep{T2VE_2025} show double latency (0.92s, 0.97s) while AEROBLADE~\citep{ricker2024aeroblade} and AIGVdet~\citep{bai2024ai} demonstrate increased latency. Our method is relatively lightweight (Supp. Section E) making it highly efficient.

\subsection{Ablation study}
\label{sec:ablations}
\noindent \textbf{Calibration set.}
We study the effect of calibration set size and dataset.
We examine different datasets as the source of the calibration set in ~\Cref{fig:cal_data}. We compare VATEX~\citep{wang2019vatex} with a combination of Kinetics-400 and PE datasets~\cite{kay2017kinetics, bolya2025perception}, samples from the real data of VideoFeedback and samples from the real data of Genvideo (different samples than the evaluation set). We add a combination of all four options. Using data from the same distribution as tested (GenVideo) results in only a slightly higher performance. Other options remain competitive, demonstrating robustness to dataset selection.
We then vary the calibration set size (using VATEX~\citep{wang2019vatex}) between 1k and 34k and report mean AUC and standard deviation over 5 sampling iterations. Results are in ~\Cref{fig:cal_size} showing that only at very small sizes (less than 5k) results drop significantly.
For whitening, we use one frame per video and all frame-to-frame transitions.
We test using a single transition per video and find that results remain identical (Supp.\ Section~D.3).

\noindent \textbf{Embedder comparison.}
We evaluate our detector with multiple vision encoders: DINOv3~\citep{siméoni2025dinov3}, lightweight MobileNet~\citep{howard2019searching}, and ResNet-18~\citep{he2016deep}, as well as video encoders VideoMAE~\citep{tong2022videomae} and ViCLIP~\citep{wang2023internvid}.
Video encoders produce a single embedding per \emph{video}, and we compute likelihood directly on this vector.
\Cref{tab:embedder_results} shows that image encoders perform strongly, even with older, lightweight backbones like MobileNet and ResNet-18. Video encoders, by contrast, perform poorly.
Collapsing an entire video to a single embedding discards frame-wise and transition statistics, undermining both spatial and temporal likelihood modeling.

\begin{table}[htbp]
	\caption{\textbf{Backbone encoder ablation.}}
	\label{tab:embedder_results}
	\begin{adjustbox}{max width=0.48\textwidth}
		\begin{tabular}{@{} | *{1}{c} | *{3}{c} | *{2}{c}  @{} | }
			\toprule
			    & \multicolumn{3}{c|}{Image encoders} & \multicolumn{2}{c|}{Video encoders}                                                                                                                 \\
			    & DINOv3~\cite{siméoni2025dinov3}     & MobileNet-v3~\citep{howard2019searching} & ResNet-18~\citep{he2016deep} & ViCLIP-L/14~\citep{wang2023internvid} & VideoMAE~\citep{tong2022videomae} \\
			\midrule
			AUC & 0.81                                & 0.82                                     & 0.79                         & 0.59                                  & 0.61                              \\
			\bottomrule
		\end{tabular}
	\end{adjustbox}
\end{table}

\noindent \textbf{Robustness analysis.}
We test robustness by applying standard image corruptions to video frames: JPEG compression, Gaussian blur, resized crop, and additive noise, at five severity levels. Perturbations are applied only at inference while keeping the calibration set unchanged.
Results in \Cref{fig:image_pert} show strong robustness across perturbation types up to the highest intensity levels;  implementation details and examples in Supp.\ Section~D.5.
We also evaluate robustness to temporal perturbations in Supp.\ Section~D.4.

\noindent We further vary the input FPS and the temporal step size between frames ($\Delta_t$) to assess sensitivity to motion sparsity and sampling rate, and also test different video durations.
Results in \Cref{fig:temp_ablations} show that our method remains robust across all temporal settings. Evaluations settings of these experiments are in Supp.\ Section~D.6.

\noindent Finally, we assess performance using higher-order temporal differences: while temporal transitions capture first-order changes, higher orders model more complex motion dynamics.
As shown in Supp.\ Section~D.1, all orders exhibit high correlation and yield nearly identical results.

\noindent \textbf{Component analysis.}
We assess three variants: (i) spatial-only, (ii) temporal-only, and (iii) the full model combining both.
For each, we report results with raw likelihoods and with percentile-ranked scores. We also test standard aggregations (min, max, mean).
Results show that either single-domain detector performs well, the combined detector performs best, and performance is robust to the choice of aggregation (see Supp. Section D.2).

\section{Conclusion}
We introduce STALL, a zero-shot detector for fully generated videos that fuses spatial (per-frame) and temporal (inter-frame) likelihoods in a single probabilistic framework.
Our method is training-free, uses no generated samples, and relies solely on real videos to define reference distributions for both spatial and temporal statistics.
Across multiple benchmarks, including recent frontier models such as Sora and Veo3, our approach consistently outperforms prior supervised and zero-shot image/video detectors.
It is also efficient and robust to spatial and temporal perturbations, calibration data size and source, and aggregation choices.
As this field continues to develop rapidly, there remains room for improvement; nevertheless, our results highlight that modeling the statistical structure of real videos is a promising path for robust detection.

\section*{Acknowledgments}
We would like to acknowledge
support by the Israel Science Foundation (Grant 1472/23) and by the Ministry of Innovation, Science and Technology (Grant 8801/25).

{
    \small
    \bibliographystyle{ieeenat_fullname}
    \bibliography{main}

@String(ICLR = {Int. Conf. Learn. Represent.})

@String(ICLR  = {ICLR})

@article{ho2020denoising,
  title={Denoising diffusion probabilistic models},
  author={Ho, Jonathan and Jain, Ajay and Abbeel, Pieter},
  journal={Advances in neural information processing systems},
  volume={33},
  pages={6840--6851},
  year={2020}
}

@article{song2019generative,
  title={Generative modeling by estimating gradients of the data distribution},
  author={Song, Yang and Ermon, Stefano},
  journal={Advances in neural information processing systems},
  volume={32},
  year={2019}
}

@inproceedings{rombach2022high,
  title={High-resolution image synthesis with latent diffusion models},
  author={Rombach, Robin and Blattmann, Andreas and Lorenz, Dominik and Esser, Patrick and Ommer, Bj{\"o}rn},
  booktitle={Proceedings of the IEEE/CVF conference on computer vision and pattern recognition},
  pages={10684--10695},
  year={2022}
}

@article{raffel2020exploring,
  title={Exploring the limits of transfer learning with a unified text-to-text transformer},
  author={Raffel, Colin and Shazeer, Noam and Roberts, Adam and Lee, Katherine and Narang, Sharan and Matena, Michael and Zhou, Yanqi and Li, Wei and Liu, Peter J},
  journal={Journal of machine learning research},
  volume={21},
  number={140},
  pages={1--67},
  year={2020}
}

@inproceedings{brokman2026training,
  title={Training-free Detection of Text-to-video Generations via Over-coherence},
  author={Brokman, Jonathan and Rachmil, Oren and Hofman, Omer and Betser, Roy and Giloni, Amit and Vainshtein, Roman and Kojima, Hisashi},
  booktitle={Proceedings of the IEEE/CVF Winter Conference on Applications of Computer Vision},
  pages={3993--4003},
  year={2026}
}

@article{rachmil2025training,
  title={Training-Free Policy Violation Detection via Activation-Space Whitening in LLMs},
  author={Rachmil, Oren and Betser, Roy and Gershon, Itay and Hofman, Omer and Yakoby, Nitay and Meron, Yuval and Yankelev, Idan and Shabtai, Asaf and Elovici, Yuval and Vainshtein, Roman},
  journal={arXiv preprint arXiv:2512.03994},
  year={2025}
}

@article{brown2020language,
  title={Language models are few-shot learners},
  author={Brown, Tom and Mann, Benjamin and Ryder, Nick and Subbiah, Melanie and Kaplan, Jared D and Dhariwal, Prafulla and Neelakantan, Arvind and Shyam, Pranav and Sastry, Girish and Askell, Amanda and others},
  journal={Advances in neural information processing systems},
  volume={33},
  pages={1877--1901},
  year={2020}
}

@inproceedings{chen2024videocrafter2,
  title={Videocrafter2: Overcoming data limitations for high-quality video diffusion models},
  author={Chen, Haoxin and Zhang, Yong and Cun, Xiaodong and Xia, Menghan and Wang, Xintao and Weng, Chao and Shan, Ying},
  booktitle={Proceedings of the IEEE/CVF Conference on Computer Vision and Pattern Recognition},
  pages={7310--7320},
  year={2024}
}

@article{hacohen2024ltx,
  title={Ltx-video: Realtime video latent diffusion},
  author={HaCohen, Yoav and Chiprut, Nisan and Brazowski, Benny and Shalem, Daniel and Moshe, Dudu and Richardson, Eitan and Levin, Eran and Shiran, Guy and Zabari, Nir and Gordon, Ori and others},
  journal={arXiv preprint arXiv:2501.00103},
  year={2024}
}

@article{wan2025wan,
  title={Wan: Open and advanced large-scale video generative models},
  author={Wan, Team and Wang, Ang and Ai, Baole and Wen, Bin and Mao, Chaojie and Xie, Chen-Wei and Chen, Di and Yu, Feiwu and Zhao, Haiming and Yang, Jianxiao and others},
  journal={arXiv preprint arXiv:2503.20314},
  year={2025}
}

@misc{using2023,
  title        = {Using AI to Simplify Content Marketing Workflows},
  author       = {Stelzner, Michael},
  howpublished = {Social Media Examiner, “Using AI to Simplify Content Marketing Workflows”},
  url          = {https://www.socialmediaexaminer.com/using-ai-to-simplify-content-marketing-workflows/},
  year         = {2023}
}

@misc{times2025,
  title        = {AI-assisted content creation will lower the barrier to creativity but raise quality: Adobe’s Govind Balakrishnan},
  author       = {TOI Tech Desk},
  howpublished = {The Times of India, “AI-assisted content creation will lower the barrier to creativity…”},
  url          = {https://timesofindia.indiatimes.com/technology/tech-news/ai-assisted-content-creation-will-lower-the-barrier-to-creativity-but-raises-quality-adobes-govind-balakrishnan/articleshow/124922128.cms},
  year         = {2025}
}

@misc{appel2023generative,
  title        = {Generative AI Has an Intellectual Property Problem},
  author       = {Appel, Gil and Neelbauer, Juliana and Schweidel, David A.},
  howpublished = {Harvard Business Review, April 7, 2023},
  url          = {https://hbr.org/2023/04/generative-ai-has-an-intellectual-property-problem},
  year         = {2023}
}

@misc{dandodiary2025deepfake,
  title        = {The Growing Threat of AI Deepfake Attacks},
  author       = {LaCroix, Kevin},
  howpublished = {Dando and O’Malley, August 19, 2025},
  url          = {https://www.dandodiary.com/2025/08/articles/cyber-liability/the-growing-threat-of-ai-deepfake-attacks/},
  year         = {2025}
}

@misc{euiphelpdesk2024deepfake,
  title        = {Deepfake – A Global Crisis},
  author       = {European Union Intellectual Property Helpdesk},
  howpublished = {EUIPO, August 28, 2024},
  url          = {https://intellectual-property-helpdesk.ec.europa.eu/news-events/news/deepfake-global-crisis-2024-08-28_en},
  year         = {2024}
}

@article{kalra2025india,
  title        = {India proposes strict rules to label AI content citing growing risks of deepfakes},
  author       = {Kalra, Aditya and Vengattil, Munsif},
  journal      = {Reuters},
  month        = {Oct},
  day          = {22},
  year         = {2025},
  url          = {https://www.reuters.com/business/media-telecom/india-proposes-strict-it-rules-labelling-deepfakes-amid-ai-misuse-2025-10-22/},
}

@misc{itu2025deepfakes,
  title        = {UN report urges stronger measures to detect AI-driven deepfakes},
  author       = {Le Poidevin, Olivia },
  howpublished = {Reuters, July 11, 2025},
  url          = {https://www.reuters.com/business/un-report-urges-stronger-measures-detect-ai-driven-deepfakes-2025-07-11/},
  year         = {2025}
}

@article{baraheem2023ai,
  title={AI vs. AI: Can AI Detect AI-Generated Images?},
  author={Baraheem, Samah S and Nguyen, Tam V},
  journal={Journal of Imaging},
  volume={9},
  number={10},
  pages={199},
  year={2023},
  publisher={MDPI}
}

@article{bird2024cifake,
  title={Cifake: Image classification and explainable identification of ai-generated synthetic images},
  author={Bird, Jordan J and Lotfi, Ahmad},
  journal={IEEE Access},
  year={2024},
  publisher={IEEE}
}

@article{cioni2024clip,
  title={Are CLIP features all you need for Universal Synthetic Image Origin Attribution?},
  author={Cioni, Dario and Tzelepis, Christos and Seidenari, Lorenzo and Patras, Ioannis},
  journal={arXiv preprint arXiv:2408.09153},
  year={2024}
}

@inproceedings{epstein2023online,
  title={Online detection of ai-generated images},
  author={Epstein, David C and Jain, Ishan and Wang, Oliver and Zhang, Richard},
  booktitle={Proceedings of the IEEE/CVF International Conference on Computer Vision},
  pages={382--392},
  year={2023}
}

@inproceedings{ojha2023towards,
  title={Towards universal fake image detectors that generalize across generative models},
  author={Ojha, Utkarsh and Li, Yuheng and Lee, Yong Jae},
  booktitle={Proceedings of the IEEE/CVF Conference on Computer Vision and Pattern Recognition},
  pages={24480--24489},
  year={2023}
}

@inproceedings{cozzolino2023raising,
  title={Raising the Bar of AI-generated Image Detection with CLIP},
  author={Cozzolino, Davide and Poggi, Giovanni and Corvi, Riccardo and Nie{\ss}ner, Matthias and Verdoliva, Luisa},
  booktitle={Proceedings of the IEEE/CVF Conference on Computer Vision and Pattern Recognition},
  pages={4356--4366},
  year={2024}
}

@inproceedings{ricker2024aeroblade,
  title={AEROBLADE: Training-Free Detection of Latent Diffusion Images Using Autoencoder Reconstruction Error},
  author={Ricker, Jonas and Lukovnikov, Denis and Fischer, Asja},
  booktitle={Proceedings of the IEEE/CVF Conference on Computer Vision and Pattern Recognition},
  pages={9130--9140},
  year={2024}
}

@article{he2024rigid,
  title={RIGID: A Training-free and Model-Agnostic Framework for Robust AI-Generated Image Detection},
  author={He, Zhiyuan and Chen, Pin-Yu and Ho, Tsung-Yi},
  journal={arXiv preprint arXiv:2405.20112},
  year={2024}
}

@inproceedings{cozzolino2024zero,
  title={Zero-shot detection of ai-generated images},
  author={Cozzolino, Davide and Poggi, Giovanni and Nie{\ss}ner, Matthias and Verdoliva, Luisa},
  booktitle={European Conference on Computer Vision},
  pages={54--72},
  year={2024},
  organization={Springer}
}

@inproceedings{brokman2025manifold,
  title={Manifold Induced Biases for Zero-shot and Few-shot Detection of Generated Images},
  author={Jonathan Brokman and Amit Giloni and Omer Hofman and Roman Vainshtein and Hisashi Kojima and Guy Gilboa},
  booktitle={International Conference on Learning Representations},
  year={2025},
  url={https://openreview.net/forum?id=7gGl6HB5Zd}
}

@inproceedings{vahdati2024beyond,
  title={Beyond deepfake images: Detecting ai-generated videos},
  author={Vahdati, Danial Samadi and Nguyen, Tai D and Azizpour, Aref and Stamm, Matthew C},
  booktitle={Proceedings of the IEEE/CVF Conference on Computer Vision and Pattern Recognition},
  pages={4397--4408},
  year={2024}
}

@inproceedings{bai2024ai,
  title={Ai-generated video detection via spatial-temporal anomaly learning},
  author={Bai, Jianfa and Lin, Man and Cao, Gang and Lou, Zijie},
  booktitle={Chinese Conference on Pattern Recognition and Computer Vision (PRCV)},
  pages={460--470},
  year={2024},
  organization={Springer}
}

@inproceedings{wang2019vatex,
  title={Vatex: A large-scale, high-quality multilingual dataset for video-and-language research},
  author={Wang, Xin and Wu, Jiawei and Chen, Junkun and Li, Lei and Wang, Yuan-Fang and Wang, William Yang},
  booktitle={Proceedings of the IEEE/CVF international conference on computer vision},
  pages={4581--4591},
  year={2019}
}

@inproceedings{xu2016msr,
  title={Msr-vtt: A large video description dataset for bridging video and language},
  author={Xu, Jun and Mei, Tao and Yao, Ting and Rui, Yong},
  booktitle={Proceedings of the IEEE conference on computer vision and pattern recognition},
  pages={5288--5296},
  year={2016}
}

@inproceedings{zeng2024benchmarking,
  title={Benchmarking the Robustness of Temporal Action Detection Models Against Temporal Corruptions},
  author={Zeng, Runhao and Chen, Xiaoyong and Liang, Jiaming and Wu, Huisi and Cao, Guangzhong and Guo, Yong},
  booktitle={IEEE Conference on Computer Vision and Pattern Recognition},
  year={2024},
}

@article{wang2023modelscope,
  title={Modelscope text-to-video technical report},
  author={Wang, Jiuniu and Yuan, Hangjie and Chen, Dayou and Zhang, Yingya and Wang, Xiang and Zhang, Shiwei},
  journal={arXiv preprint arXiv:2308.06571},
  year={2023}
}

@article{zhang2025show,
  title={Show-1: Marrying pixel and latent diffusion models for text-to-video generation},
  author={Zhang, David Junhao and Wu, Jay Zhangjie and Liu, Jia-Wei and Zhao, Rui and Ran, Lingmin and Gu, Yuchao and Gao, Difei and Shou, Mike Zheng},
  journal={International Journal of Computer Vision},
  volume={133},
  number={4},
  pages={1879--1893},
  year={2025},
  publisher={Springer}
}

@article{wang2025lavie,
  title={Lavie: High-quality video generation with cascaded latent diffusion models},
  author={Wang, Yaohui and Chen, Xinyuan and Ma, Xin and Zhou, Shangchen and Huang, Ziqi and Wang, Yi and Yang, Ceyuan and He, Yinan and Yu, Jiashuo and Yang, Peiqing and others},
  journal={International Journal of Computer Vision},
  volume={133},
  number={5},
  pages={3059--3078},
  year={2025},
  publisher={Springer}
}

@article{chen2023videocrafter1,
  title={Videocrafter1: Open diffusion models for high-quality video generation},
  author={Chen, Haoxin and Xia, Menghan and He, Yingqing and Zhang, Yong and Cun, Xiaodong and Yang, Shaoshu and Xing, Jinbo and Liu, Yaofang and Chen, Qifeng and Wang, Xintao and others},
  journal={arXiv preprint arXiv:2310.19512},
  year={2023}
}

@article{guo2023animatediff,
  title={Animatediff: Animate your personalized text-to-image diffusion models without specific tuning},
  author={Guo, Yuwei and Yang, Ceyuan and Rao, Anyi and Liang, Zhengyang and Wang, Yaohui and Qiao, Yu and Agrawala, Maneesh and Lin, Dahua and Dai, Bo},
  journal={arXiv preprint arXiv:2307.04725},
  year={2023}
}

@article{he2022latent,
  title={Latent video diffusion models for high-fidelity long video generation},
  author={He, Yingqing and Yang, Tianyu and Zhang, Yong and Shan, Ying and Chen, Qifeng},
  journal={arXiv preprint arXiv:2211.13221},
  year={2022}
}

@article{brooks2024video,
  title={Video generation models as world simulators},
  author={Brooks, Tim and Peebles, Bill and Holmes, Connor and DePue, Will and Guo, Yufei and Jing, Li and Schnurr, David and Taylor, Joe and Luhman, Troy and Luhman, Eric and others},
  journal={OpenAI Blog},
  volume={1},
  number={8},
  pages={1},
  year={2024}
}

@inproceedings{chen2024panda,
  title={Panda-70m: Captioning 70m videos with multiple cross-modality teachers},
  author={Chen, Tsai-Shien and Siarohin, Aliaksandr and Menapace, Willi and Deyneka, Ekaterina and Chao, Hsiang-wei and Jeon, Byung Eun and Fang, Yuwei and Lee, Hsin-Ying and Ren, Jian and Yang, Ming-Hsuan and others},
  booktitle={Proceedings of the IEEE/CVF Conference on Computer Vision and Pattern Recognition},
  pages={13320--13331},
  year={2024}
}

@inproceedings{betser2026general,
  title={General and Domain-Specific Zero-shot Detection of Generated Images via Conditional Likelihood},
  author={Betser, Roy and Hofman, Omer and Vainshtein, Roman and Gilboa, Guy},
  booktitle={Proceedings of the IEEE/CVF Winter Conference on Applications of Computer Vision},
  pages={7809--7820},
  year={2026}
}

@inproceedings{betser2026infonce,
  title={InfoNCE Induces Gaussian Distribution},
  author={Betser, Roy and Gofer, Eyal and Levi, Meir Yossef and Gilboa, Guy},
  booktitle={International Conference on Learning Representations (ICLR)},
  year={2026},
  eprint={2602.24012},
  archivePrefix={arXiv},
  primaryClass={cs.LG}
}

@inproceedings{esser2023structure,
  title={Structure and content-guided video synthesis with diffusion models},
  author={Esser, Patrick and Chiu, Johnathan and Atighehchian, Parmida and Granskog, Jonathan and Germanidis, Anastasis},
  booktitle={Proceedings of the IEEE/CVF international conference on computer vision},
  pages={7346--7356},
  year={2023}
}

@misc{morphstudio2023,
  title        = {MorphStudio},
  author       = {{MorphStudio}},
  year         = {2023},
  howpublished = {\url{https://www.morphstudio.com/}},
}

@misc{moonvalley2022,
  title        = {MoonValley},
  author       = {{MoonValley}},
  year         = {2022},
  howpublished = {\url{https://moonvalley.ai/}},
}

@misc{hotshot2023,
  title        = {Hotshot{-}XL},
  author       = {{HotshotCo}},
  year         = {2023},
  howpublished = {\url{https://huggingface.co/hotshotco/Hotshot-XL}},
}

@misc{hotshotxl2023,
  title        = {Hotshot{-}XL},
  author       = {HotshotCo},
  year         = {2023},
  howpublished = {\url{https://github.com/hotshotco/hotshot-xl}},
}

@misc{zeroscope_v2_2024,
  title        = {ZeroScope\,V2\,(576\,w)},
  author       = {Cerspense},
  year         = {2024},
  howpublished = {\url{https://huggingface.co/cerspense/zeroscope_v2_576w}},
}

@article{blattmann2023stable,
  title={Stable video diffusion: Scaling latent video diffusion models to large datasets},
  author={Blattmann, Andreas and Dockhorn, Tim and Kulal, Sumith and Mendelevitch, Daniel and Kilian, Maciej and Lorenz, Dominik and Levi, Yam and English, Zion and Voleti, Vikram and Letts, Adam and others},
  journal={arXiv preprint arXiv:2311.15127},
  year={2023}
}

@inproceedings{anne2017localizing,
  title={Localizing moments in video with natural language},
  author={Anne Hendricks, Lisa and Wang, Oliver and Shechtman, Eli and Sivic, Josef and Darrell, Trevor and Russell, Bryan},
  booktitle={Proceedings of the IEEE international conference on computer vision},
  pages={5803--5812},
  year={2017}
}

@misc{pika2023,
  title        = {Pika},
  author       = {{Pika Labs}},
  year         = {2023},
  howpublished = {\url{https://pika.art/}},
}

@inproceedings{khachatryan2023text2video,
  title={Text2video-zero: Text-to-image diffusion models are zero-shot video generators},
  author={Khachatryan, Levon and Movsisyan, Andranik and Tadevosyan, Vahram and Henschel, Roberto and Wang, Zhangyang and Navasardyan, Shant and Shi, Humphrey},
  booktitle={Proceedings of the IEEE/CVF International Conference on Computer Vision},
  pages={15954--15964},
  year={2023}
}

@article{chen2024demamba,
  title={Demamba: Ai-generated video detection on million-scale genvideo benchmark},
  author={Chen, Haoxing and Hong, Yan and Huang, Zizheng and Xu, Zhuoer and Gu, Zhangxuan and Li, Yaohui and Lan, Jun and Zhu, Huijia and Zhang, Jianfu and Wang, Weiqiang and others},
  journal={arXiv preprint arXiv:2405.19707},
  year={2024}
}

@inproceedings{zheng2025d3,
  title={D3: Training-Free AI-Generated Video Detection Using Second-Order Features},
  author={Zheng, Chende and Suo, Ruiqi and Lin, Chenhao and Zhao, Zhengyu and Yang, Le and Liu, Shuai and Yang, Minghui and Wang, Cong and Shen, Chao},
  booktitle={Proceedings of the IEEE/CVF International Conference on Computer Vision},
  pages={12852--12862},
  year={2025}
}

@article{he2024videoscore,
  title = {VideoScore: Building Automatic Metrics to Simulate Fine-grained Human Feedback for Video Generation},
  author = {He, Xuan and Jiang, Dongfu and Zhang, Ge and Ku, Max and Soni, Achint and Siu, Sherman and Chen, Haonan and Chandra, Abhranil and Jiang, Ziyan and Arulraj, Aaran and Wang, Kai and Do, Quy Duc and Ni, Yuansheng and Lyu, Bohan and Narsupalli, Yaswanth and Fan, Rongqi and Lyu, Zhiheng and Lin, Yuchen and Chen, Wenhu},
  journal = {ArXiv},
  year = {2024},
  volume={abs/2406.15252},
  url = {https://arxiv.org/abs/2406.15252},
}

@misc{sora2024openai,
  title        = {Video Generation Models as World Simulators – Introducing Sora},
  author       = {OpenAI},
  howpublished = {Online technical report},
  month        = {Feb},
  year         = {2024},
  url          = {https://openai.com/index/sora/},
}

@inproceedings{betser2025whitened,
  title={Whitened CLIP as a Likelihood Surrogate of Images and Captions},
  author={Betser, Roy and Levi, Meir Yossef and Gilboa, Guy},
  booktitle={42nd International conference on machine learning},
  year={2025}
}

@misc{veo32025deepmind,
  title        = {Veo 3: Google DeepMind’s Third-Generation Text-to-Video Model},
  author       = {DeepMind / Google},
  howpublished = {Online technical report},
  month        = {May},
  year         = {2025},
  url          = {https://storage.googleapis.com/deepmind-media/veo/Veo-3-Tech-Report.pdf},
}

@misc{siméoni2025dinov3,
      title={DINOv3}, 
      author={Oriane Siméoni and Huy V. Vo and Maximilian Seitzer and Federico Baldassarre and Maxime Oquab and Cijo Jose and Vasil Khalidov and Marc Szafraniec and Seungeun Yi and Michaël Ramamonjisoa and Francisco Massa and Daniel Haziza and Luca Wehrstedt and Jianyuan Wang and Timothée Darcet and Théo Moutakanni and Leonel Sentana and Claire Roberts and Andrea Vedaldi and Jamie Tolan and John Brandt and Camille Couprie and Julien Mairal and Hervé Jégou and Patrick Labatut and Piotr Bojanowski},
      year={2025},
      eprint={2508.10104},
      archivePrefix={arXiv},
      primaryClass={cs.CV},
      url={https://arxiv.org/abs/2508.10104}, 
}

@inproceedings{radford2021learning,
  title={Learning transferable visual models from natural language supervision},
  author={Radford, Alec and Kim, Jong Wook and Hallacy, Chris and Ramesh, Aditya and Goh, Gabriel and Agarwal, Sandhini and Sastry, Girish and Askell, Amanda and Mishkin, Pamela and Clark, Jack and others},
  booktitle={International conference on machine learning},
  pages={8748--8763},
  year={2021},
  organization={PmLR}
}

@inproceedings{wang2020cnn,
  title={CNN-generated images are surprisingly easy to spot... for now},
  author={Wang, Sheng-Yu and Wang, Oliver and Zhang, Richard and Owens, Andrew and Efros, Alexei A},
  booktitle={Proceedings of the IEEE/CVF conference on computer vision and pattern recognition},
  pages={8695--8704},
  year={2020}
}

@article{bammey2023synthbuster,
  title={Synthbuster: Towards detection of diffusion model generated images},
  author={Bammey, Quentin},
  journal={IEEE Open Journal of Signal Processing},
  year={2023},
  publisher={IEEE}
}

@misc{T2VE_2025,
  author       = {1129ljc},
  title        = {T2VE: Text-Vision Embedding for Generalized AI-Generated Video Detection},
  year         = {2025},
  howpublished = {\url{https://github.com/1129ljc/T2VE}},
  note         = {GitHub repository}
}

@article{martin2023detection,
  title={Detection of AI-created images using pixel-wise feature extraction and convolutional neural networks},
  author={Martin-Rodriguez, Fernando and Garcia-Mojon, Rocio and Fernandez-Barciela, Monica},
  journal={Sensors},
  volume={23},
  number={22},
  pages={9037},
  year={2023},
  publisher={MDPI}
}

@article{zhong2023rich,
  title={Rich and poor texture contrast: A simple yet effective approach for ai-generated image detection},
  author={Zhong, Nan and Xu, Yiran and Qian, Zhenxing and Zhang, Xinpeng},
  journal={arXiv preprint arXiv:2311.12397},
  year={2023}
}

@inproceedings{wang2023dire,
  title={Dire for diffusion-generated image detection},
  author={Wang, Zhendong and Bao, Jianmin and Zhou, Wengang and Wang, Weilun and Hu, Hezhen and Chen, Hong and Li, Houqiang},
  booktitle={Proceedings of the IEEE/CVF International Conference on Computer Vision},
  pages={22445--22455},
  year={2023}
}

@article{zhang2022exposing,
  title={Exposing unseen GAN-generated image using unsupervised domain adaptation},
  author={Zhang, Mingxu and Wang, Hongxia and He, Peisong and Malik, Asad and Liu, Hanqing},
  journal={Knowledge-Based Systems},
  volume={257},
  pages={109905},
  year={2022},
  publisher={Elsevier}
}

@inproceedings{sha2023fake,
  title={De-fake: Detection and attribution of fake images generated by text-to-image generation models},
  author={Sha, Zeyang and Li, Zheng and Yu, Ning and Zhang, Yang},
  booktitle={Proceedings of the 2023 ACM SIGSAC Conference on Computer and Communications Security},
  pages={3418--3432},
  year={2023}
}

@article{wen2025busterx,
  title={BusterX: MLLM-Powered AI-Generated Video Forgery Detection and Explanation},
  author={Wen, Haiquan and He, Yiwei and Huang, Zhenglin and Li, Tianxiao and Yu, Zihan and Huang, Xingru and Qi, Lu and Wu, Baoyuan and Li, Xiangtai and Cheng, Guangliang},
  journal={arXiv preprint arXiv:2505.12620},
  year={2025}
}

@article{gao2025david,
  title={DAVID-XR1: Detecting AI-Generated Videos with Explainable Reasoning},
  author={Gao, Yifeng and Ding, Yifan and Su, Hongyu and Li, Juncheng and Zhao, Yunhan and Luo, Lin and Chen, Zixing and Wang, Li and Wang, Xin and Wang, Yixu and others},
  journal={arXiv preprint arXiv:2506.14827},
  year={2025}
}

@article{liang2022mind,
  title={Mind the gap: Understanding the modality gap in multi-modal contrastive representation learning},
  author={Liang, Victor Weixin and Zhang, Yuhui and Kwon, Yongchan and Yeung, Serena and Zou, James Y},
  journal={Advances in Neural Information Processing Systems},
  volume={35},
  pages={17612--17625},
  year={2022}
}

@article{diaconis1984asymptotics,
  title={Asymptotics of graphical projection pursuit},
  author={Diaconis, Persi and Freedman, David},
  journal={The annals of statistics},
  pages={793--815},
  year={1984},
  publisher={JSTOR}
}

@inproceedings{lin2014microsoft,
  title={Microsoft coco: Common objects in context},
  author={Lin, Tsung-Yi and Maire, Michael and Belongie, Serge and Hays, James and Perona, Pietro and Ramanan, Deva and Doll{\'a}r, Piotr and Zitnick, C Lawrence},
  booktitle={Computer Vision--ECCV 2014: 13th European Conference, Zurich, Switzerland, September 6-12, 2014, Proceedings, Part V 13},
  pages={740--755},
  year={2014},
  organization={Springer}
}

@article{d1973tests,
  title={Tests for departure from normality. Empirical results for the distributions of $b^2$ and $\sqrt{b}$},
  author={D'agostino, Ralph and Pearson, Egon S},
  journal={Biometrika},
  volume={60},
  number={3},
  pages={613--622},
  year={1973},
  publisher={Oxford University Press}
}

@article{anderson1954test,
  title={A test of goodness of fit},
  author={Anderson, Theodore W and Darling, Donald A},
  journal={Journal of the American statistical association},
  volume={49},
  number={268},
  pages={765--769},
  year={1954},
  publisher={Taylor and Francis}
}

@book{van2000asymptotic,
  title={Asymptotic statistics},
  author={Van der Vaart, Aad W},
  volume={3},
  year={2000},
  publisher={Cambridge university press}
}

@inproceedings{levi2025double,
  title     = {The Double Ellipsoid Geometry of CLIP},
  author    = {Levi, Meir Yossef and Gilboa, Guy},
  booktitle = {Proceedings of the 42nd International Conference on Machine Learning},
  series    = {Proceedings of Machine Learning Research},
  volume    = {267},
  year      = {2025},
  publisher = {PMLR},
  address   = {Vancouver, Canada},
}

@inproceedings{diaconis1987dozen,
  title={A dozen de Finetti-style results in search of a theory},
  author={Diaconis, Persi and Freedman, David},
  booktitle={Annales de l'IHP Probabilit{\'e}s et statistiques},
  volume={23},
  number={S2},
  pages={397--423},
  year={1987}
}

@article{wang2023internvid,
  title={Internvid: A large-scale video-text dataset for multimodal understanding and generation},
  author={Wang, Yi and He, Yinan and Li, Yizhuo and Li, Kunchang and Yu, Jiashuo and Ma, Xin and Li, Xinhao and Chen, Guo and Chen, Xinyuan and Wang, Yaohui and others},
  journal={arXiv preprint arXiv:2307.06942},
  year={2023}
}

@misc{wei2023dreamvideocomposingdreamvideos,
      title={DreamVideo: Composing Your Dream Videos with Customized Subject and Motion}, 
      author={Yujie Wei and Shiwei Zhang and Zhiwu Qing and Hangjie Yuan and Zhiheng Liu and Yu Liu and Yingya Zhang and Jingren Zhou and Hongming Shan},
      year={2023},
      eprint={2312.04433},
      archivePrefix={arXiv},
      primaryClass={cs.CV},
      url={https://arxiv.org/abs/2312.04433}, 
}

@article{interno2025ai,
  title={AI-Generated Video Detection via Perceptual Straightening},
  author={Intern{\`o}, Christian and Geirhos, Robert and Olhofer, Markus and Liu, Sunny and Hammer, Barbara and Klindt, David},
  journal={arXiv preprint arXiv:2507.00583},
  year={2025}
}

@inproceedings{howard2019searching,
  title={Searching for mobilenetv3},
  author={Howard, Andrew and Sandler, Mark and Chu, Grace and Chen, Liang-Chieh and Chen, Bo and Tan, Mingxing and Wang, Weijun and Zhu, Yukun and Pang, Ruoming and Vasudevan, Vijay and others},
  booktitle={Proceedings of the IEEE/CVF international conference on computer vision},
  pages={1314--1324},
  year={2019}
}

@inproceedings{he2016deep,
  title={Deep residual learning for image recognition},
  author={He, Kaiming and Zhang, Xiangyu and Ren, Shaoqing and Sun, Jian},
  booktitle={Proceedings of the IEEE conference on computer vision and pattern recognition},
  pages={770--778},
  year={2016}
}

@article{tong2022videomae,
  title={Videomae: Masked autoencoders are data-efficient learners for self-supervised video pre-training},
  author={Tong, Zhan and Song, Yibing and Wang, Jue and Wang, Limin},
  journal={Advances in neural information processing systems},
  volume={35},
  pages={10078--10093},
  year={2022}
}

@article{kay2017kinetics,
  title={The kinetics human action video dataset},
  author={Kay, Will and Carreira, Joao and Simonyan, Karen and Zhang, Brian and Hillier, Chloe and Vijayanarasimhan, Sudheendra and Viola, Fabio and Green, Tim and Back, Trevor and Natsev, Paul and others},
  journal={arXiv preprint arXiv:1705.06950},
  year={2017}
}

@article{bolya2025perception,
  title={Perception encoder: The best visual embeddings are not at the output of the network},
  author={Bolya, Daniel and Huang, Po-Yao and Sun, Peize and Cho, Jang Hyun and Madotto, Andrea and Wei, Chen and Ma, Tengyu and Zhi, Jiale and Rajasegaran, Jathushan and Rasheed, Hanoona and others},
  journal={arXiv preprint arXiv:2504.13181},
  year={2025}
}

@inproceedings{chen2011collecting,
  title={Collecting highly parallel data for paraphrase evaluation},
  author={Chen, David and Dolan, William B},
  booktitle={Proceedings of the 49th annual meeting of the association for computational linguistics: human language technologies},
  pages={190--200},
  year={2011}
}

@inproceedings{zhang2025NSGVD,
  title={Physics-Driven Spatiotemporal Modeling for AI-Generated Video Detection},
  author={Zhang, Shuhai and Lian, Zihao and Yang, Jiahao and Li, Daiyuan and Pang, Guoxuan and Liu, Feng and Han, Bo and Li, Shutao and Tan, Mingkui},
  booktitle={Advances in Neural Information Processing Systems},
  year={2025}
}

@article{wang2024vidprom,
  title={VidProM: A Million-scale Real Prompt-Gallery Dataset for Text-to-Video Diffusion Models},
  author={Wang, Wenhao and Yang, Yi},
  booktitle={Thirty-eighth Conference on Neural Information Processing Systems},
  year={2024},
  url={https://openreview.net/forum?id=pYNl76onJL}
}

@article{paszke2019pytorch,
  title={Pytorch: An imperative style, high-performance deep learning library},
  author={Paszke, Adam and Gross, Sam and Massa, Francisco and Lerer, Adam and Bradbury, James and Chanan, Gregory and Killeen, Trevor and Lin, Zeming and Gimelshein, Natalia and Antiga, Luca and others},
  journal={Advances in neural information processing systems},
  volume={32},
  year={2019}
}

@article{smirnov1939estimation,
  title={On the estimation of the discrepancy between empirical curves of distribution for two independent samples},
  author={Smirnov, Nikolai V},
  journal={Bull. Math. Univ. Moscou},
  volume={2},
  number={2},
  pages={3--14},
  year={1939}
}

@misc{cao2020losslessimagecompressionsuperresolution,
      title={Lossless Image Compression through Super-Resolution}, 
      author={Sheng Cao and Chao-Yuan Wu and Philipp Krähenbühl},
      year={2020},
      eprint={2004.02872},
      archivePrefix={arXiv},
      primaryClass={eess.IV},
      url={https://arxiv.org/abs/2004.02872}, 
}

@inproceedings{caron2021emerging,
  title={Emerging properties in self-supervised vision transformers},
  author={Caron, Mathilde and Touvron, Hugo and Misra, Ishan and J{\'e}gou, Herv{\'e} and Mairal, Julien and Bojanowski, Piotr and Joulin, Armand},
  booktitle={Proceedings of the IEEE/CVF international conference on computer vision},
  pages={9650--9660},
  year={2021}
}

@misc{ni2022expandinglanguageimagepretrainedmodels,
      title={Expanding Language-Image Pretrained Models for General Video Recognition}, 
      author={Bolin Ni and Houwen Peng and Minghao Chen and Songyang Zhang and Gaofeng Meng and Jianlong Fu and Shiming Xiang and Haibin Ling},
      year={2022},
      eprint={2208.02816},
      archivePrefix={arXiv},
      primaryClass={cs.CV},
      url={https://arxiv.org/abs/2208.02816}, 
}
}


\clearpage
\setcounter{section}{0}
\renewcommand{\thesection}{\Alph{section}}
\renewcommand{\thesubsection}{\Alph{section}.\arabic{subsection}}
\onecolumn
\begin{center}
    {\LARGE\bfseries Training-free Detection of Generated Videos\\via Spatial-Temporal Likelihoods\\[0.4em]--- Supplementary Material ---}\\[1em]
    {\large Omer Ben Hayun, Roy Betser, Meir Yossef Levi, Levi Kassel, Guy Gilboa}\\[0.3em]
    {\normalsize Viterbi Faculty of Electrical and Computer Engineering\\
    Technion -- Israel Institute of Technology, Haifa, Israel}\\[0.2em]
    {\tt\small \{omerben,roybe,me.levi,kassellevi\}@campus.technion.ac.il; guy.gilboa@ee.technion.ac.il}
\end{center}
\vspace{1em}

\begin{abstract}
In this supplementary material document, we provide additional implementation details to ensure the full reproducibility of STALL. We also present extended explanations of the statistical tests used to assess the normality of embeddings and the uniformity of the temporal representation features. Furthermore, we include additional experiments and experimental details on all experiments. Next, we provide further details on the newly introduced synthetic dataset, \emph{ComGenVid}, and important details on the other used benchmarks. We conclude with an efficiency analysis, comparing our method to other zero-shot and supervised methods.
The source code, dataset, and pre-computed whitening parameters are publicly available \href{https://omerbenhayun.github.io/stall-video}{here}.
\begin{enumerate}[label=\Alph*.]
    \item Reproducibility (\Cref{supp: rep}).
    \item Statistical tests (\Cref{supp: tests}).
    \item Datasets (\Cref{supp: data}).
    \item Experimental details and additional results (\Cref{supp: exps}).
    \item Efficiency analysis (\Cref{supp: eff}).
\end{enumerate}
\end{abstract}

\section{Reproducibility}
\label{supp: rep}
\subsection{Detailed algorithms}


To ensure complete reproducibility, we provide full implementation details, including detailed algorithms for whitening (\Cref{alg:whitening}), scores (\Cref{alg:spatial_score}), (\Cref{alg:temporal_score}), calibration (\Cref{alg:calibration}), and inference (\Cref{alg:inference}).

\subsubsection{Notation}
\begin{itemize}
    \item $\mathcal{C} = \{c^{(i)}\}_{i=1}^{N_c}$: Calibration set of $N_c$ real videos
    \item $c^{(i)} = \{f_t^{(i)}\}_{t=1}^{T}$: Calibration video $i$ consisting of $T$ frames.
    \item $v = \{f_t\}_{t=1}^{T}$: query video with $T$ frames.
    \item $E: \mathbb{R}^{H \times W \times 3} \rightarrow \mathbb{R}^d$: Image vision encoder.
    \item $x_t = E(f_t) \in \mathbb{R}^d$: Embedding of frame $f_t$.
    \item $\Delta_t = x_{t+1} - x_t$: Temporal difference between consecutive embeddings.
    \item $\tilde{\Delta}_t = \Delta_t / \|\Delta_t\|_2$: $\ell_2$-normalized temporal difference.
    \item $\mu, W$: Mean and whitening matrix for spatial embeddings
    \item $\mu_\Delta, W_\Delta$: Mean and whitening matrix for temporal embeddings
    \item $s_{\text{spat}}, s_{\text{temp}}$ spatial and temporal scores.
\end{itemize}

\subsubsection{Algorithms}

\begin{algorithm}[H]
\caption{Compute Whitening Transform}
\label{alg:whitening}
\begin{algorithmic}[1]
\Require Embeddings $\mathcal{X} = \{x_i\}_{i=1}^N \subset \mathbb{R}^d$
\State $\mu \gets \frac{1}{N} \sum_{i=1}^{N} x_i$
\State $\hat{x}_i \gets x_i - \mu$ for $i = 1, \ldots, N$
\State $\mathbf{\hat{X}} \gets [\hat{x}_1, \ldots, \hat{x}_N]$
\State $\mathbf{\Sigma} \gets \frac{1}{N} \mathbf{\hat{X}} \mathbf{\hat{X}}^\top$
\State Eigendecomposition: $\mathbf{\Sigma} = V \Lambda V^\top$
\State $W \gets \Lambda^{-1/2} V^\top$
\State \Return $\mu, W$
\end{algorithmic}
\end{algorithm}

\begin{algorithm}[H]
\caption{Compute Spatial Score}
\label{alg:spatial_score}
\begin{algorithmic}[1]
\Require Frame embeddings $\{x_t\}_{t=1}^T$, parameters $(\mu, W)$

\State $y_t \gets W(x_t - \mu)$ for $t = 1, \ldots, T$
\State $\ell_{\text{spat}}(t) \gets -\frac{1}{2}(d \log(2\pi) + \|y_t\|_2^2)$ for $t = 1, \ldots, T$

\State $s_{\text{spat}} \gets \max\{\ell_{\text{spat}}(t)\}_{t=1}^T$

\State \Return $s_{\text{spat}}$
\end{algorithmic}
\end{algorithm}

\begin{algorithm}[H]
\caption{Compute Temporal Score}
\label{alg:temporal_score}
\begin{algorithmic}[1]
\Require Frame embeddings $\{x_t\}_{t=1}^T$, parameters $(\mu_\Delta, W_\Delta)$

\State $\Delta_t \gets x_{t+1} - x_t$ for $t = 1, \ldots, T-1$ \Comment{Temporal differences}
\State $\tilde{\Delta}_t \gets \frac{\Delta_t}{\|\Delta_t\|_2}$ for $t = 1, \ldots, T-1$ \Comment{Normalization}

\State $z_t \gets W_\Delta(\tilde{\Delta}_t - \mu_\Delta)$ for $t = 1, \ldots, T-1$
\State $\ell_{\text{temp}}(t) \gets -\frac{1}{2}(d \log(2\pi) + \|z_t\|_2^2)$ for $t = 1, \ldots, T-1$

\State $s_{\text{temp}} \gets \min\{\ell_{\text{temp}}(t)\}_{t=1}^{T-1}$

\State \Return $s_{\text{temp}}$
\end{algorithmic}
\end{algorithm}

\begin{algorithm}[H]
\caption{STALL Calibration}
\label{alg:calibration}
\begin{algorithmic}[1]
\Require Calibration set $\mathcal{C} = \{c^{(i)}\}_{i=1}^{N_c}$ of $N_c$ videos, each with $T$ frames, encoder $E$
\State \textit{Encode all frames from calibration set:}
\For{$i = 1$ to $N_c$}
    \State $x_t^{(i)} \gets E(f_t^{(i)})$ for $t = 1, \ldots, T_i$
\EndFor
\State \textit{Compute spatial whitening parameters:}
\State $\mathcal{X}_{\text{spat}} \gets \emptyset$
\For{$i = 1$ to $N_c$}
    \State Sample one frame: $x \sim \text{Uniform}(\{x_t^{(i)}\}_{t=1}^{T_i})$
    \State $\mathcal{X}_{\text{spat}} \gets \mathcal{X}_{\text{spat}} \cup \{x\}$
\EndFor
\State $(\mu, W) \gets \textsc{Compute Whitening Transform}(\mathcal{X}_{\text{spat}})$ \Comment{Algorithm~\ref{alg:whitening}}
\State \textit{Compute temporal whitening parameters:}
\State $\mathcal{X}_{\text{temp}} \gets \emptyset$
\For{$i = 1$ to $N_c$}
    \For{$t = 1$ to $T_i - 1$}
        \State $\Delta_t \gets x_{t+1}^{(i)} - x_t^{(i)}$
        \State $\tilde{\Delta}_t \gets \frac{\Delta_t}{\|\Delta_t\|_2}$
        \State $\mathcal{X}_{\text{temp}} \gets \mathcal{X}_{\text{temp}} \cup \{\tilde{\Delta}_t\}$
    \EndFor
\EndFor
\State $(\mu_\Delta, W_\Delta) \gets \textsc{Compute Whitening Transform}(\mathcal{X}_{\text{temp}})$ \Comment{Algorithm~\ref{alg:whitening}}
\State \textit{Compute calibration score distributions:}
\For{$i = 1$ to $N_c$}
    \State $s_{\text{spat}}^{(i)} \gets \textsc{Compute Spatial Score}(\{x_t^{(i)}\}_{t=1}^{T_i}, \mu, W)$ \Comment{Algorithm~\ref{alg:spatial_score}}
    \State $s_{\text{temp}}^{(i)} \gets \textsc{Compute Temporal Score}(\{x_t^{(i)}\}_{t=1}^{T_i}, \mu_\Delta, W_\Delta)$ \Comment{Algorithm~\ref{alg:temporal_score}}
\EndFor
\State $\mathcal{S}_{\text{spat}} \gets \{s_{\text{spat}}^{(i)}\}_{i=1}^{N_c}$
\State $\mathcal{S}_{\text{temp}} \gets \{s_{\text{temp}}^{(i)}\}_{i=1}^{N_c}$
\State \Return $\mu, W, \mu_\Delta, W_\Delta, \mathcal{S}_{\text{spat}}, \mathcal{S}_{\text{temp}}$
\end{algorithmic}
\end{algorithm}

\begin{algorithm}[H]
\caption{STALL Inference}
\label{alg:inference}
\begin{algorithmic}[1]
\Require Test video $v = \{f_t\}_{t=1}^T$, encoder $E$, calibration parameters 
$\mu, W, \mu_\Delta, W_\Delta, \mathcal{S}_{\text{spat}}, \mathcal{S}_{\text{temp}}$
\State \textit{// Encode all frames from test video}
\State $x_t \gets E(f_t)$ for $t = 1, \ldots, T$
\State \textit{Compute spatial and temporal scores:}
\State $s_{\text{spat}} \gets \textsc{Compute Spatial Score}(\{x_t\}_{t=1}^T, \mu, W)$ \Comment{Algorithm~\ref{alg:spatial_score}}
\State $s_{\text{temp}} \gets \textsc{Compute Temporal Score}(\{x_t\}_{t=1}^T, \mu_\Delta, W_\Delta)$ \Comment{Algorithm~\ref{alg:temporal_score}}
\State \textit{Compute percentile ranks from calibration distributions:}
\State $\text{perc}_{\text{spat}} \gets \frac{1}{N_c} |\{s \in \mathcal{S}_{\text{spat}} : s \leq s_{\text{spat}}\}|$
\State $\text{perc}_{\text{temp}} \gets \frac{1}{N_c} |\{s \in \mathcal{S}_{\text{temp}} : s \leq s_{\text{temp}}\}|$
\State \textit{Combine percentiles for final detection score:}
\State $s_{\text{video}} \gets \frac{1}{2}(\text{perc}_{\text{spat}} + \text{perc}_{\text{temp}})$
\State \Return $s_{\text{video}}$
\end{algorithmic}
\end{algorithm}

\subsection{Implementation details}
All of our experiments are conducted using Intel Core i9-7940X CPU and NVIDIA GeForce RTX 3090 GPU.
For our method we use DINOv3~\citep{siméoni2025dinov3} as our encoder, available at \href{https://github.com/facebookresearch/dinov3}{\texttt{DINOv3 repository}}.
For all competing methods, we rely on the official implementations when available, otherwise, we implement the corresponding baselines.

\noindent For AEROBLADE \cite{ricker2024aeroblade}, we use the official implementation
available at \href{https://github.com/jonasricker/aeroblade}{AEROBLADE repository}.
Since no official implementation is available for RIGID \cite{he2024rigid}, we implement the method using the DINO~\cite{caron2021emerging} model. As there is no official implementation for ZED \cite{cozzolino2024zero} either, we follow the lossless compression setup based on SReC~\cite{cao2020losslessimagecompressionsuperresolution}. ZED introduces four separate criteria, and we report results for the $\Delta^{01}$ criterion. For AIGVDet~\citep{bai2024ai} and T2VE~\citep{T2VE_2025}, we use the official code and pretrained weights released by the authors, available at \href{https://github.com/multimediaFor/AIGVDet}{AIGVDet repository} and \href{https://github.com/1129ljc/T2VE}{T2VE repository}, respectively. For D3~\cite{zheng2025d3}, we rely on the official implementation at \href{https://github.com/Zig-HS/D3}{D3 repository}, and use DINOv3 as the encoder. 

\subsection{Comparison Methodology}

\subsubsection{Video Preprocessing}
\label{sup:video_downsample}
In all of our evaluations, we filter out videos that are shorter than 2 seconds or have a frame rate below 8 FPS to ensure sufficient temporal coverage and quality.
After filtering, we subsample frames to achieve a target frame rate of 8 FPS.
Given an original frame rate $f_{\text{orig}}$ and target rate $f_{\text{target}} = 8$ FPS, we compute the sampling ratio $r = f_{\text{orig}} / f_{\text{target}}$ and select every $r$-th frame (approximately). 

\noindent Specifically, we maintain a continuous position that advances by $r$ at each step, selecting the frame at the rounded position:
\begin{equation*}
i_0 = 0, \quad i_{j} = \text{round}(r \cdot j), \quad j = 1, 2, \ldots
\end{equation*}
subject to $i_{j} < N$, where $N$ is the total number of frames of the original video.

\noindent When $f_{\text{orig}}$ is perfectly divisible by $f_{\text{target}}$ (i.e., $r$ is an integer), this reduces to uniform sampling of every $r$-th frame. For non-integer ratios, this approach selects frames with approximately uniform temporal spacing that best approximates the target frame rate.
The corresponding Python implementation is provided below:
\begin{lstlisting}
def downsample_frames(num_frames, current_fps, target_fps=8):
    """Downsample frame indices to achieve target fps."""
    ratio = current_fps / target_fps
    indices = []
    j = 0

    while True:
        frame_idx = round(ratio * j)
        if frame_idx >= num_frames:
            break
        indices.append(frame_idx)
        j += 1

    return indices
\end{lstlisting}

\noindent Unless stated otherwise, all videos in our experiments were sampled at 8 FPS and truncated to 2 seconds, yielding 16 frames per video.

\subsubsection{Pairwise Comparison Protocol}
\label{pairwise_comparison_protocol}
We conduct systematic pairwise comparisons between synthetic videos from each generative model and authentic videos. To ensure fair metric comparisons and to address data imbalance, we implement a balanced sampling procedure where we use equal numbers of real and generated videos, determined by the smaller class in each split.
For each generative model $M_i$ with $N_i$ synthetic videos, we sample exactly $N_i$ authentic videos from our real video datasets. Specifically, in the Videofeedback benchmark~\cite{he2024videoscore}, the real videos are drawn from two datasets~\citep{chen2024panda, anne2017localizing}. To prevent bias from over-representation of any specific real dataset source, we sample an equal number of videos from both sources such that their total equals $N_i$.

\subsection{D3 Baseline Evaluation}
\label{d3_baseline_evaluation}
We identified two systematic differences between D3's~\citep{zheng2025d3} official evaluation protocol and ours that explain the performance gap reported in the main paper.

\subsubsection{Unbalanced test set}
When fewer than 1000 synthetic videos are available, 1000 real videos are compared against fewer synthetic samples (on GenVideo~\citep{chen2024demamba}, this is most notably the case with Sora~\citep{sora2024openai}, which contains only 56 samples). Average Precision (AP) is sensitive to class imbalance and tends to inflate. Our pairwise comparison protocol (Section~\ref{pairwise_comparison_protocol}) uses equal numbers of real and generated videos to ensure unbiased evaluation.

\subsubsection{FPS upsampling by frame duplication}
In the GenVideo ~\citep{chen2024demamba} benchmark, the MSR-VTT~\citep{xu2016msr} real videos provided on ModelScope are stored at only 3~FPS. D3's official code brings these to 8~FPS by duplicating frames. This duplication applies exclusively to real videos, because the generated videos in GenVideo are already at 8~FPS or higher. The resulting temporal redundancy inflates detection scores for methods that rely on inter-frame differences. In our evaluation, we download high-frame-rate MSR-VTT videos and uniformly downsample them to 8~FPS, which removes this artifact.
for more details, see ~\Cref{sup:genvideo_benchmark,sup:video_downsample}.

\subsubsection{D3 Ablation Results}
\label{d3_ablation_results}

In our main experiments, we evaluated D3 using the same embedder as our method (DinoV3 \cite{siméoni2025dinov3}) for consistency. For completeness, we now report D3 results using X-CLIP16 \cite{ni2022expandinglanguageimagepretrainedmodels} as embedder, as conducted in the original D3 implementation, which yields a small performance increase.
\Cref{tab:D3_ablate} shows how D3 mean AP on GenVideo varies across evaluation settings: choice of embedder, test-set balancing and sampling. Each protocol difference individually inflates AP, and their combination produces a large gap relative to our evaluation.

\begin{table}[ht]
    \centering
    \caption{D3~\citep{zheng2025d3} mean AP across evaluation settings on GenVideo. The two protocol differences (class imbalance and FPS duplication) each inflate AP; together they fully explain the gap between D3's reported numbers and our evaluation.}
    \begin{adjustbox}{max width=0.65\linewidth}
    \begin{tabular}{|c|cccc|}
        \toprule
        \textbf{Real video Sampling} & \multicolumn{2}{c}{\textbf{Downsample}} & \multicolumn{2}{c|}{\textbf{Upsample (Duplication)}} \\
        \textbf{FPS} & \multicolumn{2}{c}{\textbf{$\sim27\rightarrow8$}} & \multicolumn{2}{c|}{\textbf{$3\rightarrow8$}} \\
        \cmidrule(lr){1-1} \cmidrule(lr){2-3} \cmidrule(lr){4-5}
        \textbf{Embedder} & \textbf{Balanced} & \textbf{Unbalanced} & \textbf{Balanced} & \textbf{Unbalanced} \\
        \midrule
        X-CLIP16 \cite{ni2022expandinglanguageimagepretrainedmodels} & 0.78 & 0.85 & 0.97 & 0.98 \\
        DinoV3 \cite{siméoni2025dinov3}  & 0.74 & 0.83 & 0.94 & 0.96 \\
        \bottomrule
    \end{tabular}
    \end{adjustbox}
    \label{tab:D3_ablate}
\end{table}
\clearpage
\section{Normality measures and tests}
\label{supp: tests}

\subsection{Normality and uniformity tests}
To assess whether the embeddings are approximately Gaussian, we apply two classical normality tests, Anderson–Darlin~ \cite{anderson1954test} and D’Agostino–Pearson~\cite{d1973tests}, performed on each coordinate of the embeddings independently. Each test measures the normality of the one-dimensional vector input. Below we elaborate on each test and present results.

\subsubsection{Anderson-Darling normality test}
The Anderson-Darling (AD) test \cite{anderson1954test} can be viewed as a refinement of the classical Kolmogorov-Smirnov (KS) goodness-of-fit test \cite{smirnov1939estimation}, designed to put more weight on discrepancies in the tails of the distribution. Given a sample $X = \{x_1, x_2, \dots, x_n\}$ and a target cumulative distribution function (CDF) of $F$ (in our case, a normal CDF with parameters estimated from the data), the AD statistic is defined as
\begin{equation}
A^2 = -n - \sum_{i=1}^{n} \frac{2i - 1}{n}
\bigl[\ln F(x_i) + \ln\bigl(1 - F(x_{n+1-i})\bigr)\bigr],
\label{eq:anderson_darling}
\end{equation}
where $x_1 \le \dots \le x_n$ are the ordered sample values, $F(x)$ is the CDF of the reference normal distribution, and $n$ is the sample size. Larger values of $A^2$ indicate stronger deviations from normality. these values are compared to tabulated critical values to decide whether to reject the Gaussian assumption.
 In our setting, we follow the conventional threshold $A^2 < 0.752$ as evidence to accept normality. To preform this test, we used \texttt{stats.anderson} function from \texttt{scipy} python package.

\subsubsection{D’Agostino-Pearson Test}
The D'Agostino-Pearson (DP) test \cite{d1973tests} evaluates departures from normality by combining information about sample skewness and kurtosis. Let $X = \{x_1, x_2, \dots, x_n\}$ be a univariate sample and let $\mu = \frac{1}{n}\sum_{i=1}^{n} x_i$ denote its mean. Let $m_i = \frac{1}{n}\sum_{j=1}^{n}(x_j - \mu)^i$ be the $i$-th centeral moment. The skewness $g_1$ and kurtosis $g_2$ are defined as:
\begin{align}
g_1 &= \frac{m_3}{m_2^{3/2}} = \frac{\frac{1}{n}\sum_{i=1}^{n}(x_i - \mu)^3}{\left[\frac{1}{n}\sum_{i=1}^{n}(x_i - \mu)^2\right]^{3/2}} \\
g_2 &= \frac{m_4}{m_2^2} - 3 = \frac{\frac{1}{n}\sum_{i=1}^{n}(x_i - \mu)^4}{\left[\frac{1}{n}\sum_{i=1}^{n}(x_i - \mu)^2\right]^2} - 3
\end{align}
These two statistics are then transformed into approximately standard normal variables $Z_1$ and $Z_2$, and the DP test statistic is
\begin{equation}
K^2 = Z_1^2 + Z_2^2.
\label{eq:dagostino_pearson}
\end{equation}
Under the null hypothesis of normality, $K^2$ approximately follows a chi square distribution with two degrees of freedom, so the $p$-value is
\begin{equation}
p = 1 - F_{\chi^2_2}(K^2),
\end{equation}
where $F_{\chi^2_2}$ is the cumulative distribution function of $\chi^2_2$. Positive $g_1$ indicates right-skewed data and negative $g_1$ left-skewed data, while positive $g_2$ corresponds to heavy tails and negative $g_2$ to light tails. As $g_1$ and $g_2$ approach zero, the test statistic $K^2$ decreases and the $p$-value increases, which is consistent with normality; in practice we treat $p > 0.05$ as compatible with a Gaussian distribution. We performed this test using the \texttt{stats.normaltest} function from the \texttt{scipy} Python package.

\subsubsection{Results}
To obtain stable estimates of normality, we randomly sample $40$ independent groups of $250$ embeddings each from the VATEX~\cite{wang2019vatex} calibration set, for both frame embeddings and frame embedding differences. As described and used in the main paper, we employ DINOv3~\cite{siméoni2025dinov3} as the frame level embedder, whose embedding space is $1024$ dimensional. For every group and every coordinate, we apply the Anderson Darling (AD) and D’Agostino Pearson (DP) normality tests. We then aggregate the outcomes in two separate ways: (i) we average the test statistics across all coordinates and groups, and (ii) we compute the fraction of coordinates whose average statistic satisfies the normality thresholds ($A^2 < 0.752$ for AD, $p > 0.05$ for DP). As summarized in Table~\ref{tab:normality_vatex}, raw frame embeddings $x_t$ show high proportions of coordinates passing both tests, and whitening further increases these fractions. In contrast, raw temporal differences $\Delta_t$ are strongly non-Gaussian, with essentially no coordinates passing either criterion. After $\ell_2$ normalization, temporal differences $\tilde{\Delta}_t$  exhibit high pass rates, and an additional whitening step $z_t$ yields almost all coordinates satisfying both normality thresholds.

\begin{table}[t]
    \caption{\textbf{Normality tests.} Results of Anderson Darling (AD) and D’Agostino Pearson (DP) normality tests for different representations on the VATEX~\cite{wang2019vatex} calibration set. ``Avg Score'' is the mean test statistic across embedding coordinates and groups, ``Threshold'' specifies the acceptance condition for normality, and ``Normal Features'' is the percentage of coordinates satisfying that condition.}
\begin{tabular}{llcccc}
	\hline
	Representation                                                         & Test & Avg Score & Threshold & Normal Features $(\uparrow)$ & approximately gaussian? \\
	\hline
	\multirow{2}{*}{Raw embeddings $x_t$}                             & AD   & 0.4750    & $< 0.752$ & 96.3\%          & \cmark                  \\
	                                                                  & DP   & 0.3931    & $> 0.05$  & 99.2\%          & \cmark                  \\
	\hline
	\multirow{2}{*}{Whitened embeddings $y_t$}                        & AD   & 0.5034    & $< 0.752$ & 98.2\%          & \cmark                  \\
	                                                                  & DP   & 0.3207    & $> 0.05$  & 99.3\%          & \cmark                  \\
	\hline

	\multirow{2}{*}{Transition vector $\Delta_t=x_{t+1}-x_t$}         & AD   & 3.3992    & $< 0.752$ & 0.0\%           & \xmark                  \\
	                                                                  & DP   & 0.0093    & $> 0.05$  & 0.0\%           & \xmark                  \\
	\hline
	\multirow{2}{*}{$\ell_2$ Normalized transition vector  $\tilde{\Delta}_t$} & AD   & 0.4134    & $< 0.752$ & 98.4\%          & \cmark                  \\
	                                                                  & DP   & 0.4752    & $> 0.05$  & 99.9\%          & \cmark                  \\
	\hline
	\multirow{2}{*}{Whitened $\ell_2$ normalized transition vector $z_t$}      & AD   & 0.4119    & $< 0.752$ & 99.6\%          & \cmark                  \\
	                                                                  & DP   & 0.4648    & $> 0.05$  & 100.0\%         & \cmark                  \\
	\hline
\end{tabular}
\label{tab:normality_vatex}
\end{table}
    
\subsection{Histogram comparisons}
\subsubsection{Raw vs. normalized temporal differences}

\begin{figure}
    \centering
    \includegraphics[width=1\linewidth]{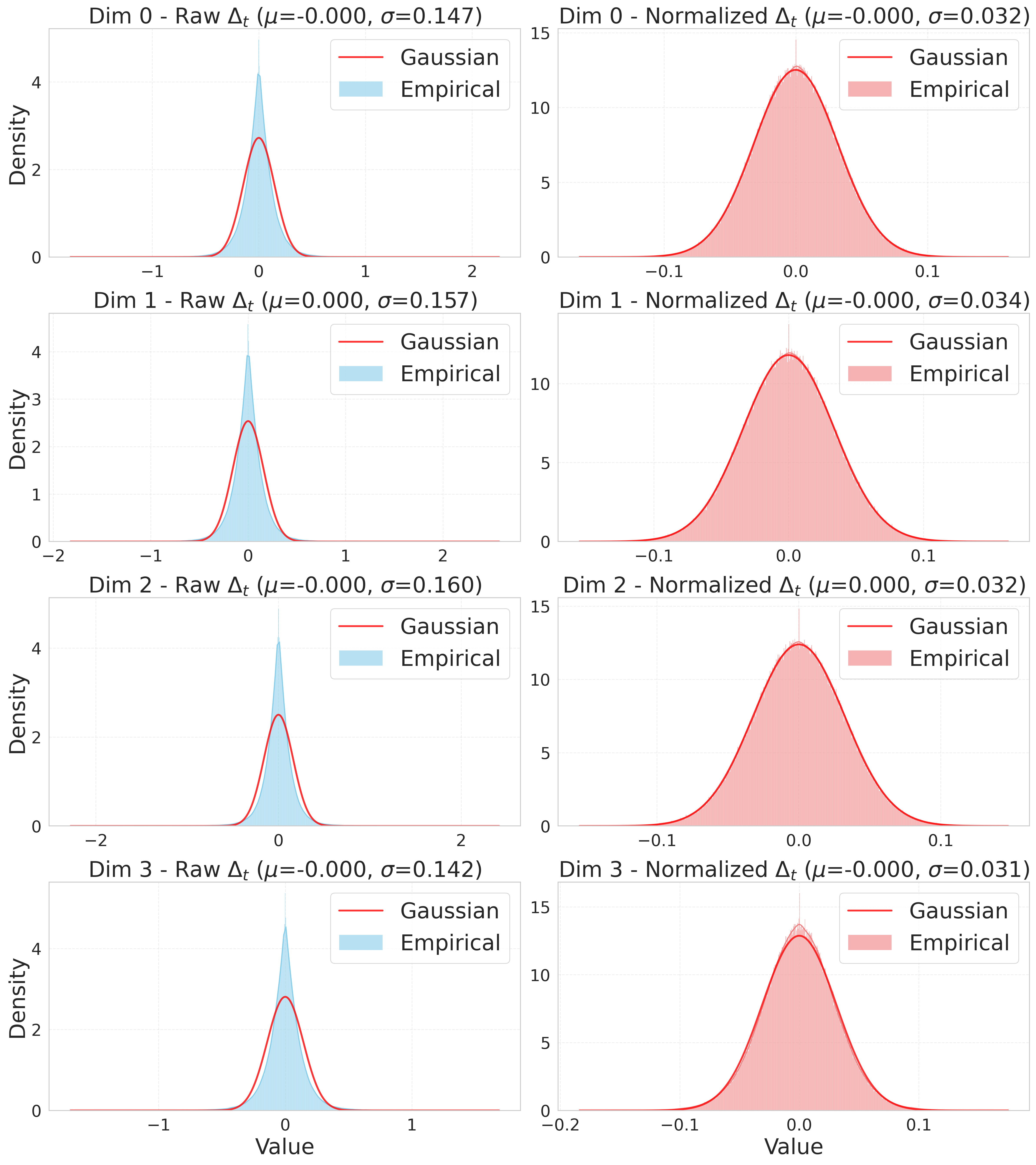}
    \caption{\textbf{Raw vs. Normalized Temporal Difference Histograms.} Histogram comparison of raw temporal differences (Raw $\Delta_t$, left) versus normalized temporal differences (Normalized $\Delta_t$, right) for the first four dimensions of DINOv3~\cite{siméoni2025dinov3} embeddings, computed from all adjacent frame pairs in the VATEX~\cite{wang2019vatex} calibration set. Red curves show Gaussian distributions fitted using the empirical mean and variance of the data (i.e., a moment-matched Gaussian). Even under this moment-matched fit, raw differences exhibit clear deviations from Gaussianity, while normalized differences closely match Gaussian distributions across all dimensions.}
    \label{fig:normality_comparison}
\end{figure}

To qualitatively illustrate these findings, Figure~\ref{fig:normality_comparison} presents histograms of temporal differences for the first four embedding dimensions of DINOv3~\cite{siméoni2025dinov3}, computed from all adjacent frame pairs in the VATEX~\cite{wang2019vatex} calibration set. The left column shows raw temporal differences (Raw $\Delta_t$), which exhibit significant deviations from the overlaid Gaussian distributions. In contrast, the right column shows $\ell_2$-normalized temporal differences (Normalized $\Delta_t$), which align closely with their corresponding Gaussian fits. This visual demonstration confirms the quantitative results in Table~\ref{tab:normality_vatex}: $\ell_2$ normalization transforms non-Gaussian temporal differences into approximately Gaussian distributions.

\subsubsection{Real and AI-generated videos embeddings}

\begin{figure}
    \centering
    \includegraphics[width=\linewidth]{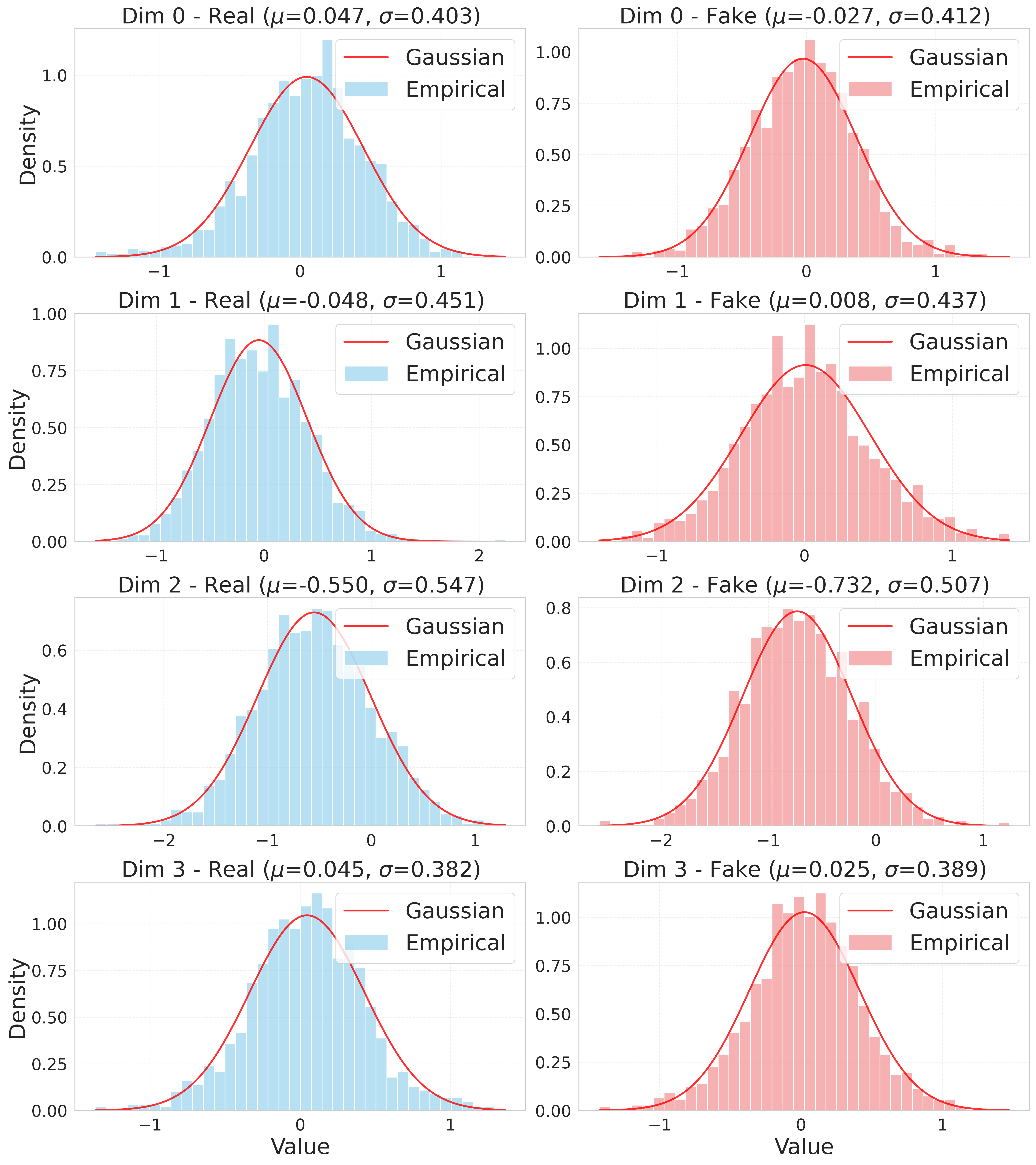}
    \caption{\textbf{Real and AI-generated videos embeddings.}
    Histograms of selected embedding dimensions from DINOv3~\cite{siméoni2025dinov3} features extracted from randomly sampled frames in the GenVideo~\cite{chen2024demamba} dataset. The left column corresponds to real videos, and the right column corresponds to generated videos. Red curves show Gaussian distributions fitted using the empirical mean and variance of each feature dimension.}
    \label{fig:real_fake_feature_hist}
\end{figure}

To further illustrate the statistical properties of the embedding space, we visualize the univariate feature distributions of embeddings extracted from real and generated videos. Figure~\ref{fig:real_fake_feature_hist} shows histograms for the first four dimensions of DINOv3~\cite{siméoni2025dinov3} embeddings, computed from randomly sampled frames from real and generated videos in the GenVideo~\cite{chen2024demamba} dataset.

For each dimension, we overlay a Gaussian distribution fitted using the empirical mean and variance of the data (i.e., a moment-matched Gaussian). The empirical distributions closely follow the corresponding Gaussian curves for both real and generated samples. While the means and variances differ slightly between real and generated content, the overall shapes remain approximately Gaussian.
These visualizations provide additional qualitative support for modeling the embedding dimensions using Gaussian statistics, which underlies the likelihood formulation used in our method.

\subsection{Maxwell Poincare Lemma}
\begin{lemma}[Maxwell-Poincaré \citep{diaconis1984asymptotics}]
\label{lem:maxwell}
Let $U_d$ be uniform on $\mathbb{S}^{d-1}$ and fix $k\in\mathbb N$. Then
\begin{equation}
\label{eq:maxwell}
\sqrt{d}(U_{d,1},\ldots,U_{d,k}) \;\Rightarrow\; \mathcal N(0,I_k) \qquad (d\to\infty).
\end{equation}

\end{lemma}
\noindent The rate at which this convergence occurs was quantified by \citet{diaconis1987dozen}:
\begin{theorem}[Maxwell-Poincaré convergence rate\citep{diaconis1987dozen}]
If $1\leq k \leq d-4$, then 
\begin{equation}
\label{eq:tv-rate}
d_{\mathrm{TV}}\!\big(\sqrt{d}(U_{d,1},\ldots,U_{d,k}),Z\big)
\;\leq\;\frac{2(k+3)}{d-k-3}\;,
\end{equation}
where $Z \sim \mathcal{N}(0,I_k)$.
\label{thm:conv_rate}
\end{theorem}

\noindent Note that both Lemma~\ref{lem:maxwell} and Theorem~\ref{thm:conv_rate} extend naturally to arbitrary coordinate selections, or more generally to any $k$-dimensional orthonormal projection of $U_d$.
\noindent When norms exhibit concentration behavior, an analogous result can be established:

\begin{lemma}[Maxwell–Poincaré with norm concentration]
\label{lem:slut}
Let $U_d$ be as in Lemma~\ref{lem:maxwell} and fix $k\in\mathbb N$.
Let $z_d = r_d U_d$ with $r_d \ge 0$.
If $r_d \xrightarrow[d\to\infty]{\mathsf P} r_0$, then
\begin{equation}
\label{eq:slutsky-zk}
r_d\,\sqrt{d}\,(U_{d,1},\ldots,U_{d,k})
\;\Rightarrow\;
\mathcal N\!\big(0,\,r_0^2 I_k\big) \qquad (d\to\infty).
\end{equation}
\end{lemma}

\noindent This is obtained by combining the Maxwell–Poincaré limit with radial concentration and applying Slutsky’s theorem~\citep{van2000asymptotic}.

\begin{figure}
    \centering
    \includegraphics[width=1\linewidth]{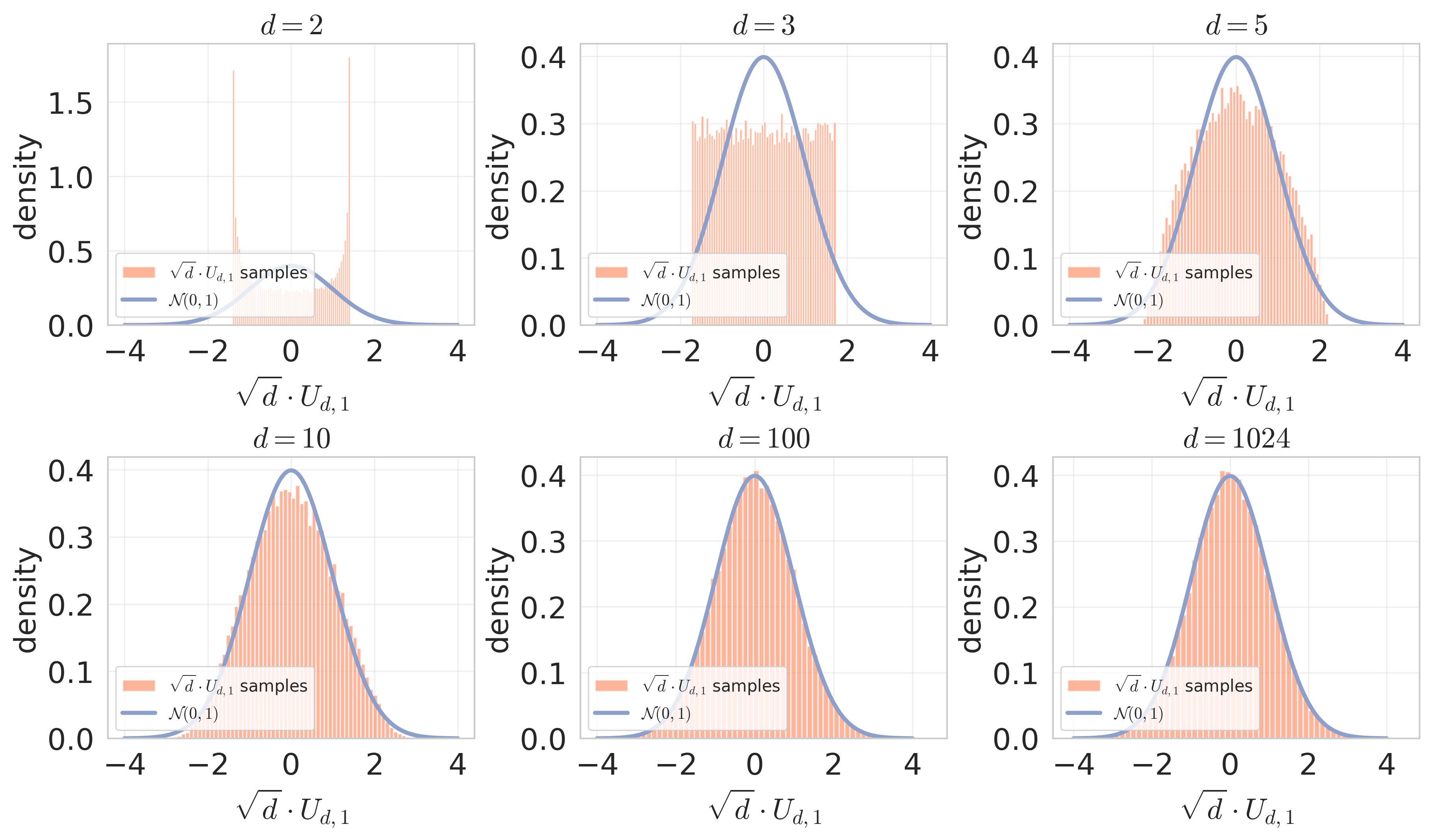}
    \caption{\textbf{Uniformity on the sphere and Gaussian distribution.}
    Histograms of the first coordinate of $U_d$, where $U_d$ is uniformly
    distributed on the $d$-dimensional unit sphere $\mathbb{S}^{d-1}$, shown
    for several values of $d$ and overlaid with the $\mathcal{N}(0,1)$ density.}
    
    \label{fig:mpb_scaled_combined_3x2}
\end{figure}

\noindent Figure~\ref{fig:mpb_scaled_combined_3x2} visualizes the distribution
of the first coordinate of $U_d$, a random vector uniformly distributed on
$\mathbb{S}^{d-1}$, for several dimensions $d$. As $d$ increases, this
distribution approaches the standard normal distribution.

\begin{figure}
    \centering
    \includegraphics[width=0.5\linewidth]{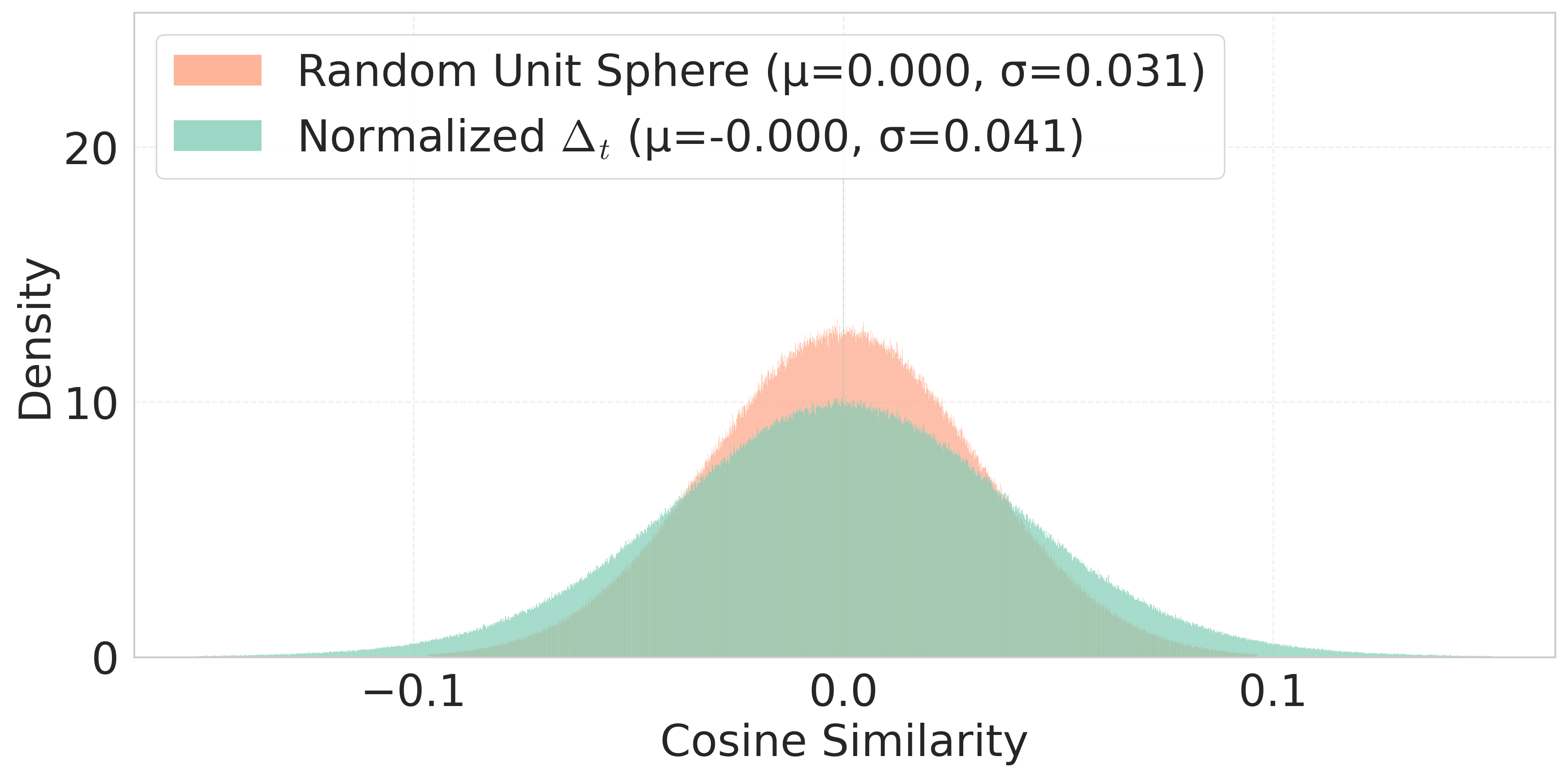}
    \caption{\textbf{Uniformity on the sphere.} Cosine similarity distributions over all unordered pairs of $3k$ randomly sampled normalized temporal differences $\tilde{\Delta}_t$ from the VATEX \cite{wang2019vatex} calibration set and $3k$ points drawn uniformly from the unit sphere. 
    }
    \label{fig:cosine_diff_vs_sphere}
\end{figure}

\noindent Figure~\ref{fig:cosine_diff_vs_sphere} shows histograms of cosine similarities over all unordered pairs of $3k0$ randomly sampled normalized temporal differences $\tilde{\Delta}_t\subset \mathbb{R}^{1024}$ from the VATEX~\cite{wang2019vatex} calibration set and of $3k$ points drawn uniformly from $\mathbb{S}^{1023}$.

\clearpage
\section{Datasets}
\label{supp: data}

\subsection{ComGenVid benchmark}
for full statistics of our \textit{ComGenVid} benchmark, refer to table \ref{tab:ComGenVidTable}

\paragraph{MSVD Dataset.}
We obtained the complete MSVD dataset~\cite{chen2011collecting} from the \href{https://www.kaggle.com/datasets/sarthakjain004/msvd-clips/data}{sarthakjain004 Kaggle repository} using the kaggle command line interface:
\begin{lstlisting}[language=bash]
kaggle datasets download -d sarthakjain004/msvd-clips --unzip
\end{lstlisting}
We sampled 1.7k videos with 2 seconds length from this dataset. 

The complete set of sampled videos is listed in \texttt{msvd\_sampled\_videos.csv}.
\paragraph{Generative Model Videos.}
To ensure fair comparison, we sampled 1.7k videos from both Sora~\cite{brooks2024video} and Veo3~\cite{veo32025deepmind} to match the MSVD dataset size.

\noindent \textbf{Veo3 Sampling.}
We randomly selected 1.7k videos from the ShareVeo3 dataset~\cite{wang2024vidprom}, available on the \href{https://huggingface.co/datasets/WenhaoWang/ShareVeo3}{WenhaoWang Hugging Face repository}, and can be download with the following python script:
\begin{lstlisting}
# pip install huggingface_hub[hf_xet]
from huggingface_hub import hf_hub_download

for i in range(1, 51):
    hf_hub_download(
        repo_id="WenhaoWang/ShareVeo3",
        filename=f"generated_videos_veo3_tar/veo3_videos_{i}.tar",
        repo_type="dataset"
    )
\end{lstlisting}

\noindent The complete list of sampled videos is provided in \texttt{veo3\_sampled\_videos.csv}.

\noindent \textbf{Sora Sampling.}
We manually collected 1.7k videos from distinct users on the OpenAI Sora public explore feed. 
The complete list of sampled videos is provided in \texttt{sora\_sampled\_videos.csv}.

\begin{table}[htbp]
\caption{\textbf{Overview of video sources and characteristics in the \textit{ComGenVid} dataset.}}

\begin{adjustbox}{max width=\textwidth}
\begin{tabular}{lcccccccc}
	\toprule
	Video Source         & Type & Length Range & Length (Mean±Std) & Resolution        & Number of Pixels (Mean±Std) & FPS (Mean±Std) & \textbf{Total Count} \\
	\midrule
	MSVD \cite{chen2011collecting}                 & Real & 2-60s        & 9.68±6.27s        & 160×112-1920×1080 & 0.29±0.35M        & 29.1±8.6       & 1700                 \\
	Sora \cite{sora2024openai}                 & Fake & 4-20s        & 6.01±2.26s        & 480×480-720×1080  & 0.36±0.05M        & 30.0±0.0       & 1700                 \\
	VEO3  \cite{veo32025deepmind}                  & Fake & 8s           & 8.00±0.00s        & 1280×720          & 0.92±0.00M        & 24.0±0.0       & 1700                 \\
	\textbf{Total Count} & -    & -            & -                 & -                 & -                 & -              & 5100                 \\
	\bottomrule
\end{tabular}
\end{adjustbox}
\label{tab:ComGenVidTable}
\end{table}

\subsection{GenVideo}
\label{sup:genvideo_benchmark}
We obtained the AI-generated videos for GenVideo~\citep{chen2024demamba} from the official \href{https://modelscope.cn/collections/Gen-Video-7de46cd6846f4e}{ModelScope
 collection}. Because the MSR-VTT~\cite{xu2016msr} real videos available on ModelScope are limited to 3 FPS, we downloaded the original MSR-VTT dataset from the \href{https://www.kaggle.com/datasets/khoahunhtngng/msrvtt}{khoahunhtngng
 Kaggle repository} to access higher-frame-rate versions.

All evaluated videos are uniformly sampled at 8 FPS for a 2-second duration (16 frames). The only exceptions are HotShot-XL and MoonValley~\citep{hotshotxl2023,moonvalley2022}, which produce clips shorter than 2 seconds, these are compared against real 1-second videos at 8 FPS (8 frames). Generative models that produce less than 2-second clips are excluded from the ablation study.
Because MSR-VTT contains far more real videos (10k) than any generative model, we selected 1.4k videos from it, following our pairwise comparison protocol (Section~\ref{pairwise_comparison_protocol}).
Comprehensive GenVideo benchmark statistics are provided in Table~\ref{tab:inf_gen_video}.


\begin{table}[htbp]
\centering
\caption{\textbf{Comprehensive statistics of the GenVideo \cite{chen2024demamba} dataset used in our evaluations.}}
\begin{adjustbox}{max width=\textwidth}
\begin{tabular}{lcccccccc}
\toprule
Video Source & Type & Length Range & Length (Mean±Std) & Resolution & Number of Pixels (Mean±Std) & FPS Range & FPS (Mean±Std) & \textbf{Total Count} \\
\midrule
MSR-VTT \cite{xu2016msr} & Real & 10-30s & 14.80±5.04s & 320×240 & 0.08±0.00M & 10-30 & 27.3±3.3 & 1400 \\
Crafter \cite{chen2023videocrafter1} & Fake & 2s & 2.00±0.00s & 1024×576 & 0.59±0.00M & 8 & 8.0±0.0 & 188 \\
Gen2 \cite{esser2023structure} & Fake & 4s & 4.00±0.00s & 896×504-1408×768 & 0.77±0.31M & 24 & 24.0±0.0 & 1380 \\
Lavie \cite{wang2025lavie} & Fake & 2-3s & 2.27±0.27s & 512×320 & 0.16±0.00M & 8-24 & 16.0±8.0 & 1400 \\
ModelScope \cite{wang2023modelscope} & Fake & 4s & 4.00±0.00s & 448×256-1280×720 & 0.70±0.36M & 8 & 8.0±0.0 & 700 \\
MorphStudio \cite{morphstudio2023} & Fake & 2s & 2.00±0.00s & 1024×576 & 0.59±0.00M & 8 & 8.0±0.0 & 700 \\
Show\_1 \cite{zhang2025show} & Fake & 4s & 3.62±0.00s & 576×320 & 0.18±0.00M & 8 & 8.0±0.0 & 700 \\
Sora \cite{sora2024openai} & Fake & 9-60s & 16.78±10.96s & 512×512-1920×1088 & 1.43±0.65M & 30 & 30.0±0.0 & 56 \\
WildScrape \cite{wei2023dreamvideocomposingdreamvideos} & Fake & 2-251s & 7.41±18.79s & 256×256-2286×1120 & 0.48±0.41M & 8-45 & 16.9±9.7 & 529 \\
HotShot ~\citep{hotshotxl2023}& Fake & 1s & 1.00±0.00s & 672×384 & 0.26±0.00M & 8 & 8.0±0.0 & 700 \\
MoonValley \cite{moonvalley2022} & Fake & 1.82s & 1.82±0.00s & 1184×672 & 0.80±0.00M & 50 & 50.0±0.0 & 626 \\
\textbf{Total Count} & - & - & - & - & - & - & - & 8379 \\
\bottomrule
\end{tabular}
\end{adjustbox}
\label{tab:inf_gen_video}
\end{table}

\subsection{VideoFeedback}

We gather the Videofeedback ~\cite{he2024videoscore} dataset 
from the official \href{https://huggingface.co/datasets/TIGER-Lab/VideoFeedback}{Hugging Face repository}.
We evaluate only videos that are at least 2 seconds long, except for HotShot-XL~\citep{hotshotxl2023}, which generates 1 s clips and is therefore compared against real 1 s videos at 8 FPS (8 frames). in the original paper ~\citep{he2024videoscore}, each clip is assigned a dynamic-degree score (1–4) indicating how clearly its motion can be distinguished from a static image. We retain only the highest-scoring videos (levels 3–4). for the full statistics of the VideoFeedback benchmark, refer to Table \ref{tab:inf_vf}.


\begin{table}[htbp]
\centering
\caption{\textbf{Comprehensive statistics for the VideoFeedback \cite{he2024videoscore} benchmark used in our evaluations.}}

\begin{adjustbox}{max width=\textwidth}
\begin{tabular}{lcccccccc}
\toprule
Video Source & Type & Length Range & Length (Mean±Std) & Resolution & Number of Pixels (Mean±Std) & FPS Range & FPS (Mean±Std) & \textbf{Total Count} \\
\midrule
DiDeMo \cite{anne2017localizing} & Real & 3s & 3.00±0.00s & 352×288-640×1138 & 0.27±0.09M & 8 & 8.0±0.0 & 1861 \\
Panda70M \cite{chen2024panda} & Real & 2-3s & 2.38±0.48s & 384×288-640×360 & 0.23±0.01M & 8 & 8.0±0.0 & 1861 \\
AnimateDiff \cite{guo2023animatediff} & Fake & 2s & 2.00±0.00s & 512×512 & 0.26±0.00M & 8 & 8.0±0.0 & 992 \\
Fast-SVD \cite{blattmann2023stable} & Fake & 3s & 3.00±0.00s & 768×432 & 0.33±0.00M & 8 & 8.0±0.0 & 959 \\
LVDM \cite{he2022latent} & Fake & 2s & 2.00±0.00s & 256×256 & 0.07±0.00M & 8 & 8.0±0.0 & 2973 \\
LaVie \cite{wang2025lavie} & Fake & 2s & 2.00±0.00s & 256×160-512×320 & 0.13±0.06M & 8 & 8.0±0.0 & 2789 \\
ModelScope \cite{wang2023modelscope} & Fake & 2s & 2.00±0.00s & 256×256 & 0.07±0.00M & 8 & 8.0±0.0 & 3722 \\
Pika \cite{pika2023} & Fake & 3s & 3.00±0.00s & 768×640 & 0.49±0.00M & 8 & 8.0±0.0 & 1906 \\
Sora \cite{sora2024openai} & Fake & 2-3s & 2.73±0.27s & 512×512-1920×1088 & 1.25±0.58M & 8 & 8.0±0.0 & 898 \\
Text2Video \cite{khachatryan2023text2video} & Fake & 2s & 2.00±0.00s & 256×256 & 0.07±0.00M & 8 & 8.0±0.0 & 3722 \\
VideoCrafter2 \cite{chen2024videocrafter2} & Fake & 2s & 2.00±0.00s & 512×320 & 0.16±0.00M & 8 & 8.0±0.0 & 3543 \\
ZeroScope \cite{zeroscope_v2_2024} & Fake & 3s & 3.00±0.00s & 256×256 & 0.07±0.00M & 8 & 8.0±0.0 & 2022 \\
Hotshot-XL ~\citep{hotshotxl2023} & Fake & 1s & 1.00±0.00s & 512×512 & 0.26±0.00M & 8 & 8.0±0.0 & 2736 \\

\textbf{Total Count} & - & - & - & - & - & - & - & 29984 \\
\bottomrule
\end{tabular}
\end{adjustbox}
\label{tab:inf_vf}
\end{table}

\subsection{VATEX (Calibration set)}

We obtained the VATEX dataset \cite{wang2019vatex} from the 
 khaledatef1's Kaggle repositories. The dataset is distributed across three parts: \href{https://www.kaggle.com/datasets/khaledatef1/vatex0110}{Vatex
 1}, \href{https://www.kaggle.com/datasets/khaledatef1/vatex01101}{Vatex
 2}, and \href{https://www.kaggle.com/datasets/khaledatef1/vatex011011}{Vatex
 3}. We downloaded all parts using the Kaggle command-line interface:
\begin{lstlisting}[language=bash]
kaggle datasets download -d khaledatef1/vatex0110 --unzip
kaggle datasets download -d khaledatef1/vatex01101 --unzip
kaggle datasets download -d khaledatef1/vatex011011 --unzip
\end{lstlisting}

\noindent For the complete statistics of VATEX \cite{wang2019vatex} calibaration set, see Table~\ref{tab:rep_vatex}.

\begin{table}[htbp]
\centering
\caption{\textbf{Comprehensive statistics of the VATEX \cite{wang2019vatex} calibration set.}}

\begin{adjustbox}{max width=\textwidth}
\begin{tabular}{lcccccccc}
\toprule
Video Source & Type & Length Range & Length (Mean±Std) & Resolution & Number of Pixels (Mean±Std) & FPS Range & FPS (Mean±Std) & \textbf{Total Count} \\
\midrule
VATEX \cite{wang2019vatex} & Real & 2-10s & 9.68±1.10s & 128×88-720×1280 & 0.44±0.37M & 8-30 & 27.2±5.0 & 33976 \\
\textbf{Total Count} & - & - & - & - & - & - & - & 33976 \\
\bottomrule
\end{tabular}
\end{adjustbox}
\label{tab:rep_vatex}
\end{table}

\subsection{Other datasets}
\label{pe_kinetics_calib_set}
We collected approximately 1.5k real videos from Kinetics400~\cite{kay2017kinetics} and PE~\cite{bolya2025perception}. The Kinetics400 clips were taken from the test split available in the \href{https://github.com/cvdfoundation/kinetics-dataset}{cvdfoundation repository}, while the PE videos were downloaded from the \href{https://huggingface.co/datasets/facebook/PE-Video}{Facebook PE Hugging Face} page. These datasets were used exclusively for the calibration set ablation experiment reported in Fig.~6(c) in the main paper. A detailed specification of this calibration set configuration is provided in Table~\ref{tab:rep_SotaGen}. A complete list of the sampled videos is provided in \texttt{pe\_kinetics400\_sampled\_videos.csv}.

\begin{table}[htbp]
\centering
\caption{\textbf{Comprehensive statistics of the Kinetics+PE calibration set.}}
\begin{adjustbox}{max width=\textwidth}
\begin{tabular}{lcccccccc}
\toprule
Video Source & Type & Length Range & Length (Mean±Std) & Resolution & Number of Pixels (Mean±Std) & FPS Range & FPS (Mean±Std) & \textbf{Total Count} \\
\midrule
Kinetics400 \cite{kay2017kinetics} & Real & 2-10s & 9.62±1.19s & 128×96-1280×720 & 0.55±0.38M & 24-30 & 26.9±3.0 & 1496 \\
PE \cite{bolya2025perception} & Real & 5-60s & 16.49±9.37s & 608×254-608×1152 & 0.21±0.03M & 24-60 & 37.7±14.5 & 1500 \\
\textbf{Total Count} & - & - & - & - & - & - & - & 2996 \\
\bottomrule
\end{tabular}
\end{adjustbox}
\label{tab:rep_SotaGen}
\end{table}
\section{Additional results and experimental details}
\label{supp: exps}
Unless stated otherwise, all ablation studies are conducted on the GenVideo \cite{chen2024demamba} benchmark, restricted to generative-model videos sampled at 8 FPS and 2 seconds duration (16 frames). The only exceptions are HotShot-XL and MoonValley~\citep{hotshotxl2023,moonvalley2022}, which generate videos shorter than 2 seconds and are therefore omitted from our ablation experiments. We use DINOv3 \cite{siméoni2025dinov3} as the frame-level embedder and the VATEX \cite{wang2019vatex} dataset as the calibration set.

\subsection{Derivative order ablations}

\begin{figure*}[t]
    \centering
    \begin{subfigure}{0.3\textwidth}
        \centering
        \includegraphics[width=\linewidth]{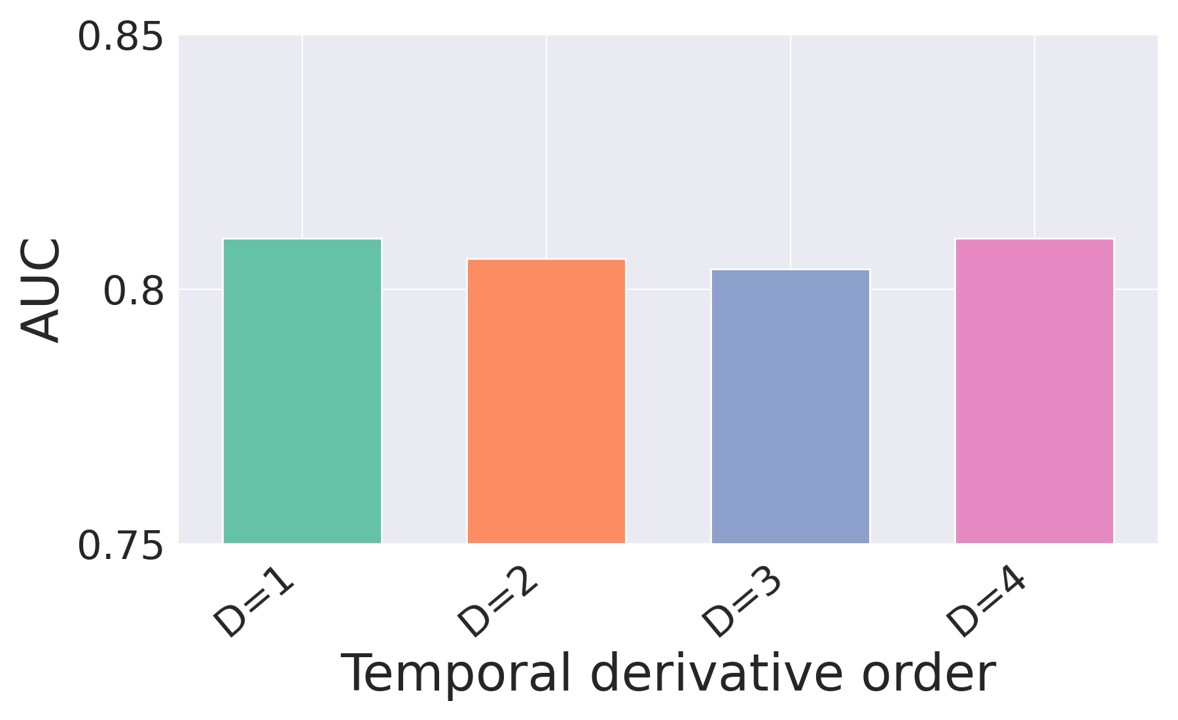}
        \caption{AUC - different derivative orders.}
        \label{fig:deriv_auc}
    \end{subfigure}\hfill
    \begin{subfigure}{0.3\textwidth}
        \centering
        \includegraphics[width=\linewidth]{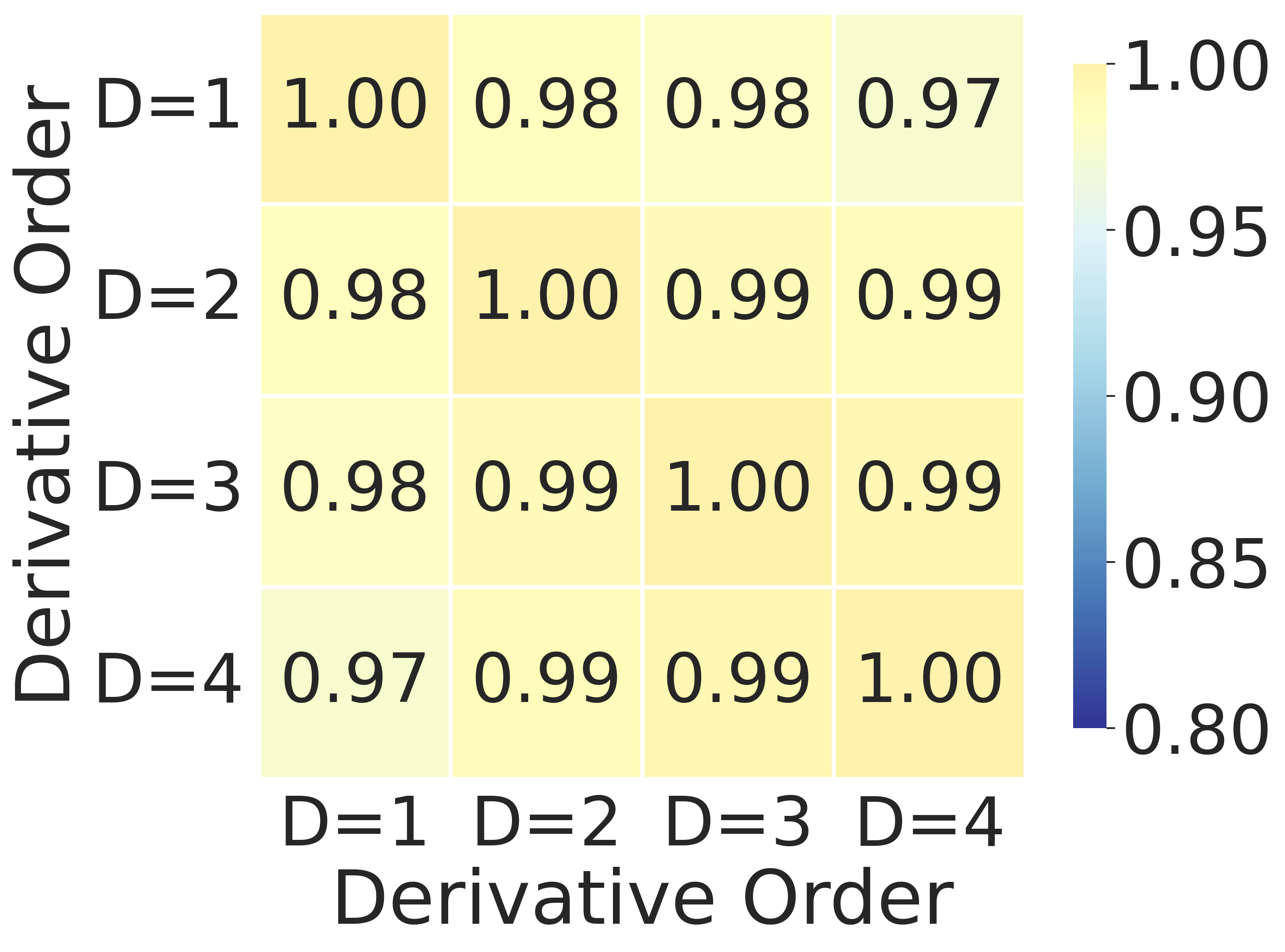}
        \caption{Pearson correlation.}
        \label{fig:deriv_corr_pearson}
    \end{subfigure}\hfill
    \begin{subfigure}{0.3\textwidth}
        \centering
        \includegraphics[width=\linewidth]{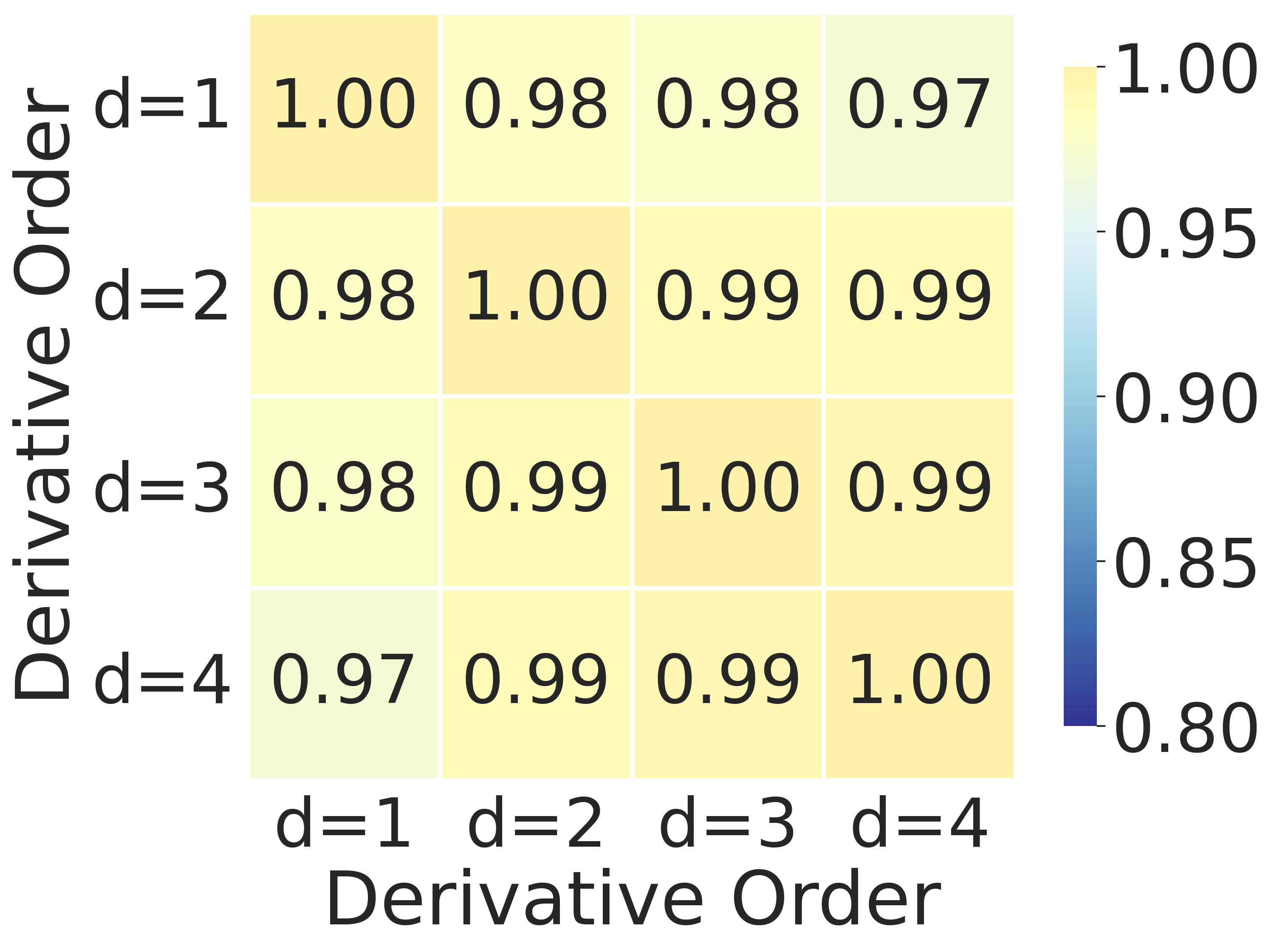}
        \caption{Spearman correlation.}
        \label{fig:deriv_corr_spearman}

    \end{subfigure}
    \caption{\textbf{Ablations on temporal derivative order.} 
    STALL performance for different temporal derivative orders \(D \in \{1,2,3,4\}\): 
    (a) AUC across orders, 
    (b) Pearson correlation between scores from different orders, 
    and (c) Spearman correlation between scores from different orders.}
    \label{fig:high_deriv_results}
\end{figure*}

We isolate the effect of higher-order temporal differences by keeping the entire STALL pipeline fixed and changing only the temporal derivative order \(D \in \{1,2,3,4\}\). For each video $v = \{f_t\}_{t=1}^{T}$, we extract frame-wise  embeddings \(E(v) \in \mathbb{R}^{T \times d}\), compute the \(D\)-th order finite-difference trajectory along time, and apply frame-wise \(\ell_2\) normalization. Importantly, we fit a separate whitening transform for each temporal derivative order, yielding parameters \((\mu_{\Delta=i}, W_{\Delta=i})\) for every \(i \in D\), estimated on the corresponding derivative trajectories from the VATEX \cite{wang2019vatex} calibration set. The temporal log-likelihood sequence for each video is aggregated using the same statistic 
as in the main STALL score, and we then average this temporal percentile with the spatial log-likelihood percentile to obtain a single score for each derivative order. An example implementation of the derivative and normalization computation is shown below:

\begin{lstlisting}
def differences_vec(features):
    return features[:, 1:, :] - features[:, :-1, :]

def temporal_diff_with_order(features, derivative_order: int):
    """
    Apply a temporal finite-difference operator of the given order
    to frame-wise embeddings and L2-normalize the result.

    Args:
        frame_embeddings: array of shape [N, T, d]
        derivative_order: positive integer order of the derivative

    Returns:
        Array of shape [N, T - derivative_order, d].
    """
    if derivative_order < 1:
        raise ValueError("derivative_order must be positive")
    for _ in range(derivative_order):
        features = differences_vec(features)
    norms = np.linalg.norm(features, axis=-1, keepdims=True) + 1e-8
    return features / norms
\end{lstlisting}
Ablation results across all derivative orders, including AUC, Pearson correlation, and Spearman correlation, are summarized in Figure~\ref{fig:high_deriv_results}.
All derivative orders are strongly correlated and produce very similar performance.


\subsection{spatial and temporal aggregation methods}

\subsubsection{Components ablation}
\begin{figure*}[t]
    \centering
    
    \begin{subfigure}{0.49\textwidth}
        \centering
        \includegraphics[width=\linewidth]{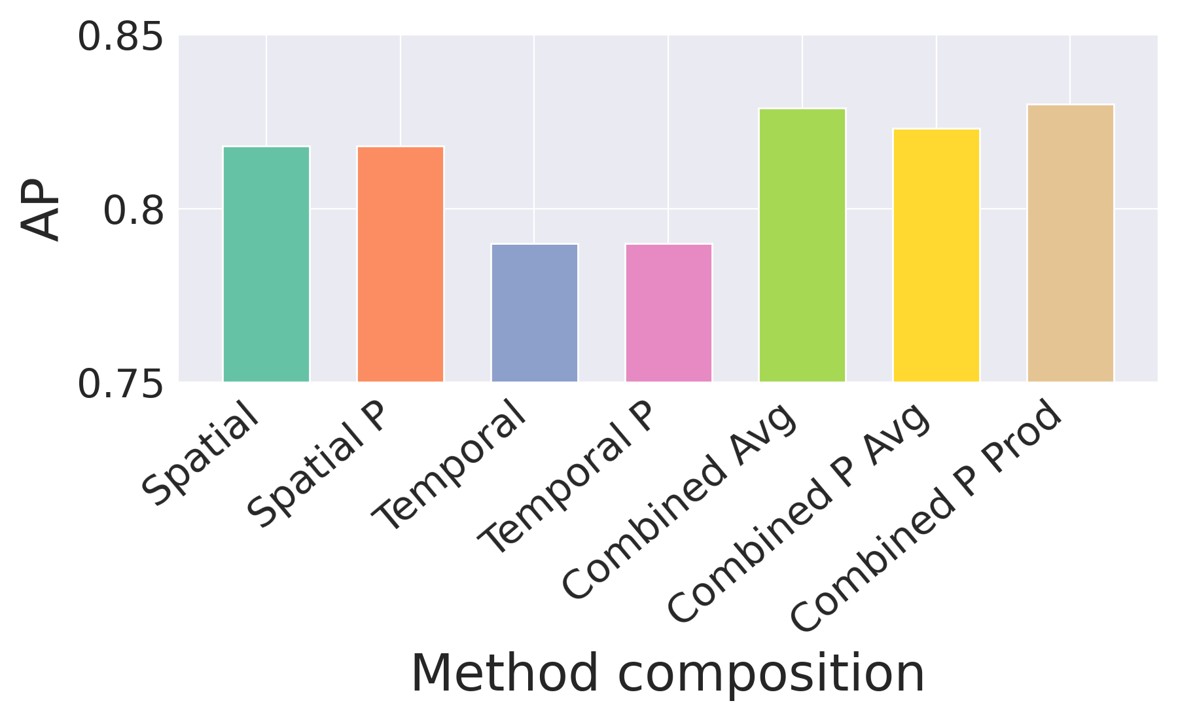}
        \caption{AP - different components}
        \label{fig:comps_ap}
    \end{subfigure}\hfill
    \begin{subfigure}{0.49\textwidth}
        \centering
        \includegraphics[width=\linewidth]{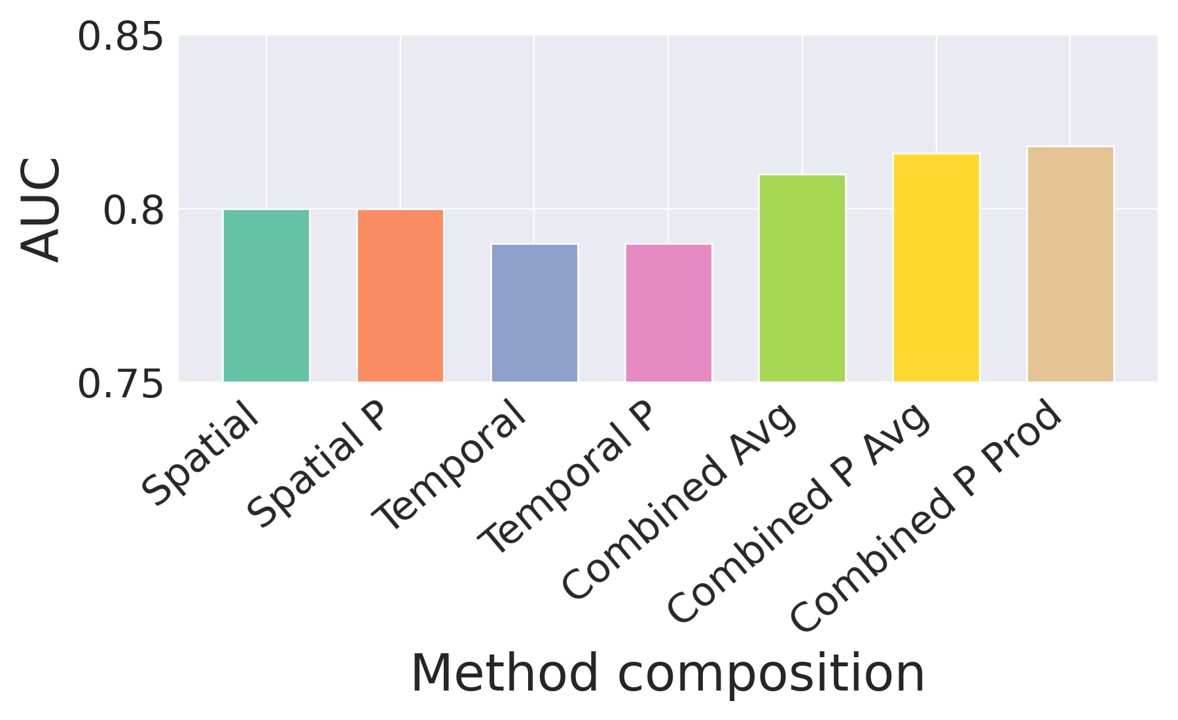}
        \caption{AUC - different components}
        \label{fig:comps_auc}
    \end{subfigure}\hfill


    \caption{\textbf{Components ablation.}
    Bar plots of spatial-only, temporal-only, and combined detectors, using both raw scores and percentile-ranked scores.
    Spatial and Temporal denote single-branch detectors, and Combined refers to the fused Spatial and Temporal detector.
    Configurations labeled with ``P'' 
    use percentile-ranked scores, where ``P'' stands for percentile. ``Avg'' stands for average of the two scores and ``Prod'' stands for the product of the two scores.}
    \label{fig:comps_ap_auc}
\end{figure*}

We analyze the contribution of the spatial and temporal branches by evaluating three detector variants: a spatial-only model, a temporal-only model, and a combined model that fuses both branches. For each variant, we consider raw scores and percentile-ranked scores, and for the combined model we also test mean-based and product-based fusions of the spatial and temporal percentile scores. For every configuration, we report the average AP and AUC across all three benchmarks. See results in Fig.~\ref{fig:comps_ap_auc}. 

\subsubsection{Aggregation ablation}
\begin{figure*}[t]
    \centering    
    \begin{subfigure}{0.33\textwidth}
        \centering
        \includegraphics[width=\linewidth]{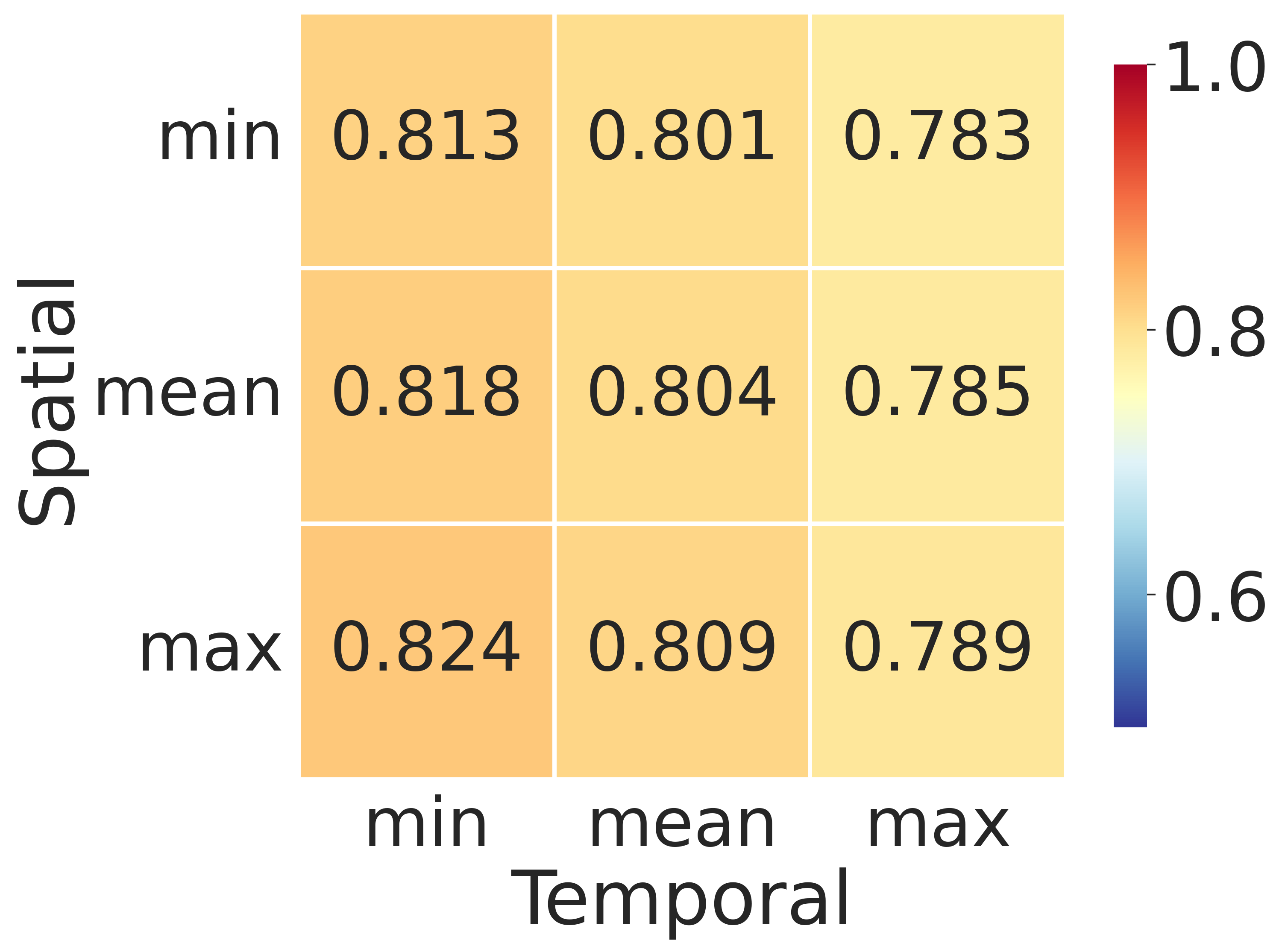}
        \caption{AP for different spatial and temporal aggregation pairs.}
        \label{fig:agg_ap}
    \end{subfigure}
    \hspace{0.05\textwidth}
    \begin{subfigure}{0.33\textwidth}
        \centering
        \includegraphics[width=\linewidth]{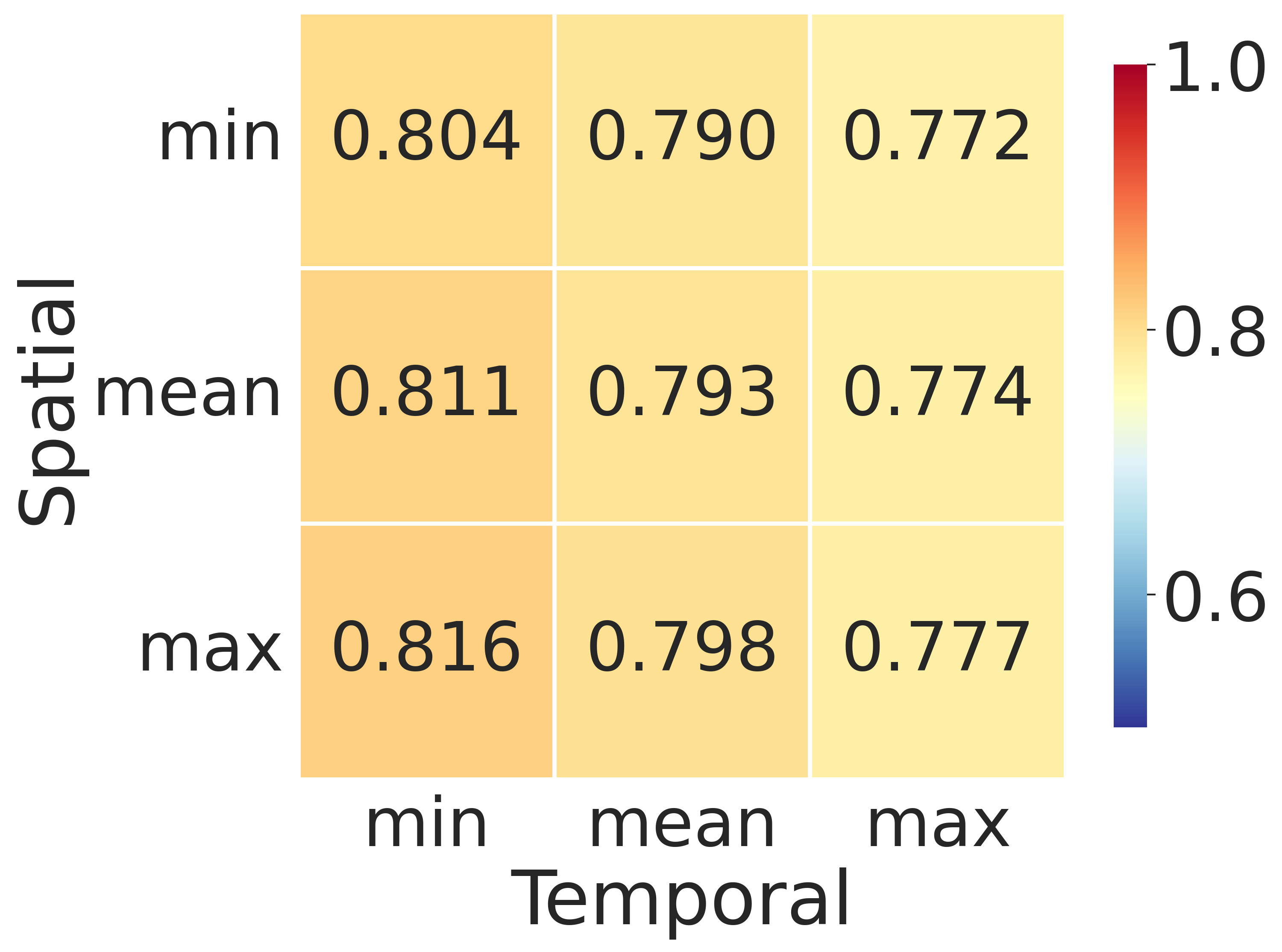}
        \caption{AUC for different spatial and temporal aggregation pairs.}
        \label{fig:agg_auc}
    \end{subfigure}
    \caption{\textbf{Frame-level aggregation ablation.}
    Heatmaps of AP and AUC, for different combinations of spatial and temporal frame-level aggregation operators.
    Rows correspond to the spatial aggregation and columns to the temporal aggregation, while all other parts of the pipeline follow the default configuration described in the main paper.}
    \label{fig:agg_auc_ap}
\end{figure*}

To analyze the effect of the frame-level aggregation operators, we fix our pipeline as defined in the main paper and vary only the frame-level aggregation used within each branch. We sweep over \texttt{min}, \texttt{mean}, and \texttt{max} for both the spatial and temporal components. For each spatial and temporal aggregation pair, we compute the average AUC and AP across all three benchmarks and report the resulting scores as heatmaps in Supp. Fig.~\ref{fig:agg_auc_ap}.

\subsection{Sampling a single frame difference for temporal whitening}

In this ablation, we test whether temporal whitening requires all frame-to-frame transitions or can be reliably estimated from a single transition per video. We keep the spatial branch (and its whitening transform) fixed, and re-compute the temporal whitening statistics \((\mu_{\Delta}, W_{\Delta})\) twice on the calibration set: (i) using all normalized frame differences from all videos, and (ii) using only one randomly sampled frame difference per video. Re-evaluating the combined score under these two temporal-whitening calibration variants yields essentially the same performance, with average AUC \textbf{0.8110} and average AP \textbf{0.8046} for (i) and average AUC \textbf{0.8105} and average AP \textbf{0.8044} for (ii) (Pearson correlation $\textbf{0.9994}$, Spearman correlation $\textbf{0.9992}$ between (i) and (ii)), indicating that our temporal whitening is robust to the number of transitions used from each video for its estimation.

\subsection{Temporal likelihood under temporal perturbations}

\begin{figure}[t]
  \centering
  \includegraphics[width=0.7\linewidth]{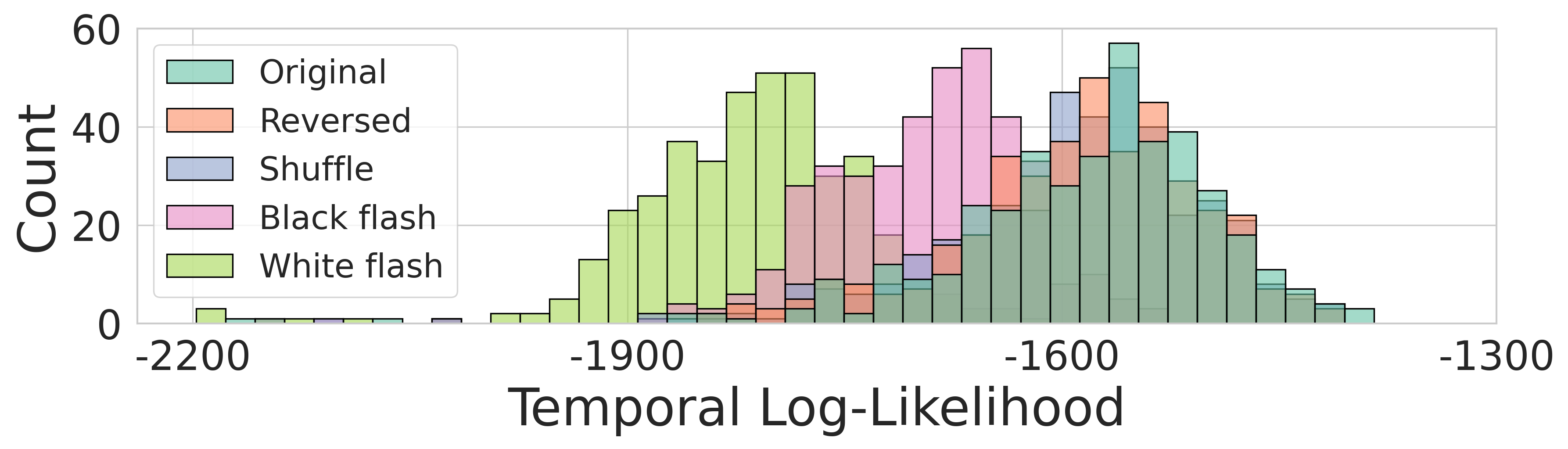}
  \caption{%
    \textbf{Temporal scores under temporal perturbations.}
    Each condition applies a different perturbation to 400 real MSR-VTT videos: \emph{Original} (no perturbation), \emph{Reversed} (frames in reverse order), \emph{Shuffle} (consecutive frames shuffled), \emph{Black flash} (one black frame inserted mid-video), and \emph{White flash} (one white frame inserted mid-video).
    The temporal likelihood is largely unaffected by reversal and shuffling, since adjacent-frame difference statistics are preserved; however, abrupt flash frames cause a strong drop in the score, with white flashes producing a larger degradation than black flashes.
  }
  \label{fig:temporal_ll_perturbations}
\end{figure}

We analyze the temporal likelihood under realistic perturbations: reversing frame order (rewind), shuffling consecutive frames, and inserting black or white frames, representing data transmission issues~\citep{zeng2024benchmarking}.
As shown in \Cref{fig:temporal_ll_perturbations}, the method is robust to reversal and shuffling, since the score depends only on statistics of adjacent-frame differences, which are largely preserved under these operations.
In contrast, inserting a flash frame introduces an abrupt temporal inconsistency that strongly reduces the likelihood.
The experiment is conducted on 400 real videos from MSR-VTT~\citep{xu2016msr}.
The temporal likelihood is stable under common temporal distortions but responds strongly to abrupt temporal anomalies.

\subsection{Image perturbation experiment details} 

To explore robustness to perturbations, we apply four standard image corruptions to GenVideo~\citep{chen2024demamba} frames at inference time only, keeping the calibration set fixed. We use a reduced GenVideo subset with 250 videos per generative model and corrupt every video with each perturbation type at all five predefined severity levels. We then measure the impact of each corruption setting on detection performance. We additionally include level~0, which corresponds to uncorrupted frames (no perturbation). As shown in Fig.~7(b) of the main paper, STALL maintains strong separation across all perturbation types and severities.

\noindent The specific parameter settings for each perturbation type and level are summarized in Tab.~\ref{tab:image_perturbations}. Gaussian blur severity is controlled by the blur radius $r$, with larger $r$ producing stronger smoothing. JPEG compression reduces the image quality parameter, where lower quality introduces stronger compression artifacts. The random resized crop perturbation is parameterized by scale ranges, where a random crop is sampled at each application: sometimes removing more content and sometimes less. Higher severity levels use wider ranges with smaller minimum scales, increasing the chance of more aggressive crops. Gaussian noise severity is adjusted by increasing the standard deviation of zero-mean noise added to each frame, which progressively degrades fine texture while preserving global structure.

\begin{table}[t]
	\centering
	\caption{\textbf{Image perturbation details.} These perturbations are used in the robustness ablation (Sec.~D.5). Levels 1--5 span the full range of corruption strength for each type; level~0 corresponds to no perturbation.}
	\label{tab:image_perturbations}
	\begin{tabular}{lll}
		\toprule
		Perturbation & Implementation                                   & Severity levels (1$\rightarrow$5) \\
		\midrule
		Gaussian blur
		             & \texttt{TF.gaussian\_blur}
		             & $(r,\sigma,k) =$                                                                     \\
		             &
		             & $\{(1,0.5,3),(2,1.0,5),$                                                               \\
		             &
		             & $(3,1.5,7),(5,2.5,11),$                                                              \\
		             &
		             & $(10,5.0,21)\}$                                                                        \\
		JPEG compression
		             & \texttt{PIL.save(...,}
		             & JPEG quality                                                                         \\
		             & \texttt{format="JPEG", quality)}
		             & $q \in \{80,50,30,10,1\}$ \\
		Random resized crop
		             & \texttt{transforms.RandomResizedCrop}
		             & scale ranges                                                                         \\
		             &
		             & $\{(0.85,0.9),(0.7,0.85),$                                                            \\
		             &
		             & \quad $(0.5,0.8),(0.3,0.9),$                                                        \\
		             &
		             & \quad $(0.08,1.0)\}$                                                                 \\
		Gaussian noise
		             & \texttt{torch.randn\_like}, \texttt{torch.clamp}
		             & $\sigma \in \{0.02,0.05,0.1,0.2,0.5\}$                                               \\
		\bottomrule
	\end{tabular}
\end{table}

\noindent Perturbation examples can be found in Figs.~\ref{fig:noise_mandaril} and~\ref{fig:noise_boat}.

\begin{figure}[t]
  \centering
  \perturbationtabular{figs/perturbation_ex/mandrill_demo}
  \caption{\textbf{Perturbations example.} mandrill.}
  \label{fig:noise_mandaril}
\end{figure}

\begin{figure}[t]
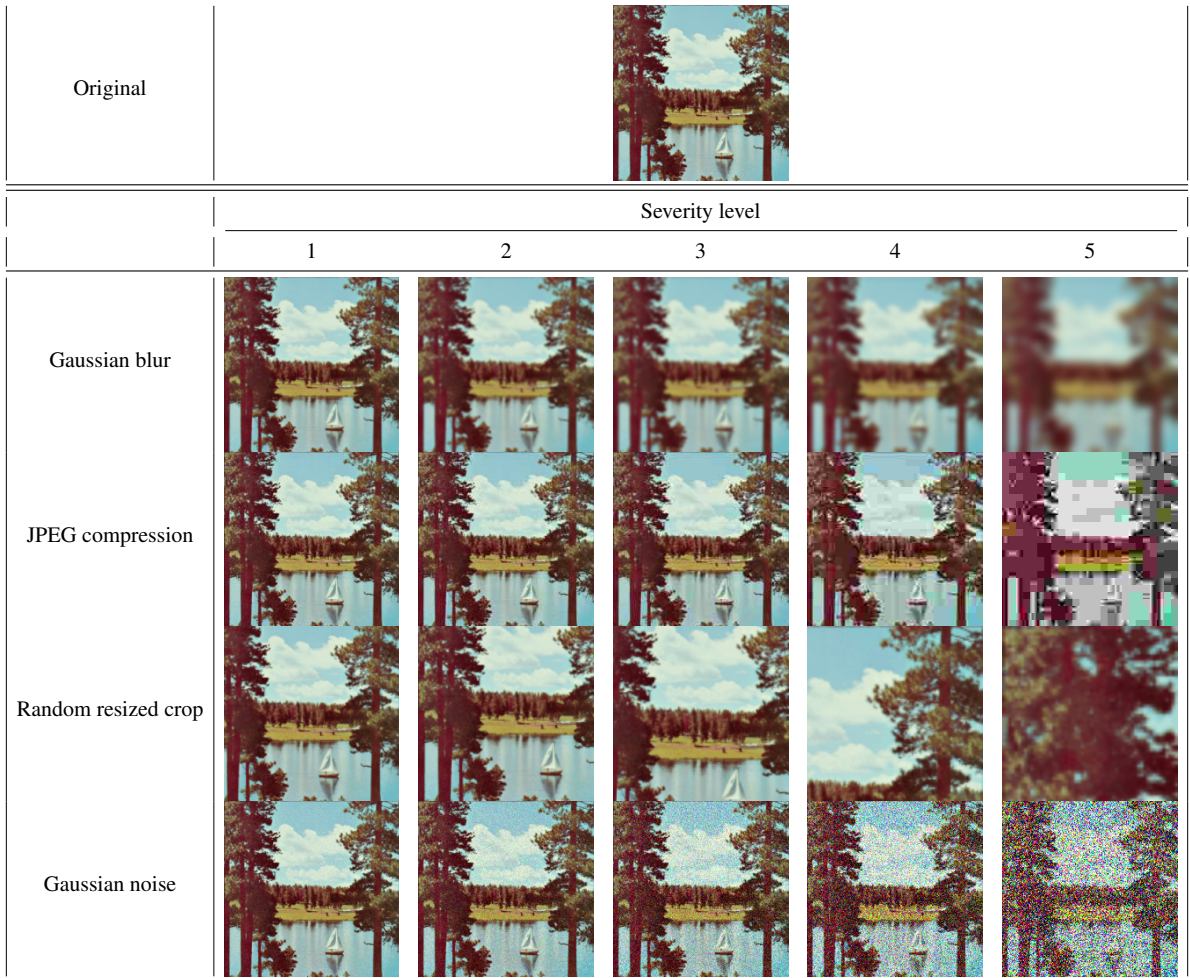

  \centering
  \perturbationtabular{figs/perturbation_ex/sail_boat}
  \caption{\textbf{Perturbations example.} Sailboat on lake.}
  \label{fig:noise_boat}
\end{figure}

\subsection{Additional experimental details}

\subsubsection{Step size ablation}

To assess sensitivity to temporal sampling rate, we vary the frame step size applied to videos already sampled at 8 FPS. For step size $s \in \{1,2,3,4\}$, we subsample every $s$-th frame from all videos, reducing the effective frame rate by a factor of $s$ (yielding 8, 4, 2.67, and 2 FPS respectively). Specifically, this subsampling is applied uniformly to both the VATEX \cite{wang2019vatex} calibration set used to estimate the whitening transform and to the GenVideo \cite{chen2024demamba} test videos being evaluated. We then compute the combined spatial and temporal score using these subsampled sequences. From another perspective, the setup can be viewed as using a larger temporal stride of $s$ between successive differences, without overlap between stride segments in the original video. This ablation tests whether the detector remains robust when operating at lower effective frame rates, which is relevant for computational efficiency. 
The average AUC results are shown in Fig.~8(a) of the main paper.

\subsubsection{FPS ablation} 

To evaluate robustness to different frame rates, we subsample only the inference videos while keeping the whitening transform fixed (estimated on VATEX \cite{wang2019vatex} at 8 FPS as in all other experiments). We select videos from GenVideo \cite{chen2024demamba} that are originally at 24 FPS with at least 2 seconds duration (425 videos from Gen2 \cite{esser2023structure}, 425 from Lavie \cite{wang2025lavie}, and 42 from WildScrape \cite{wei2023dreamvideocomposingdreamvideos}, and a matching number of real videos from MSR-VTT~\cite{xu2016msr}), isolating the effect of frame rate from other factors. In contrast to our standard 8 FPS setup, in this experiment we downsample the videos to target frame rates in $\{2, 4, 8, 12, 24\}$ FPS using exact subsampling. The downsampling factor and target frame rate are chosen such that
\begin{equation*}
\frac{\text{current\_fps}}{\text{target\_fps}} = n \in \mathbb{N},
\end{equation*}
so the downsampled sequence is obtained by retaining every $n$-th frame.
 This ensures deterministic frame selection without approximations. The score is computed on the downsampled sequences. 
Results, summarized in Fig.~8(c) of the main paper, show that performance is essentially unchanged across this range of frame rates, indicating that our method is robust to frame rate variation and that calibrating the whitening transform at 8 FPS does not degrade inference at other frame rates.
\subsubsection{Length of video ablation} 

To evaluate robustness to video length, we follow a procedure similar in nature to the FPS ablation. We select videos from GenVideo \cite{chen2024demamba} that were originally 4 seconds in duration (1380 videos from Gen2 \cite{esser2023structure}, 700 from ModelScope \cite{wang2023modelscope}, 214 from WildScrape \cite{wei2023dreamvideocomposingdreamvideos}, 56 from Sora \cite{sora2024openai}, and 1400 real videos from MSR-VTT~\cite{xu2016msr}), and then truncate each to $\{1, 2, 3, 4\}$ seconds. The whitening transform remains fixed throughout (estimated on VATEX \cite{wang2019vatex} at 2 seconds as in all other experiments). 
The score is then computed on these truncated videos. Results, shown in Fig.~8(b) of the main paper, demonstrate that our method remains robust across this range of video durations, indicating that calibrating at 2 seconds does not degrade performance on shorter or longer clips.

\subsubsection{Backbone encoders ablation} 

To assess the impact of the feature extractor, we evaluate STALL with five different backbone encoders: three image-based encoders and two video-based encoders. For image encoders, we use DINOv3~\citep{siméoni2025dinov3} (\href{https://github.com/facebookresearch/dinov3}{\texttt{dinov3\_vitl16}}), the lightweight MobileNetV3~\citep{howard2019searching} (\texttt{mobilenetv3\_large\_100} from \texttt{timm} python pacagke), and ResNet-18~\citep{he2016deep} (from \texttt{torchvision.models}). For video encoders, we test VideoMAE~\citep{tong2022videomae} (\href{https://huggingface.co/MCG-NJU/videomae-base}{\texttt{MCG-NJU/videomae-base}}) and ViCLIP~\citep{wang2023internvid} (\href{https://huggingface.co/OpenGVLab/ViCLIP-L-14-hf}{\texttt{OpenGVLab/ViCLIP-L-14-hf}}), both from HuggingFace. All encoders are used with pretrained weights, and for image encoders we extract per-frame features and apply them independently to each frame in the video sequence.
Results are presented in Table~2 of the main paper. 

\subsubsection{Calibration set size} 

To evaluate the sensitivity of our method to the size of the calibration set used for estimating the whitening transform, we systematically vary the number of videos sampled from VATEX \cite{wang2019vatex}. We test calibration set sizes ranging from 1,000 to the full VATEX dataset (33,976 videos) in increments of 1,000 videos, evaluating each configuration across 5 random seeds to ensure statistical reliability. For each calibration set size, we randomly sample the specified number of videos from VATEX, estimate the whitening parameters on this subset, and then evaluate the resulting detector on the GenVideo \cite{chen2024demamba} benchmark. All other pipeline components remain fixed, results are presented in Fig. 7(a) of the main paper.

\subsubsection{Calibration set sources} 

To assess the impact of calibration set composition on detection performance, we evaluate STALL using five different calibration sets drawn from diverse real video sources. We test: (1) VATEX \cite{wang2019vatex} (33,976 videos), (2) GenVideo \cite{he2024videoscore} real subset from MSR-VTT \cite{xu2016msr} (8,584 videos, corresponding to all MSR-VTT clips excluded from the test set), (3) VideoFeedback \cite{chen2024demamba} combining DiDeMo \cite{anne2017localizing} (1k videos) and Panda70M \cite{chen2024panda} (1k videos), (4) a combination of Kinetics400   (1,496 videos) and PE \cite{bolya2025perception} (around 1,500 videos from each, see section \ref{pe_kinetics_calib_set} for more details), and (5) a balanced hybrid set sampling 1k videos from each of the six sources: MSR-VTT, PE, Panda70M, DiDeMo, VATEX, and Kinetics400. For each calibration set, we estimate the whitening transform using only videos from that set and evaluate the resulting detector on the GenVideo \cite{chen2024demamba} benchmark. All other pipeline components remain fixed. Results are presented in Fig. 6(c) of the main paper.

Table ~\ref{tab:results_calib_AUC} further reports average AUC across all three benchmarks for each calibration source (2K videos each), confirming stable performance across calibration choices.

\begin{table}[htbp]
\centering

\begin{adjustbox}{max width=\linewidth}
\begin{tabular}{l|ccccc}
\toprule
Benchmark & VATEX \cite{wang2019vatex}& Kinetics400 \cite{kay2017kinetics}& DiDeMo \cite{anne2017localizing} & Panda-70M \cite{chen2024panda}& MSR-VTT \cite{xu2016msr}\\
\midrule
VideoFeedback \cite{he2024videoscore}& 0.82 & 0.82 & 0.86 & 0.76 & 0.73 \\
GenVideo      \cite{chen2024demamba}& 0.78 & 0.77 & 0.76 & 0.83 & 0.83 \\
ComGenVid     & 0.82 & 0.81 & 0.87 & 0.75 & 0.76 \\
\bottomrule
\end{tabular}
\end{adjustbox}
\caption{Average AUC for different calibration datasets across all three benchmarks.}
\label{tab:results_calib_AUC}
\end{table}

More broadly, the calibration set defines the reference feature statistics of the detector.
If the test domain is not represented in the calibration set (e.g., surveillance or aerial videos),
performance may degrade, which is a limitation of the method.
Conversely, this also enables domain adaptiveness, as the detector can be adapted to a target domain
by choosing an appropriate calibration set.
Within the general video regime studied here, we observe stable behavior under calibration and test variation.

\subsection{Additional qualitative results}
\begin{figure}[H]
    \centering
    \includegraphics[width=1\linewidth]{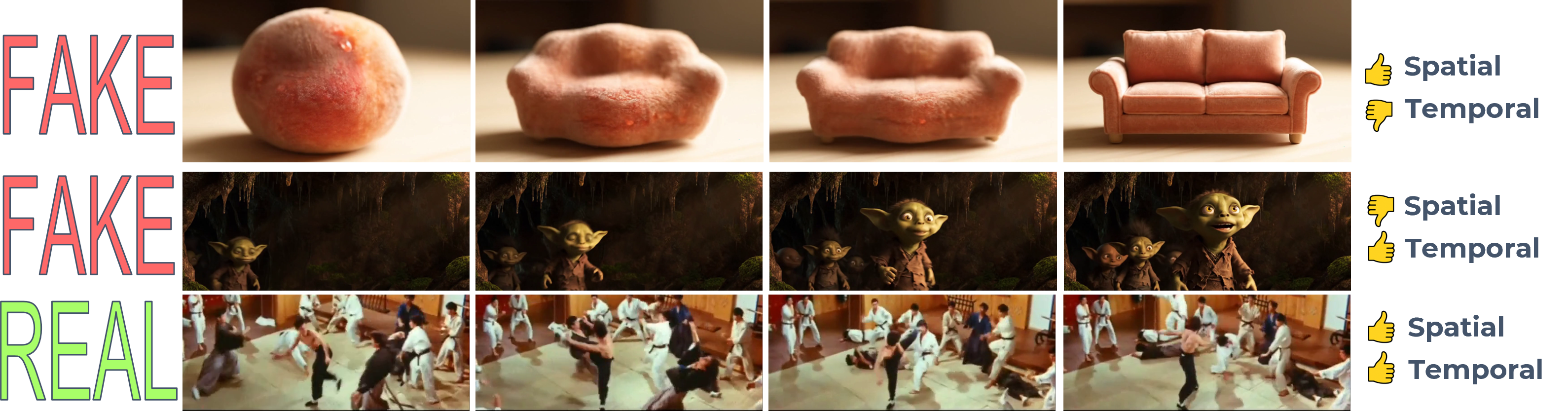}
    \caption{\textbf{Qualitative examples.} Each row shows sampled frames from a video clip, with indicators marking whether its spatial and temporal behavior appears natural or unnatural.}
    \label{fig:placeholder}
\end{figure}

\begin{figure}[H]
    \centering
    \includegraphics[width=1\linewidth]{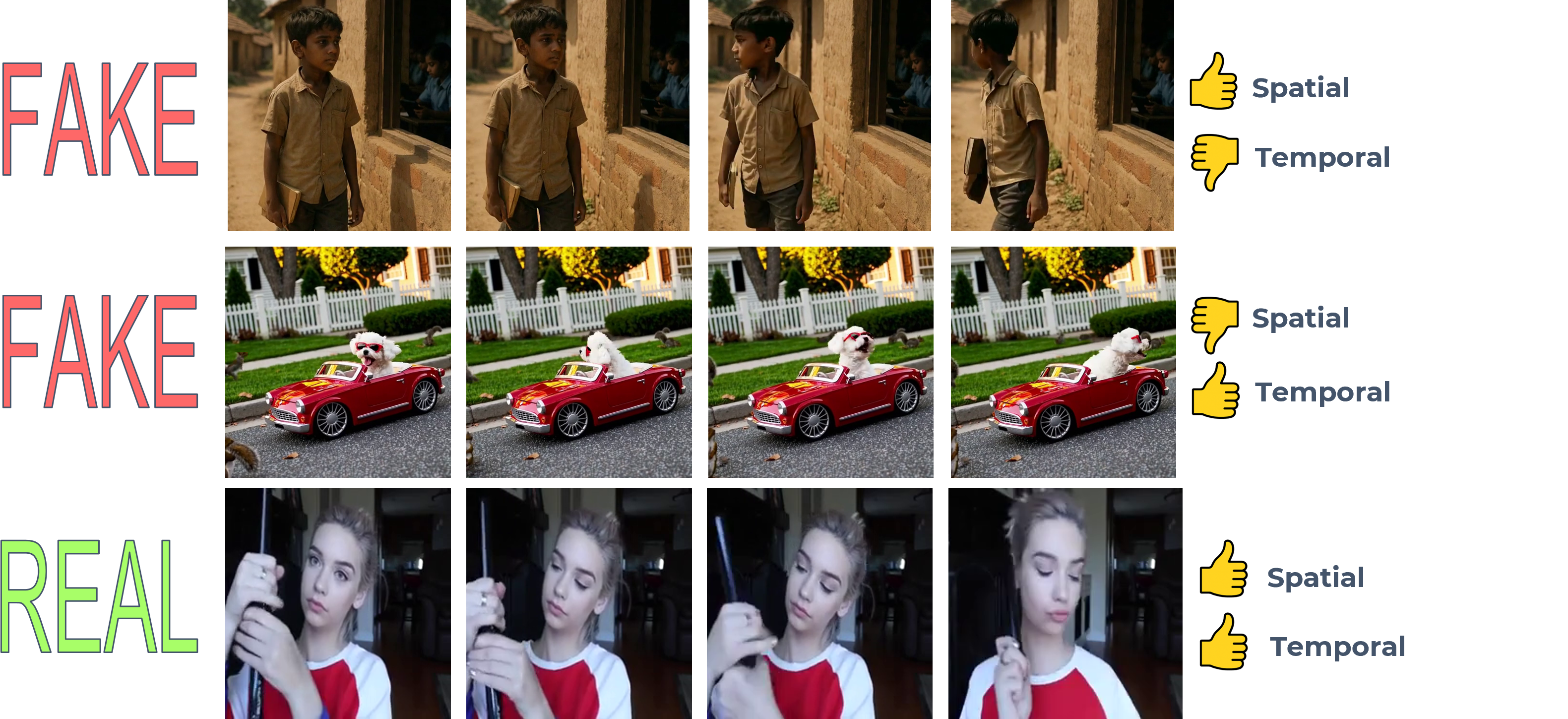}
    \caption{\textbf{Qualitative examples.} Each row shows sampled frames from a video clip, with indicators marking whether its spatial and temporal behavior appears natural or unnatural.} 
    \label{fig:placeholder}
\end{figure}
\section{Efficiency analysis}
\label{supp: eff}

We conducted a comprehensive efficiency analysis to evaluate the computational performance of all detection methods, measuring model inference time and memory usage under controlled conditions.

\subsection{Infrence time analysis}


This experiment was performed on a fixed set of 20 videos. Each method ran inference on each video separately (without batching), and we repeated this process 5 times over the same video set to account for performance variability, yielding 100 total inference evaluations per method.

All methods were initialized before timing measurements to ensure a fair comparison. We used Python's \texttt{timeit.repeat} function to measure execution times, defining the inference time of each method to include video loading. This design ensures that the measured times reflect realistic end-to-end performance, covering both data loading and inference. The complete inference time analysis is provided in Table~\ref{tab:timing}.

\begin{table}[htbp]
\caption{\textbf{Inference time comparison for all methods.}}
\centering
\label{tab:timing}
\begin{tabular}{llcc}
\toprule
Domain & Method & Mean [sec] & Std [sec] \\
\midrule
\multirow{3}{*}{Zero shot images}
&AEROBLADE \cite{ricker2024aeroblade} & 2.5266 & 0.0243 \\
&ZED \cite{cozzolino2024zero}& 1.1394 & 0.0067 \\
&RIGID \cite{he2024rigid}& 0.4363 & 0.0024 \\
\hline
\multirow{2}{*}{Supervised video}
&T2VE \cite{T2VE_2025}& 1.9950 & 0.0102 \\
&AIGVdet \cite{bai2024ai}& 5.4216 & 0.0787 \\
\hline
\multirow{3}{*}{Zero-shot video}
&D3 cos \cite{zheng2025d3}& 0.2157 & 0.0043 \\
&D3 L2 \cite{zheng2025d3}& 0.2220 & 0.0015 \\
&STALL (ours) & 0.2230 & 0.0010 \\
\bottomrule
\end{tabular}
\end{table}

\subsection{memory analysis}

To comprehensively evaluate the memory requirements of all methods, we conducted a profiling study that measures both model loading and inference memory consumption. Each method was executed in complete isolation within separate processes to eliminate any potential memory pollution or interference between measurements. We distinguished between two critical phases: (1) model loading memory, which captures the one-time cost of initializing model parameters and loading them onto CPU and GPU, and (2) inference memory, which measures the runtime memory footprint during actual video processing. For each method, we repeated measurements across multiple videos (10 videos × 3 repetitions) to ensure statistical reliability. Memory measurements were captured at two levels: CPU memory was tracked using \texttt{psutil} to monitor RAM consumption, while GPU memory was measured using PyTorch's \cite{paszke2019pytorch} CUDA memory tracking facilities to capture peak memory usage. To ensure measurement accuracy, we performed garbage collection and CUDA cache clearing between measurements
(\texttt{gc.collect()}) and GPU cache clearing (\texttt{torch.cuda.empty\_cache()} and \texttt{torch.cuda.reset\_peak\_memory\_stats()}) performed between each measurement to ensure clean memory states 
, with deliberate delays to allow the system to stabilize. A detailed analysis of memory consumption appears in Table \ref{tab:memory_all}.
\begin{table}[htbp]
	\caption{\textbf{Memory usage comparison for all methods}}
	\label{tab:memory_all}
	\resizebox{\textwidth}{!}{
		\begin{tabular}{llccccccc}
			\toprule
			domain & method       & model loading cpu (MB) & model loading gpu (MB) & inference cpu peak (MB) & inference gpu peak (MB) \\
			\midrule
			\multirow{3}{*}{zero shot images}
			       & AEROBLADE \cite{ricker2024aeroblade}   & 7875.93                & 141.02                 & 9711.18                 & 2624.43                 \\
			       & ZED \cite{cozzolino2024zero}         & 101.11                 & 16.06                  & 1707.72                 & 310.08                  \\
			       & RIGID      \cite{he2024rigid}  & 142.41                 & 327.30                 & 1373.67                 & 567.57                  \\
			\hline
			\multirow{2}{*}{supervised video}
			       & T2VE      \cite{T2VE_2025}   & 1852.16                & 1271.43                & 2795.96                 & 140.96                  \\
			       & AIGVdet   \cite{bai2024ai}   & 488.77                 & 182.59                 & 1849.03                 & 673.18                  \\
			\hline
			\multirow{3}{*}{zero-shot video}
			       & D3 cos      \cite{zheng2025d3} & 315.64                 & 1157.72                & 1470.08                 & 160.30                  \\
			       & D3 l2       \cite{zheng2025d3} & 315.61                 & 1157.72                & 1416.19                 & 160.30                  \\
			       & STALL (ours) & 321.21                 & 1166.77                & 1647.60                 & 160.30                  \\
			\bottomrule
		\end{tabular}
	}
\end{table}

\noindent The memory profiling results reveal significant variations in resource consumption across different detection approaches. Among zero-shot image methods, AEROBLADE \cite{ricker2024aeroblade} demonstrates substantially higher memory footprint during both loading and inference phases, while ZED \cite{cozzolino2024zero} achieves the most efficient performance. In the supervised video category, T2VE \cite{T2VE_2025} requires notably more GPU memory for model loading compared to AIGVdet \cite{bai2024ai}, though the latter exhibits higher inference-time consumption. Our proposed STALL method, alongside D3 \cite{zheng2025d3}, maintains a balanced and efficient memory profile with moderate CPU and GPU usage across both phases, demonstrating comparable efficiency to existing temporal consistency zero-shot method while requiring significantly fewer resources than supervised alternatives. We note that all methods receive video data through CPU memory, which contributes to the observed inference CPU peak measurements across all approaches.

\end{document}